\documentclass[a4paper,12pt,titlepage,oneside]{kuphdthesis}

\usepackage{url}
\usepackage[bottom]{footmisc}
\usepackage[caption=false,font=footnotesize]{subfig}
\usepackage{amsmath}
\usepackage{amssymb}
\usepackage{amsfonts}
\usepackage{breakcites}
\usepackage{graphicx}
\usepackage{nccmath}
\usepackage{booktabs} 
\usepackage{caption}
\usepackage{subcaption}
\usepackage{multirow}
\usepackage{adjustbox}
\usepackage{colortbl}
\usepackage{float}
\usepackage{placeins}
\usepackage[hidelinks]{hyperref}

\graphicspath{{./Figures/}}

\usepackage{xspace}
\usepackage{epstopdf}

\newcommand \COMMENT  [1] {}       %

\begin{document}

\Title{Learning Object-Centric Representations Based on Slots in Real World Scenarios} \Author{Adil Kaan Akan} \Year{September 16, 2025}
\Program{Computer Science and Engineering}

\TTitle{Ger\c cek D\"unya Senaryolar\i nda Slot Tabanl\i\ Nesne-Merkezli Temsillerin \"O\u grenilmesi
} \TYear{16 Eyl\"{u}l 2019}
\TProgram{Bilgisayar Bilimleri ve M\"{u}hendisli\u{g}i}

\Signature{Prof. Y\"{u}cel Yemez (Advisor)}
\Signature{Assoc. Prof. Aykut Erdem}
\Signature{Prof. Erkut Erdem}
\Signature{Prof. Engin Erzin}
\Signature{Asst. Prof. Ayşegül Dündar}
\prelimpages
\titlepage
\newcommand{\Perp}{\perp\!\!\! \perp}
\newcommand{\bK}{\mathbf{K}}
\newcommand{\bX}{\mathbf{X}}
\newcommand{\bY}{\mathbf{Y}}
\newcommand{\bk}{\mathbf{k}}
\newcommand{\bx}{\mathbf{x}}
\newcommand{\by}{\mathbf{y}}
\newcommand{\bhy}{\hat{\mathbf{y}}}
\newcommand{\bty}{\tilde{\mathbf{y}}}
\newcommand{\bG}{\mathbf{G}}
\newcommand{\bI}{\mathbf{I}}
\newcommand{\bg}{\mathbf{g}}
\newcommand{\bS}{\mathbf{S}}
\newcommand{\bs}{\mathbf{s}}
\newcommand{\bw}{\mathbf{w}}
\newcommand{\eye}{\mathbf{I}}
\newcommand{\bU}{\mathbf{U}}
\newcommand{\bV}{\mathbf{V}}
\newcommand{\bW}{\mathbf{W}}
\newcommand{\bn}{\mathbf{n}}
\newcommand{\bv}{\mathbf{v}}
\newcommand{\bwv}{\mathbf{wv}}
\newcommand{\bq}{\mathbf{q}}
\newcommand{\bR}{\mathbf{R}}
\newcommand{\bi}{\mathbf{i}}
\newcommand{\bj}{\mathbf{j}}
\newcommand{\bp}{\mathbf{p}}
\newcommand{\bt}{\mathbf{t}}
\newcommand{\bJ}{\mathbf{J}}
\newcommand{\bu}{\mathbf{u}}
\newcommand{\bB}{\mathbf{B}}
\newcommand{\bD}{\mathbf{D}}
\newcommand{\bz}{\mathbf{z}}
\newcommand{\bP}{\mathbf{P}}
\newcommand{\bC}{\mathbf{C}}
\newcommand{\bA}{\mathbf{A}}
\newcommand{\bZ}{\mathbf{Z}}
\newcommand{\bff}{\mathbf{f}}
\newcommand{\bF}{\mathbf{F}}
\newcommand{\bo}{\mathbf{o}}
\newcommand{\bO}{\mathbf{O}}
\newcommand{\bc}{\mathbf{c}}
\newcommand{\bT}{\mathbf{T}}
\newcommand{\bQ}{\mathbf{Q}}
\newcommand{\bL}{\mathbf{L}}
\newcommand{\bl}{\mathbf{l}}
\newcommand{\ba}{\mathbf{a}}
\newcommand{\bE}{\mathbf{E}}
\newcommand{\bH}{\mathbf{H}}
\newcommand{\bd}{\mathbf{d}}
\newcommand{\br}{\mathbf{r}}
\newcommand{\be}{\mathbf{e}}
\newcommand{\bb}{\mathbf{b}}
\newcommand{\bh}{\mathbf{h}}
\newcommand{\bhh}{\hat{\mathbf{h}}}
\newcommand{\btheta}{\boldsymbol{\theta}}
\newcommand{\bTheta}{\boldsymbol{\Theta}}
\newcommand{\bpi}{\boldsymbol{\pi}}
\newcommand{\bphi}{\boldsymbol{\phi}}
\newcommand{\bpsi}{\boldsymbol{\psi}}
\newcommand{\bPhi}{\boldsymbol{\Phi}}
\newcommand{\bmu}{\boldsymbol{\mu}}
\newcommand{\bsigma}{\boldsymbol{\sigma}}
\newcommand{\bSigma}{\boldsymbol{\Sigma}}
\newcommand{\bGamma}{\boldsymbol{\Gamma}}
\newcommand{\bbeta}{\boldsymbol{\beta}}
\newcommand{\bomega}{\boldsymbol{\omega}}
\newcommand{\blambda}{\boldsymbol{\lambda}}
\newcommand{\bLambda}{\boldsymbol{\Lambda}}
\newcommand{\bkappa}{\boldsymbol{\kappa}}
\newcommand{\btau}{\boldsymbol{\tau}}
\newcommand{\balpha}{\boldsymbol{\alpha}}
\newcommand{\nR}{\mathbb{R}}
\newcommand{\nN}{\mathbb{N}}
\newcommand{\nL}{\mathbb{L}}
\newcommand{\cN}{\mathcal{N}}
\newcommand{\cA}{\mathcal{A}}
\newcommand{\cM}{\mathcal{M}}
\newcommand{\cR}{\mathcal{R}}
\newcommand{\cB}{\mathcal{B}}
\newcommand{\cG}{\mathcal{G}}
\newcommand{\cL}{\mathcal{L}}
\newcommand{\cH}{\mathcal{H}}
\newcommand{\cS}{\mathcal{S}}
\newcommand{\cT}{\mathcal{T}}
\newcommand{\cO}{\mathcal{O}}
\newcommand{\cC}{\mathcal{C}}
\newcommand{\cP}{\mathcal{P}}
\newcommand{\cE}{\mathcal{E}}
\newcommand{\cI}{\mathcal{I}}
\newcommand{\cF}{\mathcal{F}}
\newcommand{\cK}{\mathcal{K}}
\newcommand{\cV}{\mathcal{V}}
\newcommand{\cY}{\mathcal{Y}}
\newcommand{\cX}{\mathcal{X}}
\newcommand{\cZ}{\mathcal{Z}}
\def\bgamma{\boldsymbol\gamma}

\newcommand{\specialcell}[2][c]{%
  \begin{tabular}[#1]{@{}c@{}}#2\end{tabular}}

\newcommand{\figref}[1]{\Fig~\ref{#1}}
\newcommand{\secref}[1]{Section~\ref{#1}}
\newcommand{\algref}[1]{Algorithm~\ref{#1}}
\newcommand{\eqnref}[1]{Eq.~\ref{#1}}
\newcommand{\tabref}[1]{Table~\ref{#1}}

\newcommand{\rulesep}{\unskip\ \vrule\ }

\newcommand{\KLD}[2]{D_{\mathrm{KL}} \Big(#1 \mid\mid #2 \Big)}

\renewcommand{\b}{\ensuremath{\mathbf}}

\def\mc{\mathcal}
\def\mb{\mathbf}

\newcommand{\T}{^{\raisemath{-1pt}{\mathsf{T}}}}

\makeatletter
\DeclareRobustCommand\onedot{\futurelet\@let@token\@onedot}
\def\@onedot{\ifx\@let@token.\else.\null\fi\xspace}
\def\eg{{\em e.g}\onedot} \def\Eg{E.g\onedot}
\def\ie{{\em i.e}\onedot} \def\Ie{I.e\onedot}
\def\cf{cf\onedot} \def\Cf{Cf\onedot}
\def\etc{{\em etc}\onedot} \def\vs{vs\onedot}
\def\wrt{wrt\onedot}
\def\dof{d.o.f\onedot}
\def\etal{et~al\onedot} \def\iid{i.i.d\onedot}
\def\Fig{Fig\onedot} \def\Eqn{Eqn\onedot} \def\Sec{Sec\onedot} \def\Alg{Alg\onedot}
\makeatother

\newcommand{\xdownarrow}[1]{%
  {\left\downarrow\vbox to #1{}\right.\kern-\nulldelimiterspace}
}

\newcommand{\xuparrow}[1]{%
  {\left\uparrow\vbox to #1{}\right.\kern-\nulldelimiterspace}
}

\renewcommand\UrlFont{\color{blue}\rmfamily}

\newcommand*\rot{\rotatebox{90}}
\newcommand{\boldparagraph}[1]{\noindent{\bf #1:} }
\newcommand{\boldquestion}[1]{\noindent{\bf #1} }

\newcommand{\cmark}{\ding{51}}%
\newcommand{\xmark}{\ding{55}}%

\thesissignaturepage
\dedication{To those who supported me throughout this journey.}
\abstract{
A central goal in artificial intelligence is to enable machines to perceive the visual world as a composition of distinct objects. This ability for object-centric understanding is essential for generative models that support fine-grained, controllable content creation and editing. However, state-of-the-art diffusion models process images holistically and are conditioned on text, creating a semantic misalignment when tasked with object-level manipulation. As a result, researchers face a fundamental challenge: either adapt powerful but text-biased models or build specialized models from scratch, often with reduced capacity. This dissertation addresses this problem by introducing a framework that adapts pretrained generative models for object-centric image and video synthesis.

Our analysis highlights a core challenge in current approaches: achieving high-quality generation requires balancing global scene coherence with disentangled, object-level control. To address this, we propose an adaptation strategy that integrates object-specific conditioning into pretrained models while preserving their valuable priors. Extending this framework to video further amplifies the difficulty, as maintaining temporal coherence and consistent object identity across frames is critical.

For static images, we introduce SlotAdapt, a method that augments diffusion models with lightweight slot-based modules. A register token captures background and style, while slot-conditioned components encode object-specific information. This dual-pathway design mitigates text-conditioning bias and provides precise, object-centric control, leading to state-of-the-art results in object discovery, segmentation, compositional editing, and controllable image generation.

We then extend the framework to video. Using Invariant Slot Attention (ISA) to disentangle object identity from pose, combined with a Transformer-based temporal aggregator, our approach ensures consistent object representation and dynamics across time. This framework sets new benchmarks in unsupervised video object segmentation and reconstruction, while enabling advanced video editing capabilities, including object removal, replacement, and insertion, all without explicit supervision.

Overall, this work establishes a general and scalable approach to object-centric generative modeling for both images and videos. Beyond setting new technical baselines, it expands the design space for interactive and controllable generative tools, bridging the gap between human object-based perception and machine learning models. These contributions open new directions for structured, intuitive, and user-driven AI applications in creative, scientific, and practical domains.
}
\oz{
Yapay zek\=an\i n temel hedeflerinden biri, makinelerin g\"orsel d\"unyay\i\ ayr\i\ nesnelerden olu\c san bir b\"ut\"un olarak alg\i lamas\i n\i\ sa\u glamakt\i r. Nesne-merkezli anlay\i \c s, ince ayarl\i\ ve kontrol edilebilir i\c cerik \"uretimi ve d\"uzenlemesini destekleyen \"uretici modeller i\c cin vazge\c cilmezdir. Ancak g\"uncel dif\"uzyon modelleri g\"orselleri b\"ut\"unc\"ul olarak i\c sler ve metinlere ko\c sulland\i r\i l\i r; bu da nesne d\"uzeyindeki manip\"ulasyon g\"orevlerinde anlamsal bir uyumsuzluk yarat\i r. Bu nedenle ara\c st\i rmac\i lar, ya g\"ucl\"u fakat metin-\:on yarg\i l\i\ modelleri uyarlamak ya da kapasitesi s\i n\i rl\i\ yeni modeller geli\c stirmek gibi temel bir sorunla kar\c s\i\ kar\c s\i yad\i r. Bu tez, \"oncenden e\u gitilmi\c s \"uretici modelleri nesne-merkezli g\"or\"unt\"u ve video sentezi i\c cin uyarlayan bir \c cer\c ceve sunarak bu soruna \c c\"oz\"um getirmektedir.  

Analizimiz, mevcut yakla\c s\i mlarda temel bir zorluk ortaya koymaktad\i r: y\"uksek kaliteli \"uretim elde etmek i\c cin k\"uresel sahne tutarl\i l\i\u g\i n\i\ ayr\i \c st\i r\i lm\i \c s nesne d\"uzeyinde kontrol ile dengelemek gerekir. Bu dengeyi sa\u glamak i\c cin, de\u gerli \"onbilgileri korurken \"oncenden e\u gitilmi\c s modellere nesneye \"ozg\"u ko\c sulland\i rma entegre eden bir uyarlama stratejisi \"oneriyoruz. Bu \c cer\c cevenin videoya geni\c sletilmesi sorunu daha da zorla\c st\i rmaktad\i r; \c c\"unk\"u zamansal tutarl\i l\i k ve kareler boyunca nesne kimli\u ginin korunmas\i\ kritik \"onem ta\c s\i r.  

Dura\u gan g\"or\"unt\"uler i\c cin SlotAdapt y\"ontemini sunuyoruz. Bu y\"ontem, dif\"uzyon modellerini hafif slot-tabanl\i\ mod\"ullerle geni\c sletir. Bir kay\i t token'\i\ arka plan\i\ ve tarz\i\ yakalarken, slot-ko\c sullu bile\c senler nesneye \"ozg\"u bilgileri kodlar. Bu \c cift-yollu tasar\i m, metin ko\c sulland\i rma \"onyarg\i s\i n\i\ azalt\i r ve hassas, nesne-merkezli kontrol sa\u glar; nesne ke\c sfi, segmentasyon, bile\c simsel d\"uzenleme ve kontrol edilebilir g\"or\"unt\"u \"uretiminde alan\i n en iyi sonu\c clar\i n\i\ elde eder.  

Ard\i ndan \c cer\c cevemi zi videoya geni\c sletiyoruz. Invariant Slot Attention (ISA) kullanarak nesne kimli\u gini pozdan ayr\i\c st\i r\i yor, ayr\i ca Transformer-tabanl\i\ bir zamansal toplay\i c\i\ ile zaman boyunca nesne temsillerinin ve dinamiklerinin tutarl\i\ olmas\i n\i\ sa\u gl\i yoruz. Bu \c cer\c ceve, g\"ozetimsiz video nesne segmentasyonu ve yeniden yap\i land\i rmada yeni \"ol\c c\"utler belirlerken, nesne kald\i rma, de\u gi\c stirme ve ekleme gibi geli\c smi\c s video d\"uzenleme yeteneklerini de do\u rudan g\"ozetimsiz temsiller \"uzerinden m\"umk\"un k\i lmaktad\i r.  

Genel olarak, bu \c cal\i \c sma hem g\"or\"unt\"uler hem de videolar i\c cin nesne-merkezli \"uretici modelleme ad\i na genel ve \"ol\c ceklenebilir bir yakla\c s\i m ortaya koymaktad\i r. Yaln\i zca yeni teknik standartlar belirlemekle kalmay\i p, ayn\i\ zamanda etkile\c simli ve kontrol edilebilir \"uretici ara\c clar\i n tasar\i m alan\i n\i\ da geni\c sletmektedir. Bu katk\i lar, insan\i n nesne-merkezli alg\i s\i\ ile makine \"o\u grenmesi modelleri aras\i ndaki bo\c slu\u gu kapatarak, yarat\i c\i, bilimsel ve pratik alanlarda yap\i land\i r\i lm\i \c s, sezgisel ve kullan\i c\i\ odakl\i\ yapay zek\=a uygulamalar\i\ i\c cin yeni ufuklar a\c cmaktad\i r.  
 
}
\acknowledgments{{\small I would first like to express my deepest gratitude to my advisor, Prof. Yücel Yemez. At a time when I was without an advisor, you accepted me into your group and provided me with guidance, encouragement, and trust. Your support not only gave me the opportunity to continue my PhD but also shaped the direction of this work. It has been a privilege to learn from your insight and to carry out this research under your supervision.

I would also like to sincerely thank Assoc. Prof. Aykut Erdem and Assoc. Prof. Erkut Erdem, who served on my thesis monitoring committee. Their thoughtful feedback, constructive suggestions, and continued encouragement greatly improved the quality of this work. I am equally grateful to the jury members, Prof. Engin Erzin and Asst. Prof. Ayşegül Dündar, for their time, effort, and valuable evaluations of this dissertation.

I am grateful to my friends who made this journey more meaningful. I would like to especially thank Shadi Hamdan and Burak Can Biner for both their friendship and the many academic discussions we shared, which enriched my perspective and work. I am equally thankful to Batuhan Uluçesme for his support and friendship, and for the joy and encouragement he brought outside the academic setting.

I owe my deepest thanks to my family — my mother Dilek, my father Mehmet, and my sister Ayça — for their endless love, patience, and belief in me. Their constant support has been my anchor through the most difficult times, and their encouragement has given me the strength to persevere in this long journey. Everything I have accomplished is built upon the foundation of their sacrifices and unwavering trust.

Finally, my most profound gratitude goes to my fiancee, Aleyna Gürkan. Throughout this journey, she has been my closest companion, my source of strength, and my greatest joy. Her love, patience, and unshakable belief in me carried me through moments when I doubted myself, and her presence reminded me of what truly matters beyond the struggles of research. As we look forward to building our life together, I am deeply grateful to have shared this chapter with her.}}
\tableofcontents
\listoftables
\listoffigures
\abbreviations{
\begin{tabular}{lp{1cm}l}
	AI & & Artificial Intelligence\\
	CFG & & Classifier-Free Guidance\\
	DDPM & & Denoising Diffusion Probabilistic Models\\
	DiT & & Diffusion Transformers\\
	FG-ARI & & Foreground Adjusted Rand Index\\
	FID & & Fréchet Inception Distance\\
	FVD & & Fréchet Video Distance\\
	ISA & & Invariant Slot Attention\\
	KID & & Kernel Inception Distance\\
	LDM & & Latent Diffusion Model\\
	LLM & & Large Language Model\\
	LPIPS & & Learned Perceptual Image Patch Similarity\\
	mBO & & Mean Best Overlap\\
	mIoU & & Mean Intersection over Union\\
	MSE & & Mean Squared Error\\
	OCL & & Object-Centric Learning\\
	PSNR & & Peak Signal-to-Noise Ratio\\
	SSIM & & Structural Similarity Index\\
	VAE & & Variational Autoencoder\\
	ViT & & Vision Transformer\\
	VQ-VAE & & Vector Quantized-Variational Autoencoder\\
\end{tabular}
}
\textpages
\chapter{Introduction}
\label{chapter:introduction}

In this thesis we investigate how to equip generative models with object-centric representations that enable structured understanding of the visual world. Humans do not experience their surroundings as undifferentiated pixels but as a three-dimensional environment composed of distinct objects that persist, move, and interact. We develop computational methods to discover and represent these objects from visual data without human supervision. We first introduce a framework for unsupervised object discovery and compositional generation in static images. Then we extend this approach to video, where the challenges of identity preservation and temporal coherence arise. Our central argument is that object-centric, compositional representations provide the foundation for generative models that move beyond statistical pattern recognition toward structured and interpretable visual understanding.

\section{Compositionality in Dynamic Environments}
\label{sec:intro_compositionality}
The physical world is, by its nature, compositional. It is a structured system composed of individual entities that can be combined and arranged in arbitrarily many ways. This principle of compositionality is not merely an observation about the world, but a cornerstone of human intelligence itself. It allows for \textit{systematicity}, the capacity to understand and produce a vast number of novel combinations from a known set of components \cite{fodor1988connectionism}. This ability is so fundamental that from a young age, we as humans learn to parse complex scenes into discrete objects and their relationships, a process that enables core cognitive functions like causal reasoning, planning, and robust generalization to new situations \cite{lake2017building, scholkopf2021toward}. A key mechanism underlying this is the ability to create and maintain temporary representations of objects, often called ``object files," which allow humans to track their identity and properties through dynamic events \cite{kahneman1992reviewing}. It is this compositional, object-based view of the world that allows us to reason effectively within it.

This challenge is magnified when we consider the dimension of time. The physical environment is not static: objects move, interact, deform, and are frequently occluded from view. Humans possess an inherent and powerful ability to perceive object permanence and maintain a coherent understanding of a scene despite these continuous changes. This capacity is considered a form of ``core knowledge," present even in infants, suggesting it is a foundational element of how intelligent agents model their environment \cite{spelke2007core}. For an artificial system, however, this is an immense hurdle. The constant flux of pixel information caused by motion, changing illumination, and occlusions makes it incredibly difficult to recognize that an object at one moment is the same entity as an object at the next. Modeling this \textit{temporal compositionality}—understanding a scene as a collection of persistent objects undergoing transformations over time—is therefore a critical step toward building machines that perceive the world with human-like coherence \cite{ullman2017mind}.

For the field of artificial intelligence, bridging the gap between its current capabilities and this form of structured understanding is a primary objective. While deep learning models have achieved superhuman performance on a range of specific recognition tasks, their success is often based on learning statistical correlations in large datasets rather than acquiring a structured model of the world \cite{bottou2014from}. This can lead to brittle systems that fail to generalize systematically when faced with novel arrangements of known objects \cite{bahdanau2019systematic}. The prevailing hypothesis motivates this thesis: developing models with a built-in inductive bias for discovering and representing distinct objects is a necessary step toward more general and robust AI. By learning to see the world in terms of its compositional and dynamic parts, we can hope to build systems that not only recognize patterns but can also reason about, interact with, and generate faithful simulations of the world around them.

\section{Architectures and Capabilities of Generative Models}
\label{sec:intro_generative_models}

The primary ambition of generative modeling is to create an algorithm that can learn the underlying probability distribution of complex data, like natural images, and then synthesize novel, realistic samples from this learned knowledge. The last decade has seen this field transformed by several waves of innovation. The first was dominated by two major paradigms. \textit{Variational Autoencoders} (VAEs) introduced a principled probabilistic approach for learning a compressed latent representation of data via an encoder-decoder framework \cite{kingma2013auto}. In parallel, \textit{Generative Adversarial Networks} (GANs) proposed a novel two-player game between a generator network, which creates samples, and a discriminator network, which learns to distinguish real samples from fake ones \cite{goodfellow2014generative}. GANs became renowned for their ability to produce exceptionally sharp images and matured significantly over time. Architectures like StyleGAN demonstrated unprecedented control over the synthesis process by manipulating a hierarchical, style-based latent code, allowing for intuitive edits to pose, texture, and other attributes, though their adversarial training remained notoriously difficult to stabilize \cite{karras2019style}.

The next major turning point was driven by advances in Natural Language Processing (NLP). The introduction of the \textit{Transformer} architecture, with its core \textit{self-attention} mechanism, provided a highly scalable and parallelizable method for modeling long-range dependencies in sequential data \cite{vaswani2017attention}. This innovation rapidly led to the development of Large Language Models (LLMs). These models generally followed two main paths: autoencoding models like BERT used a masked language modeling objective to learn deep bidirectional representations for language understanding tasks \cite{devlin2019bert}. In contrast, autoregressive models from the GPT family demonstrated a remarkable ability to generate fluent, coherent text by simply predicting the next token in a sequence \cite{brown2020language}. The success of these LLMs established a powerful new paradigm: With a simple, scalable architecture and a self-supervised objective, a model trained on massive datasets could learn incredibly rich and useful representations of its domain.

Inspired by these advances, the vision community sought to bridge the gap between continuous images and the sequential nature of transformers. A key innovation in this area was the development of visual tokenization. The \textit{Vector Quantised-Variational Autoencoder} (VQ-VAE), and its powerful successor the \textit{VQ-GAN}, introduced a method to compress an image into a grid of discrete visual tokens from a learned codebook \cite{esser2021taming}. The VQ-GAN notably added a GAN-style discriminator to ensure that the decoded tokens resulted in high-fidelity reconstructions. This approach was famously exemplified by the original DALL-E model, which trained a large autoregressive transformer to generate these visual tokens conditioned on text prompts, effectively treating image generation as a language modeling problem \cite{ramesh2021zero}.

This provided the basis for the current state-of-the-art: \textit{diffusion models} \cite{sohl2015deep}. First proposed in 2015 but popularized years later, these models learn to reverse a gradual noising process, offering the stable training of VAEs with the high-quality output of mature GANs \cite{ho2020denoising}. Their true power was unlocked when conditioned on the rich semantic information provided by powerful, jointly-trained text-and-image encoders like CLIP \cite{radford2021learning}. The final piece of the puzzle came with \textit{Latent Diffusion Models} (LDMs), which made this entire process highly efficient by performing the denoising in the compressed latent space of a VAE \cite{rombach2022high}. This foundational LDM framework served as a launchpad for rapid scaling and refinement. Models like SDXL improved upon the original with a larger backbone and multiple text encoders for more nuanced prompt understanding \cite{podell2023sdxl}. A significant architectural evolution came with Diffusion Transformers (DiT), which replaced the conventional U-Net backbone with a pure Transformer, demonstrating its scalability and effectiveness for the denoising task \cite{peebles2023scalable}. This line of research has continued to advance at a rapid pace with the development of new training paradigms like Flow matching, which promise faster and more efficient generation \cite{liu2022flow}, and ever-larger models like FLUX, which employ novel hybrid transformer architectures to push the boundaries of quality and efficiency \cite{black2024flux}.

Naturally, the next step was to extend these powerful architectures to the dynamic domain of video. Recent text-to-video models like Imagen Video, Make-A-Video, and Phenaki have demonstrated the ability to generate short, coherent video clips from textual descriptions \cite{ho2022imagen, singer2022make, villegas2022phenaki}. These models typically adapt the successful image generation frameworks by incorporating temporal information, for instance, by using 3D convolutions or adding temporal attention layers into their backbones, allowing them to synthesize content consistent across space and time.

Across this entire history, from StyleGANs to the latest diffusion transformers, a consistent pattern can be observed. These systems have become exceptionally skilled at translating a holistic input, such as a text prompt, into a holistic output. However, their internal representation is not based on a disentangled, object-centric framework. They lack independent ``slots" or ``handles" for the individual entities in the scene. The consequence of this monolithic approach is a lack of fine-grained compositional control. The model cannot easily count objects, reason about their precise spatial relationships, or allow direct manipulation of one object without affecting the rest of the scene. While they can produce scenes containing objects, they do not reason \textit{about} or \textit{with} objects. As generative AI becomes increasingly integrated into creative and scientific workflows, this absence of an underlying compositional structure is no longer a theoretical concern but a practical bottleneck, creating an urgent need for models that offer more intuitive and granular control.

\section{Object-Centric Learning: Toward Structured Representation}
\label{sec:intro_ocl}

In response to the limitations of holistic models, a distinct research paradigm has emerged with the explicit goal of learning structured representations of the world: \textit{Unsupervised Object-Centric Learning} (OCL). The core motivation behind OCL is the belief that a crucial step towards more general and robust artificial intelligence involves building models that perceive, reason about, and interact with the world in terms of its constituent objects \cite{greff2020on, bottou2014from}. Instead of learning a flat mapping from pixels to labels, OCL seeks to discover a latent representation that explicitly factorizes a scene into entities, their properties, and their relationships. Such a representation is thought to be a prerequisite for more advanced capabilities like causal reasoning \cite{scholkopf2021toward}, systematic generalization \cite{bahdanau2019systematic}, and interpretable decision-making, moving beyond what is possible with purely monolithic systems.

The earliest modern approaches to this problem often employed an iterative inference framework. Models like Attend-Infer-Repeat (AIR) and its sequential video-based successor, SQAIR, learned to parse scenes in a sequential manner \cite{eslami2016attend, kosiorek2018sequential}. These models would typically use a recurrent network to scan a scene, attend to a specific location, infer the properties of a single object, and then subtract that object from the scene representation before repeating the process for the next object. While pioneering, these methods were often limited to simple synthetic environments with a small number of non-overlapping objects. They relied on patch-based decoders, which struggled to generate globally coherent scenes.

A breakthrough came with the introduction of \textit{Slot Attention}, a mechanism that enabled the parallel discovery of objects through a competitive attention process \cite{locatello2020object}. Instead of inferring objects sequentially, Slot Attention begins with a set of learnable vectors, or ``slots," which simultaneously compete to explain different parts of an image's feature map. Through several rounds of iterative refinement, each slot specializes to represent a distinct object or part of the scene, effectively decomposing the image representation into a collection of entities. This parallel approach proved far more scalable and robust, enabling successful object discovery in more complex synthetic scenes. These early Slot Attention models were often paired with a spatial broadcast decoder, reconstructing the final image by decoding each slot's representation and combining them via an alpha-compositing operation \cite{watters2019spatial}.

With the parallel discovery mechanism of Slot Attention established, subsequent research has focused on scaling OCL to the complexity of real-world images. One direction has been to equip the models with more powerful decoders, such as the transformer-based architectures used in SLATE and STEVE \cite{singh2021illiterate, singh2022simple}. Another line of inquiry has explored improvements to the object discovery and representation process itself. More recently, hierarchical approaches have been explored, such as COCA-Net \cite{kucuksozen2025cvpr}, which introduces a hierarchical clustering strategy with spatial broadcast decoding to achieve superior segmentation performance on synthetic datasets, and can also present results on realistic datasets that are not very complex. In parallel, some methods like DINOSAUR have bypassed the need for a generative decoder entirely, focusing purely on object discovery through self-supervised objectives. However, this sacrifices the ability to synthesize or edit images \cite{seitzer2023bridging}.

Extending these ideas to the video domain introduces the additional, major challenge of maintaining temporal consistency. Early methods in video OCL, like SAVi and SAVi++, often enforced this consistency by relying on external supervisory signals such as optical flow or depth information to guide the tracking of slots from one frame to the next \cite{kipf2021conditional, elsayed2022savi++}. More recently, self-supervised alternatives have been proposed to address the fragility of these external cues. Methods like TC-Slot and SOLV instead learn temporal consistency by enforcing similarity between slot representations across frames, either through contrastive learning or by clustering rich features from powerful pretrained vision models like DINO \cite{manasyan2024temporally, aydemir2023self}. While these models have achieved state-of-the-art results in unsupervised video object segmentation and tracking, they have a crucial limitation in common: they almost exclusively focus on the perception task and lack a pixel-level generative decoder. As a result, they cannot be used for photorealistic video synthesis or compositional editing tasks, leaving a critical gap between object discovery and content creation.

\section{Integrating Object-Centric Representations with Diffusion Models}
\label{sec:intro_juncture}

Given the respective strengths of the two fields, a logical and powerful next step for research was to combine the structured decomposition of Object-Centric Learning with the state-of-the-art synthesis quality of diffusion models. The aim was to use an OCL model as an encoder to discover objects, and a diffusion model as a powerful decoder to generate photorealistic images from these discovered object representations. However, the initial efforts at this integration revealed a core technical challenge, resulting in two distinct approaches, each ultimately limited in effectiveness.

The first line of research, illustrated by Latent Slot Diffusion (LSD), was to leverage a large, publicly available pretrained diffusion model as the slot decoder \cite{jiang2023object}. The motivation was to harness the immense generative power and world knowledge embedded in these models without the prohibitive cost of retraining them. This strategy, however, faced a significant challenge related to the models' conditioning mechanism. The cross-attention layers within these pretrained models were designed and optimized to receive text embeddings as their conditioning input. When fed object-slot representations instead, a ``text-conditioning bias" arose, creating a semantic misalignment between the learned slots and the expectations of the diffusion model's attention pathways. This often led to degraded generative quality, as the model struggled to interpret the non-textual slot information.

A contrasting strategy, taken by SlotDiffusion, sought to avoid this problem entirely by building a fully specialized system from the ground up \cite{wu2023slotdiffusion}. This required training the entire stack, the VAE for image compression, the U-Net diffusion model for the decoder, and the slot attention encoder, from scratch on the target dataset. While this eliminated the text-bias issue by design, it came with severe drawbacks. Training a diffusion model from scratch is not only a computationally massive undertaking, but more importantly, this approach sacrifices the powerful, general-purpose prior learned by large-scale pretrained models. The resulting generative capability is therefore limited, making the model less effective at handling the diversity and complexity of real-world images when compared to its large, pretrained counterparts.

This revealed a trade-off: On one hand, using pretrained models offered generative power but suffered from a conditioning mismatch, and On the other, training from scratch offered correct conditioning but lacked generative power. Therefore, a crucial question remained unanswered: How can one effectively adapt a pretrained diffusion model for object-centric generation, leveraging its powerful priors while simultaneously enabling it to understand and utilize conditioning information from object slots? Answering this question is one of the main goals of this thesis.

\section{Our Contributions}
\label{sec:intro_contributions}

To address the critical challenge of uniting object-centric learning with pretrained generative models, we introduce a novel framework that enables effective and controllable compositional synthesis. This framework is not merely an engineering solution; it directly provides generative models with the structural awareness they otherwise lack. By enabling models to represent and generate scenes in terms of their constituent objects, we shift generative AI from a tool of unconstrained synthesis to one of structured manipulation, opening new directions for fine-grained content creation and editing. Our contribution proceeds in two stages: first, we address the fundamental challenge of adapting pretrained models to static images; second, we extend this solution to the significantly more complex temporal domain of video.

\subsection{SlotAdapt: A Novel Adaptation for Compositional Image Generation}

The first significant contribution of this thesis is a method we term \textbf{SlotAdapt}, a new approach for object-centric learning that successfully integrates slot-based representations with pretrained diffusion models \cite{akan2025slot-guided}. Instead of retraining a diffusion model from scratch or forcing object slots into a text-centric conditioning framework, SlotAdapt \textit{adapts} a large-scale pretrained diffusion model. This is achieved through three key innovations, as explained below.

First, inspired by recent work on efficient fine-tuning \cite{mou2024t2i-adapter}, we introduce lightweight \textbf{adapter layers} into the frozen architecture of a pretrained diffusion model. These adapters are new cross-attention modules dedicated to receiving conditioning information from the learned object slots. This design allows the powerful, pretrained weights of the diffusion model to remain untouched. At the same time, the new adapter layers learn to map object-specific semantics to the generative process, effectively resolving the text-conditioning bias.

\begin{figure}[ht]
    \centering
    \includegraphics[width=0.8\linewidth]{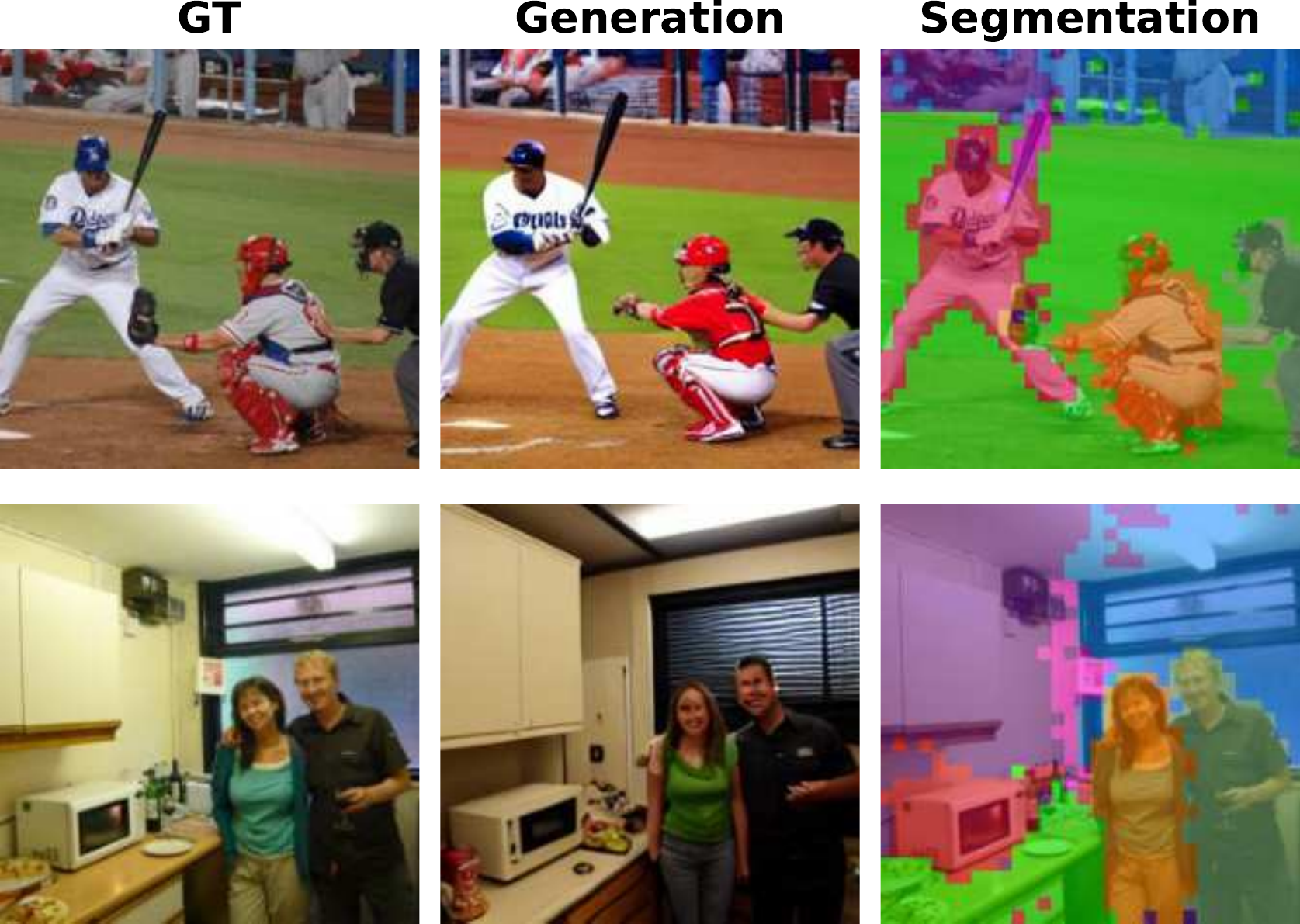}
    \caption[Unsupervised Object Discovery and Reconstruction]{
        \textbf{Unsupervised Object Discovery and Reconstruction.}
        From a single input image (a), our model autonomously discovers and separates the scene into distinct object-centric representations, visualized here as a segmentation mask (b). These learned object slots are rich enough to enable a high-fidelity reconstruction (c) of the original scene.
    }
    \label{fig:discovery_reconstruction}
\end{figure}

Second, to handle global scene context such as background and lighting, we introduce a \textbf{register token}. This token, created by pooling the learned slot representations, is fed into the diffusion model's original, text cross-attention layers. This strategy, which shares conceptual similarities with recent work on vision transformers \cite{darcet2024vision}, allows the model to store global information in a dedicated representation, freeing the individual slots to focus purely on their corresponding objects.

Third, we propose a self-supervised \textbf{attention guidance loss}. This objective enforces consistency between the attention masks produced by the Slot Attention encoder and the attention masks from our new adapter layers in the decoder. This mutual guidance helps the model resolve the part-whole hierarchy problem, learning to generate object masks that are better aligned with complete objects rather than fragmented parts, all without any external supervision.

Together, these innovations create a model that can effectively perceive and regenerate complex scenes. Figure \ref{fig:discovery_reconstruction} provides a qualitative example of SlotAdapt's capabilities. From a complex input image (a), our model autonomously discovers the constituent objects in the scene, producing a clean segmentation mask (b). The learned slots are sufficiently detailed to allow for a high-fidelity reconstruction of the original image (c), demonstrating a comprehensive understanding of the scene's structure and appearance.

\begin{figure}[ht]
    \centering
    \includegraphics[width=0.8\linewidth]{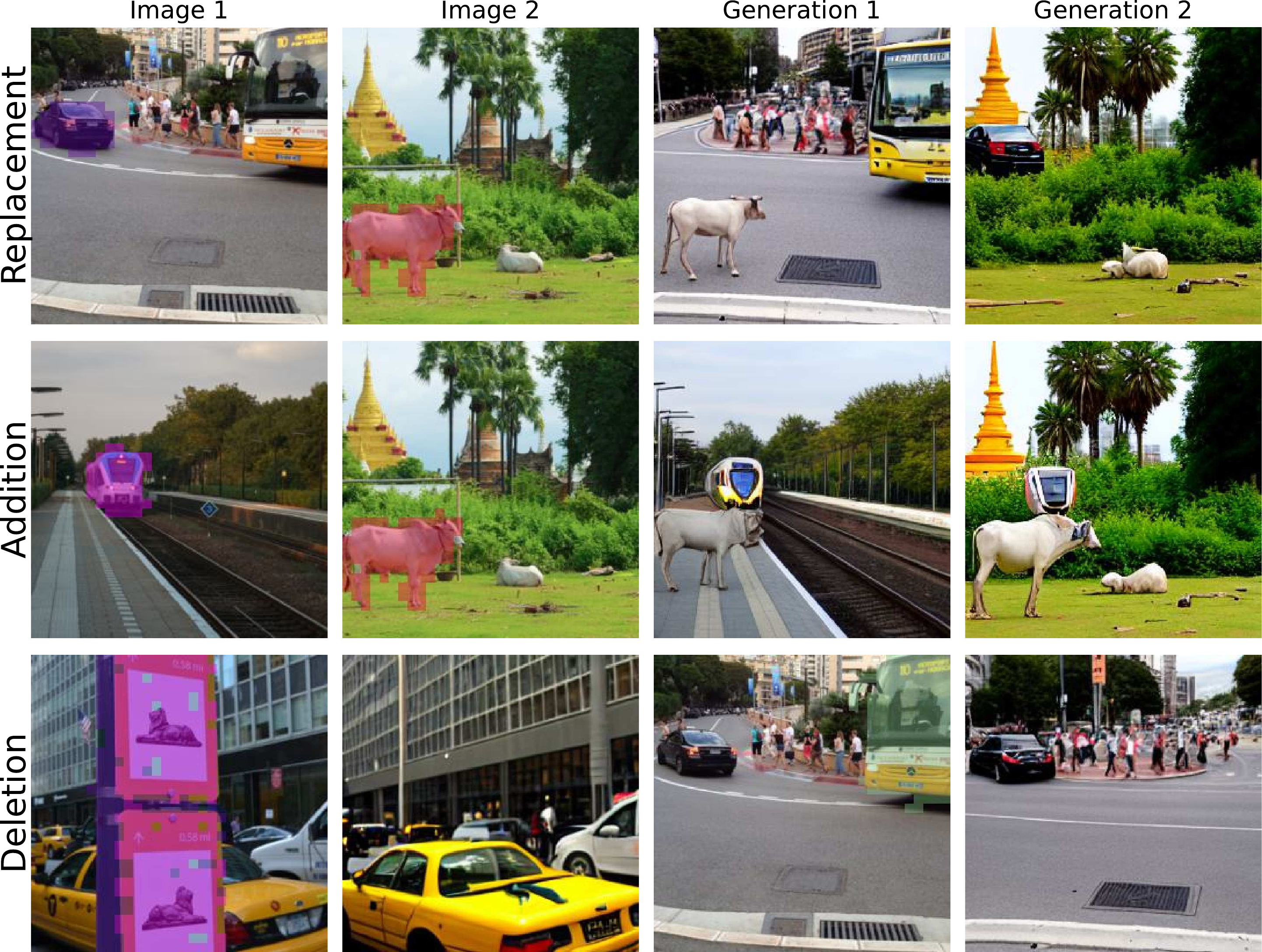}
    \caption[Compositional Image Editing via Slot Manipulation]{
        \textbf{Compositional Image Editing via Slot Manipulation.}
        Our object-centric framework enables direct, structured editing of scenes. By exchanging slots between different scenes, we can perform seamless \textbf{object replacement} (top row), by adding a slot from another image, we can perform \textbf{object addition} (middle row), by deactivating a learned object slot, we can perform targeted \textbf{object deletion} (bottom row), demonstrating true compositional control over the generative process.
    }
    \label{fig:compositional_editing}
\end{figure}

Through this framework, SlotAdapt not only achieves state-of-the-art performance in unsupervised object discovery but also unlocks new possibilities for content creation. The true power of its learned object-centric representation lies in the compositional control it offers. Because the model understands a scene as a collection of discrete, manipulable entities, we can perform structured edits by operating on the corresponding slots. As illustrated in Figure \ref{fig:compositional_editing}, this enables a range of powerful editing operations such as targeted object deletion, replacement, and addition. This thesis presents the first framework to demonstrate high-fidelity compositional editing from unsupervised representations on a challenging dataset like COCO, demonstrating that unsupervised slots can provide practical control for image generation \cite{akan2025slot-guided}.

\subsection{Extending Compositionality to the Temporal Domain}

Building on the success of SlotAdapt for static images, the second major contribution of this thesis is to extend this compositional framework to the video domain. This requires tackling the formidable challenges of maintaining object identity over time, modeling motion, and ensuring temporal coherence in the generated output. To this end, we present a novel framework for \textbf{compositional video synthesis} that learns temporally consistent object-centric representations \cite{akan2025compositional}.

Our approach introduces a \textbf{temporal object-centric encoder} that learns to discover and track objects across video frames. We use \textit{Invariant Slot Attention} (ISA) \cite{biza2023invariant} to extract object slots for each frame, helping to disentangle object identity from its pose and position. The core innovation is a \textbf{Transformer-based temporal aggregator}, which takes the sequence of per-frame slots as input and produces an enriched, temporally-aware representation for each object. These ``video slots" capture not only the appearance of an object but also its trajectory and context within the video clip.

We employ a highly efficient \textbf{1-frame training strategy} for the generative process. At each training step, the model's task is to reconstruct only a single, randomly selected video frame. However, it performs this reconstruction conditioned on the temporally-aware video slots aggregated from the \textit{entire} clip. This clever design allows us to leverage the power of a large, pretrained \textit{image} diffusion model as our decoder, teaching it to understand video dynamics without needing to train a full video diffusion model from scratch.

This method achieves state-of-the-art results in object-centric video synthesis, producing high-fidelity, temporally coherent reconstructions. Crucially, it pioneers a new capability in the field: direct and flexible \textbf{compositional video generation and editing}. Existing video editing methods usually rely on textual guidance or post-hoc fine-tuning. In contrast, our framework enables editing directly through unsupervised object representations, allowing objects to be added, removed, or replaced while preserving temporal coherence. This extension treats video not as a sequence of pixels but as a set of persistent objects. It enables structured editing with applications in visual effects, animation, and synthetic data generation~\cite{akan2025compositional}.

\subsection{Summary of Innovations}

The primary contributions of this thesis are organized into two main research thrusts, each building upon the last, which can be summarized by the following innovations:

\begin{itemize}
    \item We propose \textbf{SlotAdapt}, a novel and efficient method for adapting large, pretrained text-to-image diffusion models for object-centric image generation. Its key innovations include:
    \begin{itemize}
        \item The introduction of dedicated \textbf{adapter layers} for slot-based conditioning, resolving the semantic misalignment and text-conditioning bias present in the original pretrained models.
        \item A \textbf{register token} mechanism to disentangle global scene information from object-specific slot representations, improving the model's focus on individual object properties.
        \item A self-supervised \textbf{attention guidance loss} that improves object segmentation quality by enforcing consistency between the encoder's slot attention and the decoder's adapter attention.
        \item The first demonstration of high-fidelity \textbf{compositional image editing} from unsupervised object-centric representations on a complex, real-world dataset.
    \end{itemize}

    \item We extend this framework to the temporal domain with a novel method for \textbf{compositional video synthesis}. This second contribution introduces:
    \begin{itemize}
        \item A \textbf{Transformer-based temporal aggregator} designed to learn coherent object representations that maintain their identity and track motion across video frames.
        \item A novel framework that, for the first time, enables direct and flexible \textbf{compositional video editing}—such as object removal, replacement, and addition—by manipulating learned, unsupervised video object slots.
    \end{itemize}
\end{itemize}

\section{Publications}
\label{sec:intro_publications}

The core contributions of this thesis have led to the following publications:

\begin{itemize}
    \item Adil Kaan Akan and Yucel Yemez. ``Slot-Guided Adaptation of Pre-trained Diffusion Models for Object-Centric Learning and Compositional Generation." \textit{Proceedings of the International Conference on Learning Representations (ICLR)}, 2025.

    \item Adil Kaan Akan and Yucel Yemez. ``Compositional Video Synthesis by Temporal Object-Centric Learning." \textit{Submitted to IEEE Transactions on Pattern Analysis and Machine Intelligence (TPAMI)}.
\end{itemize}

\section{Outline of the Thesis}
\label{sec:intro_outline}

The remainder of this thesis is organized as follows:

\textbf{Chapter~\ref{chapter:background}} provides the necessary background for this work. It offers a detailed technical review of the foundational concepts, including the mathematical formulations of the Slot Attention mechanism and the denoising diffusion probabilistic model framework.

\textbf{Chapter~\ref{chap:related_work}} presents a comprehensive literature review of the fields relevant to this thesis. This includes a survey of unsupervised object-centric learning for both images and video, a review of how diffusion models have been integrated into object-centric frameworks, and an overview of existing compositional image and video editing methods.

\textbf{Chapter~\ref{chap:slotadapt}} details the first significant contribution of this thesis: the SlotAdapt framework. We describe its architecture, including the novel adapter layers and register token mechanism, and present a thorough experimental evaluation of its performance on object discovery and compositional image generation tasks.

\textbf{Chapter~\ref{ch:video}} details the second main contribution: extending our framework to the temporal domain. We explain the architecture for compositional video synthesis, focusing on the temporal aggregator and training strategy. We then present extensive experiments for both unsupervised video object segmentation and compositional video generation and editing.

\textbf{Chapter~\ref{chap:conc}} concludes the thesis. It summarizes the key findings and contributions, discusses the limitations of the current work, and suggests promising directions for future research in object-centric generative modeling.

\newpage
\chapter{Technical Background}
\label{chapter:background}

In this chapter, we introduce the technical foundations on which the contributions of this thesis are built. Our discussion centers on two interrelated areas: object-centric representation learning, which provides a structured way of parsing visual scenes into entities, and diffusion-based generative modeling, which has emerged as the state of the art for high-fidelity synthesis. Together, these frameworks establish the basis for generative models that can not only produce realistic images and videos but also represent them as compositions of distinct objects.

We first review the mechanisms that enable object-centric models to discover compositional structure in visual data. We begin with the limitations of early sequential inference methods, which motivated the development of the Slot Attention mechanism for parallel object discovery~\cite{locatello2020object}. We then describe the Spatial Broadcast Decoder~\cite{watters2019spatial}, commonly paired with Slot Attention, and discuss its shortcomings, which highlight the need for more powerful generative decoders. The section concludes with Invariant Slot Attention~\cite{biza2023invariant}, an extension designed to disentangle object identity from pose.

In the second part of the chapter, we cover generative modeling with diffusion processes. We introduce the Denoising Diffusion Probabilistic Model (DDPM)~\cite{ho2020denoising} framework and explain its reparameterized training objective. Building on this, we review Latent Diffusion Models (LDMs)~\cite{rombach2022high}, which perform diffusion in a compressed representation space, and discuss their cross-attention conditioning mechanism. This analysis sets the stage for the adaptation strategies developed in subsequent chapters, where diffusion models are integrated with object-centric representations.

\section{Object-Centric Representation Learning}
Unsupervised Object-Centric Learning (OCL) is a paradigm focused on developing models that can autonomously segment a visual scene into a set of discrete, meaningful, and manipulable representations, often termed ``slots'' \cite{greff2020on}. This approach moves beyond holistic feature extraction, explicitly seeking to decompose a scene into its constituent parts. The motivation is to equip models with an inductive bias that mirrors a fundamental aspect of human perception: viewing the world not as a flat canvas of pixels, but as a composition of distinct entities.

\subsection{Slot Attention Mechanism}
\label{sec:slot_attention_background}
Early approaches to OCL, such as Attend-Infer-Repeat (AIR) \cite{eslami2016attend}, often employed an iterative, sequential inference framework. While pioneering, these methods were computationally demanding and struggled to scale to scenes with more than a few simple objects. The primary motivation for the \textbf{Slot Attention} mechanism, proposed by Locatello et al. \cite{locatello2020object}, was to overcome this bottleneck by enabling the \textit{parallel} discovery of objects through a competitive attention process.

The mechanism aims to decompose a feature map $\mathbf{f} \in \mathbb{R}^{M \times D_f}$, extracted from an image, into a set of $K$ slot vectors $\mathbf{S} \in \mathbb{R}^{K \times D_s}$. The process is an iterative refinement, which we detail below.

\paragraph{Initialization.}
The process begins with a set of $K$ slot vectors, $\mathbf{S}^{(0)} \in \mathbb{R}^{K \times D_s}$, which are randomly initialized, typically from a standard normal distribution, $\mathcal{N}(0, 1)$. These vectors have no prior knowledge of the scene.

\paragraph{Iterative Refinement.}
For a fixed number of iterations ($m=1, \dots, R$), the slots compete for responsibility over the input pixels and are updated by information from the input features.

\textbf{1. Projections for Attention:} At each iteration $m$, the current slots $\mathbf{S}^{(m-1)}$ and the input features $\mathbf{f}$ are linearly projected to create queries ($\mathbf{Q}$), keys ($\mathbf{K}$), and values ($\mathbf{V}$). This is a standard step in attention mechanisms to map the inputs to a space where similarity can be computed.
\begin{ceqn}
\begin{align}
    \mathbf{Q} &= \text{LayerNorm}(\mathbf{S}^{(m-1)}) \cdot \mathbf{W}_q \\
    \mathbf{K} &= \text{LayerNorm}(\mathbf{f}) \cdot \mathbf{W}_k \\
    \mathbf{V} &= \text{LayerNorm}(\mathbf{f}) \cdot \mathbf{W}_v
\end{align}
\end{ceqn}
where $\mathbf{W}_q \in \mathbb{R}^{D_s \times D_k}$, $\mathbf{W}_k \in \mathbb{R}^{D_f \times D_k}$, and $\mathbf{W}_v \in \mathbb{R}^{D_f \times D_s}$ are learnable projection matrices.

\textbf{2. Attention Computation and Competition:} The core of the mechanism lies in how attention is computed. First, raw similarity scores, or attention logits, are calculated as the dot product between keys and queries, and then scaled by:
\begin{ceqn}\begin{equation}
    \mathbf{A}_{\text{logits}} = \mathbf{K} \mathbf{Q}^T
\end{equation}\end{ceqn}
\begin{ceqn}\begin{equation}
    \mathbf{A}_{\text{scaled}} = \frac{\mathbf{A}_{\text{logits}}}{\sqrt{D_k}}
\end{equation}\end{ceqn}
The scaling factor stabilizes training by preventing dot products from saturating the softmax. Next, the attention weights $\mathbf{W} \in \mathbb{R}^{M \times K}$ are obtained by applying the softmax function along the dimension of the slots (i.e., the columns of $\mathbf{A}_{\text{scaled}}$).
\begin{ceqn}
\begin{equation}
    \mathbf{W} = \text{softmax}(\mathbf{A}_{\text{scaled}}, \text{dim=slots})
\end{equation}
\end{ceqn}
For each input feature $f_i$, the softmax ensures that the attention weights assigned to it across all slots, $\sum_{j=1}^K W_{i,j}$, sum up to 1. This forces the slots to compete for ``ownership'' of each feature, encouraging specialization.

\textbf{3. Information Aggregation and GRU Update:} The slots are then updated by the input. First, an update vector $\mathbf{U}^{(m)} \in \mathbb{R}^{K \times D_s}$ is computed via a weighted sum over the values, where the weights are the transposed attention matrix, which aggregates the relevant visual information for each slot.
\begin{ceqn}
\begin{equation}
    \mathbf{U}^{(m)} = \mathbf{W}^T \mathbf{V}
\end{equation}
\end{ceqn}
To incorporate this new information, a Gated Recurrent Unit (GRU) cell is employed. The final slot representations are updated according to:
\begin{ceqn}
\begin{equation}
    \mathbf{S}^{(m)} = \text{GRU}(\mathbf{S}^{(m-1)}, \mathbf{U}^{(m)})
\end{equation}
\end{ceqn}
The use of GRU is critical for the stability of the refinement process, acting as a learned gate that selectively incorporates new information, which prevents noisy or drastic updates to the slot representations between iterations. After $R$ iterations, the final slot vectors $\mathbf{S}^{(R)}$ represent a disentangled, object-centric summary of the scene.

\subsection{Spatial Broadcast Decoder}
Once a scene is decomposed into a set of object slots, a decoder is required to reconstruct the image from these representations. The standard decoder used in early Slot Attention-based models was the Spatial Broadcast Decoder \cite{watters2019spatial}. The motivation was to provide a simple, structured, and parameter-efficient way to map non-spatial slot vectors back to a spatial, pixel-based output.

The decoder operates on each slot individually. For a single slot vector $\mathbf{s}_j \in \mathbb{R}^{D_s}$, the decoding process is as follows:
\begin{enumerate}
    \item \textbf{Broadcast}: The slot vector $\mathbf{s}_j$ is tiled, or ``broadcast,'' across a spatial grid of the desired output dimensions, for instance, $H \times W$, creating a feature map of shape $H \times W \times D_s$.
    \item \textbf{Coordinate Concatenation}: Two additional channels represent a normalized coordinate grid. These channels contain each pixel's $x$ and $y$ coordinates, typically scaled to the range $[-1, 1]$. This grid is added to the broadcast slot tensor to provide explicit spatial information to the subsequent layers.
    \item \textbf{Convolutional Refinement}: The resulting tensor is passed through a simple convolutional network (e.g., 4 to 6 layers) to produce a reconstructed image patch $\mathbf{x}_j$ and a corresponding alpha mask $\mathbf{m}_j$ for the object represented by slot $\mathbf{s}_j$.
\end{enumerate}
The final image reconstruction $\mathbf{x}_{\text{recon}}$ is then formed by combining all the individually reconstructed objects via alpha compositing:
\begin{ceqn}
\begin{equation}
    \mathbf{x}_{\text{recon}} = \sum_{j=1}^{K} \mathbf{m}_j \cdot \mathbf{x}_j
\end{equation}
\end{ceqn}

To ensure a coherent final image, the individual alpha masks $\mathbf{m}_j$ are typically generated as unnormalized logits and then normalized across the slot dimension (e.g., via a softmax operation). This guarantees that the compositing weights at each pixel sum to one, preventing artifacts from overlapping masks and ensuring a properly exposed reconstruction.

While effective for simple synthetic scenes, the Spatial Broadcast Decoder has a significant limitation: it frequently struggles to generate high-fidelity, textured images, often yielding blurry or overly smooth reconstructions. This is because the simple convolutional network lacks the generative capacity to model the complex appearance of real-world objects. This capability gap is the primary motivation for the core work of this thesis: replacing this simple decoder with a powerful, pretrained diffusion model to achieve state-of-the-art compositional generation.

\subsection{Invariant Slot Attention (ISA)}
A limitation of the standard Slot Attention mechanism is that it tends to entangle an object's identity with its pose (i.e., its position, scale, and rotation). This is problematic for robust object-centric reasoning. The motivation behind \textbf{Invariant Slot Attention (ISA)} \cite{biza2023invariant} is to learn a truly disentangled representation by explicitly separating object identity from spatial properties. 

ISA achieves this by introducing a canonical object coordinate frame for each slot. In addition to a vector $\mathbf{z}^j \in \mathbb{R}^{D_z}$, each slot $\mathbf{s}_j$ is associated with a scale parameter $\mathbf{s}_s^j \in \mathbb{R}^2$ and a position parameter $\mathbf{s}_p^j \in \mathbb{R}^2$. Given an absolute coordinate grid $\mathbf{G}_{abs} \in \mathbb{R}^{M \times 2}$ providing the spatial position of each of the $M$ input feature patches, ISA first computes a relative coordinate grid for each slot:
\begin{ceqn}
\begin{equation}
    \mathbf{G}_{rel}^{j} = \frac{\mathbf{G}_{abs} - \mathbf{s}_p^j}{\mathbf{s}_s^j}
\end{equation}
\end{ceqn}
This transformation normalizes the spatial features with respect to the slot's current estimated pose, making the subsequent attention computation invariant to that pose. The attention mechanism is then augmented as follows, detailed for a single slot $j$:

\textbf{1. Augmented Key Projection:} The keys are formed not just from the visual features $\mathbf{f}$, but also from a learnable projection $g(\cdot)$ of the relative spatial features. This injects pose-invariant spatial information directly into the attention process.
\begin{ceqn}
\begin{equation}
    \mathbf{K}_{\text{augmented}} = p(k(\mathbf{f}) + g(\mathbf{G}_{rel}^{j}))
\end{equation}
\end{ceqn}
where $p, k, g$ are learnable linear projections. The queries are projected from the identity vectors, $\mathbf{q}_j = q(\mathbf{s}_j)$.

\textbf{2. Attention Score Calculation:} The attention logits (raw scores) $\mathbf{M}_j$ for slot $j$ are computed as the scaled dot product between the augmented keys and the slot's query vector:
\begin{ceqn}
\begin{align}
    \bA_{logits} &= \mathbf{K}_{\text{augmented}} \mathbf{q}_j^T \\
    \bA_{\text{scaled}} &= \frac{\bA_{\text{logits}}}{\sqrt{d}} 
\end{align}
\end{ceqn}
The final attention weights, $\mathbf{W} \in \mathbb{R}^{M \times K}$, are then obtained by applying the softmax function to these scores.
\begin{ceqn}
\begin{equation}
    \mathbf{W} = \text{softmax}(\bA_{\text{scaled}}, \text{dim}=\text{slots})
\end{equation}
\end{ceqn}
Crucially, the slot's scale and position parameters are updated iteratively based on this attention mask, computed as the weighted mean and standard deviation of the absolute coordinates \cite{biza2023invariant}. The slot vector $\mathbf{s}_j$ is then updated via a GRU, but the information it receives is now conditioned on the object's local coordinate frame, similar to the original Slot Attention.

\paragraph{Decoding with Pose Information.}
The ISA mechanism produces in an slot vector $\mathbf{s}_j$ largely invariant to pose. While this is ideal for representation learning, this slot vector alone is insufficient for reconstructing the object with a decoder like the Spatial Broadcast Decoder, as the pose information has been factored out. To reconstruct the object, pose information must be reintroduced. This is done by adding the projected relative grid, $\mathbf{G}_{rel}$, at decoding time, ensuring that slots remain pose-invariant during learning.

\begin{ceqn}
\begin{align}
    \textbf{Z}_{\text{decode}} &= \text{SpatialBroadcast}(\textbf{S}) \\
    \textbf{Z}_{\text{decode}} &= \textbf{Z}_{\text{decode}} + h(\mathbf{G}_{rel})
\end{align}
\end{ceqn}
where $h(\cdot)$ is a learnable projection of the relative grid tensor. This complete representation, now containing both the object's identity and its specific pose in the scene, is then passed to the decoder with convolutional layers. This ensures the object is reconstructed at the correct position and scale. This process allows ISA to learn a disentangled representation where $\mathbf{s}_j$ encodes the invariant identity (``what''), while $\mathbf{s}_s^j$ and $\mathbf{s}_p^j$ encode its transient spatial properties (``how big'' and ``where'').

\section{Generative Modeling with Diffusion Models}
In recent years, the landscape of generative modeling has been reshaped by the widespread success of diffusion models. These models have become the state-of-the-art paradigm for high-quality synthesis tasks, valued for their ability to generate diverse and photorealistic images while offering stable training dynamics, which is a notable advantage over adversarial approaches such as GANs~\cite{dhariwal2021diffusion}. At their core, all diffusion models are based on a fundamental principle: systematically disrupt structure in data through a fixed noising process, and then learn to carefully reverse that process to generate new data from pure noise. In this section, we provide a detailed technical overview of the foundational DDPM framework and its computationally efficient successor, the LDM~\cite{rombach2022high}.

\subsection{Denoising Diffusion Probabilistic Models (DDPM)}
\label{sec:ddpm_background}
The foundational framework for diffusion models is the Denoising Diffusion Probabilistic Model (DDPM), first proposed by Sohl-Dickstein et al. \cite{sohl2015deep} and later popularized and simplified by Ho et al. \cite{ho2020denoising}. DDPM consists of two complementary processes as explained below.

\paragraph{Forward Process (Diffusion).}
The forward process, denoted by $q$, is a fixed, non-learned Markov chain that gradually injects Gaussian noise into an initial data point $\mathbf{x}_0$ over a sequence of $T$ discrete timesteps. The transition at each step $t$ is defined as a Gaussian distribution where the mean is a slightly contracted version of the previous state, and a small amount of variance is added:
\begin{ceqn}
\begin{equation}
    q(\mathbf{x}_{t} | \mathbf{x}_{t-1}) = \mathcal{N}(\mathbf{x}_{t}; \sqrt{1 - \beta_{t}}\mathbf{x}_{t-1}, \beta_{t}\mathbf{I})
\end{equation}
\end{ceqn}
where $\{\beta_{t}\}_{t=1}^{T}$ is a predefined variance schedule of small positive constants that typically increase with $t$. The joint distribution of the entire noise trajectory given the starting image is:
\begin{ceqn}\begin{equation}
    q(\mathbf{x}_{1:T} | \mathbf{x}_0) = \prod_{t=1}^T q(\mathbf{x}_t | \mathbf{x}_{t-1})
\end{equation}\end{ceqn}
An important property of this process is that due to the nature of Gaussian distributions, we can sample the noisy image $\mathbf{x}_t$ at any arbitrary timestep $t$ directly from the initial image $\mathbf{x}_0$ in a single, closed-form step:
\begin{ceqn}
\begin{equation}
    \mathbf{x}_{t} = \sqrt{\bar{\alpha}_{t}}\mathbf{x}_0 + \sqrt{1 - \bar{\alpha}_{t}}\mathbf{\epsilon}
    \label{eq:ddpm_forward_detailed}
\end{equation}
\end{ceqn}
where $\alpha_{t} = 1 - \beta_{t}$, $\bar{\alpha}_{t} = \prod_{i=1}^{t} \alpha_i$, and $\mathbf{\epsilon} \sim \mathcal{N}(0, \mathbf{I})$. As $t \to T$, the cumulative product $\bar{\alpha}_{T}$ approaches zero, meaning the signal from $\mathbf{x}_0$ vanishes and $\mathbf{x}_T$ becomes indistinguishable from pure Gaussian noise.

\paragraph{Reverse Process (Denoising).}
The generative part of the model is the reverse process, $p_\theta$, which learns to reverse the diffusion, starting from pure noise $\mathbf{x}_T \sim \mathcal{N}(0, \mathbf{I})$ and producing a clean sample. This process is also a Markov chain:
\begin{ceqn}
\begin{equation}
    p_\theta(\mathbf{x}_{0:T}) = p(\mathbf{x}_T) \prod_{t=1}^{T} p_\theta(\mathbf{x}_{t-1} | \mathbf{x}_t)
\end{equation}
\end{ceqn}
where each reverse transition $p_\theta(\mathbf{x}_{t-1} | \mathbf{x}_t) = \mathcal{N}(\mathbf{x}_{t-1}; \boldsymbol{\mu}_\theta(\mathbf{x}_t, t), \boldsymbol{\Sigma}_\theta(\mathbf{x}_t, t))$ is parameterized by a neural network with parameters $\theta$. The theoretical motivation for training this network comes from optimizing the Variational Lower Bound (VLB) on the data log-likelihood. 

However, the full VLB objective is complex. Ho et al. \cite{ho2020denoising} made a crucial observation that allows for a much simpler training objective. The key insight is that if we condition the true reverse posterior $q(\mathbf{x}_{t-1} | \mathbf{x}_t, \mathbf{x}_0)$ on the initial image $\mathbf{x}_0$, it becomes tractable and is also a Gaussian, with a mean that can be expressed as:
\begin{ceqn}
\begin{equation}
    \tilde{\boldsymbol{\mu}}_t(\mathbf{x}_t, \mathbf{x}_0) = \frac{\sqrt{\bar{\alpha}_{t-1}}\beta_t}{1-\bar{\alpha}_t}\mathbf{x}_0 + \frac{\sqrt{\alpha_t}(1-\bar{\alpha}_{t-1})}{1-\bar{\alpha}_t}\mathbf{x}_t
\end{equation}
\end{ceqn}
While $\mathbf{x}_0$ is unknown during inference, we can use Equation \ref{eq:ddpm_forward_detailed} to express it in terms of $\mathbf{x}_t$ and the noise $\mathbf{\epsilon}$ that was added to create it. By substituting this expression for $\mathbf{x}_0$ into the equation for $\tilde{\boldsymbol{\mu}}_t$, we can reparameterize the mean of the learned reverse distribution $\boldsymbol{\mu}_\theta$ in terms of a function $\mathbf{\epsilon}_\theta$ that predicts the noise from the noisy image $\mathbf{x}_t$:
\begin{ceqn}
\begin{equation}
    \boldsymbol{\mu}_\theta(\mathbf{x}_t, t) = \frac{1}{\sqrt{\alpha_t}}\left(\mathbf{x}_t - \frac{\beta_t}{\sqrt{1-\bar{\alpha}_t}}\mathbf{\epsilon}_\theta(\mathbf{x}_t, t)\right)
\end{equation}
\end{ceqn}
This reparameterization shows that training a network to predict the noise $\mathbf{\epsilon}$ is mathematically equivalent to learning the mean of the reverse distribution $\boldsymbol{\mu}_\theta$. While the variance $\boldsymbol{\Sigma}_\theta$ can also be learned, it is often fixed to a constant, as learning the mean is critical for sample quality. This reparameterization allows for a much simpler and more stable training objective, which is a re-weighted variant of the VLB that focuses solely on the noise prediction error:
\begin{ceqn}
\begin{equation}
    \mathcal{L}_{\text{simple}} = \mathbb{E}_{\mathbf{x}_0 \sim p(\mathbf{x}), {\mathbf{\epsilon}}_t\sim \mathcal{N}(0,1), t \sim \mathcal{U}(1, T)} \left[ ||\mathbf{\epsilon} - \mathbf{\epsilon}_{\theta}(\sqrt{\bar{\alpha}_{t}}\mathbf{x}_0 + \sqrt{1 - \bar{\alpha}_{t}}\mathbf{\epsilon}, t)||_2^2 \right]
    \label{eq:ddpm_loss_detailed}
\end{equation}
\end{ceqn}
where $\mathbf{\epsilon}_{\theta}$ is typically a U-Net architecture. This simplified objective has been shown empirically to produce higher-quality samples than directly optimizing the full VLB.

\subsection{Latent Diffusion Models (LDM) and Cross-Attention Conditioning}
The primary motivation for moving from DDPMs to \textbf{Latent Diffusion Models (LDMs)} \cite{rombach2022high} was to address the immense computational and memory bottleneck of operating in the high-dimensional pixel space. For the high-resolution images common in real-world datasets like COCO, running a large U-Net over the full pixel grid for hundreds of sampling steps is often prohibitively expensive. LDMs solve this problem by performing the diffusion process in a much smaller, compressed latent space.

\paragraph{Architecture and Training.}
The LDM framework consists of two main components:
\begin{enumerate}
    \item \textbf{A Perceptual Compression Model:} A powerful, pretrained Variational Autoencoder (VAE) is used, where encoder, $\mathcal{E}$, maps a high-resolution image $\mathbf{x}$ to a spatially smaller latent representation, $\mathbf{z}_0 = \mathcal{E}(\mathbf{x})$, and the decoder, $\mathcal{D}$, maps the latent back to the pixel space, $\tilde{\mathbf{x}} = \mathcal{D}(\mathbf{z}_0)$.
    \item \textbf{A Latent Diffusion Model:} The DDPM, as described above, is trained entirely in the latent space, where U-Net, $\mathbf{\epsilon}_\theta$, now learns to denoise noisy latents $\mathbf{z}_t$ instead of noisy images $\mathbf{x}_t$.
\end{enumerate}
The VAE is pretrained once, and its weights are frozen during the training of the diffusion model. This decouples perceptual compression from generative learning, allowing the U-Net to focus its capacity on the semantic composition of the data in a much more computationally tractable space. The LDM training objective, including a conditioning term $\mathbf{y}$, is a direct application of Equation \ref{eq:ddpm_loss_detailed} in the latent space:
\begin{ceqn}
\begin{equation}
    \mathcal{L}_{\text{LDM}} = \mathbb{E}_{\mathbf{z} \sim \mathcal{E}(\mathbf{x}), {\mathbf{\epsilon}}_t\sim \mathcal{N}(0,1), t \sim \mathcal{U}(1, T)} \left[ ||\mathbf{\epsilon} - \mathbf{\epsilon}_{\theta}(\mathbf{z}_{t}, t, \mathbf{y})||_2^2 \right]
    \label{eq:ldm_loss_detailed}
\end{equation}
\end{ceqn}

\paragraph{Conditional Generation via Cross-Attention.}
A main motivation for the widespread adoption of LDMs is their ability to be guided by external information, enabling controllable synthesis. This control is achieved via a flexible \textbf{cross-attention} mechanism integrated into the U-Net denoiser.

Let the external conditioning signal (e.g., a text prompt) be denoted by $\mathbf{y}$. This signal is first mapped to an intermediate embedding space by a dedicated domain-specific encoder, $\tau_\theta(\mathbf{y})$. For text-to-image models like Stable Diffusion, $\tau_\theta$ is typically a pretrained CLIP text encoder. The output, $\mathbf{c} = \tau_\theta(\mathbf{y})$, is a sequence of embedding vectors.

This conditioning representation $\mathbf{c}$ is then injected into the U-Net at multiple layers, typically within its residual and transformer blocks. The mechanism works as follows: the intermediate spatial feature maps of the U-Net at a given layer act as the queries ($\mathbf{Q}$), while the conditioning representation $\mathbf{c}$ is linearly projected to provide the keys ($\mathbf{K}$) and values ($\mathbf{V}$). The cross-attention operation is then given by:
\begin{ceqn}
\begin{equation}
    \text{Attention}(\mathbf{Q}, \mathbf{K}, \mathbf{V}) = \text{softmax}\left(\frac{\mathbf{Q}\mathbf{K}^T}{\sqrt{D_k}}\right)\mathbf{V}
\end{equation}
\end{ceqn}
This allows the U-Net to ``pay attention'' to different parts of the conditioning signal (e.g., different words) when generating different image parts. However, this powerful mechanism introduces a critical challenge that motivates the core contributions of this thesis. The cross-attention layers in widely-used pretrained LDMs are optimized for and expect conditioning vectors that reside in a \textit{textual} embedding space produced by a specific encoder like CLIP. This creates a significant ``text-conditioning bias,'' leading to a semantic misalignment when attempting to condition them on non-textual information, such as the object-centric slot representations detailed previously. Resolving this conflict between the structured representations of OCL and the text-biased conditioning of pretrained diffusion models is one of the challenges that this thesis addresses.
\newpage
\chapter{Related Work}
\label{chap:related_work}

In this chapter, we place the contributions of this thesis within the context of current research in computer vision and generative modeling. We first establish the foundational work in learning structured, object-centric representations, tracing its development from static images to dynamic videos. Next, we review the concurrent advances in generative modeling, focusing on the rise of diffusion models. Finally, we examine the critical intersection of these two fields, detailing the specific challenges this thesis addresses. This culminates in a discussion of compositional generation and editing, a key capability enabled by our work.

\section{Unsupervised Object-Centric Learning (OCL)}
\label{sec:related_ocl}

The ability to perceive the world as a collection of distinct objects is fundamental to human intelligence. Unsupervised Object-Centric Learning (OCL) is a research area that aims to give machines this same ability. The goal is to create models that can automatically break down a visual scene into a set of meaningful parts that correspond to objects, without needing human-provided labels or supervision \cite{greff2020on}. This section reviews the progress of OCL, from its beginnings with static images to its application to the more complex domain of video.

\subsection{Foundations in Image-Based OCL}
\label{ssec:related_ocl_image}

Early works in OCL often used a sequential, iterative inference process to identify objects. For example, the Attend-Infer-Repeat (AIR) model \cite{eslami2016attend} used a recurrent network to scan a scene, attend to a specific location to infer the properties of a single object, and then subtract that object from the scene representation before repeating the process. This approach was extended to video in Sequential AIR (SQAIR) \cite{kosiorek2018sequential}. While these methods were important first steps, their sequential nature made them slow and difficult to apply to scenes with many overlapping objects.

The introduction of \textbf{Slot Attention} \cite{locatello2020object} marked a shift toward parallel object discovery, overcoming the scalability limits of earlier sequential approaches. Slot Attention initializes a set of learnable vectors (“slots”) that compete through attention to bind with different regions of an image’s feature map. After iterative updates, each slot specializes to a particular object or scene component. This parallel process was more efficient and robust, making it possible to work with more complex scenes. Concurrent works like MONet \cite{burgess2019monet}, IODINE \cite{greff2019multi}, and SPACE \cite{lin2019space} also explored iterative variational inference and spatial attention mechanisms to decompose scenes, further establishing the effectiveness of this general approach.

Early models used a \textbf{Spatial Broadcast Decoder} \cite{watters2019spatial} to reconstruct an image from these learned slots. This decoder takes each non-spatial slot vector, tiles it across a spatial grid, adds coordinate information to reintroduce a spatial bias, and then uses a small convolutional network to generate an image of the object and a corresponding transparency (alpha) mask. The final image is then created by layering all the reconstructed objects. The main problem with this decoder was its limited generative power; it often produced blurry or overly simple reconstructions that could not capture the detailed appearance of real-world objects. This limitation was a key reason for investigating more powerful generative models as decoders, which is a central topic of this thesis.

Later research pursued two main strategies to scale OCL for complex real-world images. The first strategy focused on developing more powerful decoders. SLATE \cite{singh2021illiterate} and STEVE \cite{singh2022simple}, for example, replaced convolutional decoders with slot-conditioned Transformers. These models first tokenize an image using a VQ-VAE and then autoregressively predict the token sequence, reframing generation as a language modeling task. Methods like SPOT further refined this line of work \cite{kakogeorgiou2024spot} introducing new training techniques and hierarchical representations. Furthermore, COCA-Net \cite{kucuksozen2025cvpr} replaces the slot attention with an hierarchical clustering based attention module.

The second, alternative strategy was to bypass the generative decoder entirely. Methods like DINOSAUR \cite{seitzer2023bridging} operate directly on the semantically rich features of a pretrained DINO model, discovering objects through direct feature clustering. While highly effective for the perception task of segmentation, this approach inherently sacrifices the ability to generate or edit images—a key capability this thesis aims to preserve and enhance.

\subsection{The Temporal Challenge: OCL for Video}
\label{ssec:related_ocl_video}

Applying OCL to video adds the significant challenge of temporal consistency. A model must not only find objects in one frame but also track their identities as they move, change appearance, or become hidden by other objects. This requires a mechanism for temporal binding, ensuring that the representation for a specific object remains stable across frames.

Initial video OCL models like SAVi \cite{kipf2021conditional} and SAVi++ \cite{elsayed2022savi++} often use external information to help with tracking. They used a predictor-corrector system where slot representations from a previous frame were used to initialize or predict slots in the current frame. These predictions were refined or corrected using signals like optical flow or depth maps, which provide explicit motion information. However, these external signals can be unreliable in real-world videos, especially with fast motion, non-rigid deformations, or occlusions, making the models less robust.

To address these issues, recent work has focused on learning temporal consistency from the video data itself, without external cues. These methods build architectural priors or learning objectives that encourage consistency. For example, SlotFormer \cite{wu2023slotformer} uses a slot-aligned cross-frame attention mechanism to model object dynamics and predict future states. Other methods like TC-Slot \cite{manasyan2024temporally} use contrastive learning to ensure that the slot representation for a given object remains similar across frames. Another approach, seen in Betrayed-by-Attention \cite{ding2024betrayed}, introduces a combination of hierarchical clustering and a consistency objective to stabilize slot attention over time. RIV takes an alternative perspective \cite{qian2024rethinking}, which rethinks image-to-video adaptation from an object-centric viewpoint. Another method in this area is SOLV \cite{aydemir2023self}, which clusters features from a powerful DINO vision transformer across time to obtain consistent object tracks. While these models are very effective for unsupervised video object segmentation, they generally do not have pixel-level generative decoders. As a result, they cannot be used to create or edit photorealistic videos, a critical gap this thesis aims to fill.

A key technique for robust tracking is to separate an object's identity from its specific pose (position and scale). \textbf{Invariant Slot Attention (ISA)} \cite{biza2023invariant} was designed for this purpose. It modifies the Slot Attention mechanism to estimate the pose of each object explicitly. This allows the model to learn a representation for each object's identity independent of its location or size in the frame. This ability to disentangle ``what" an object is from ``where" it is forms a critical part of the temporal model developed in this thesis.

Beyond benchmarks, object-centric representations have also proven valuable in applied domains such as autonomous driving. Prior work in trajectory and world modeling, including StretchBEV \cite{stretchbev}, FTGN \cite{ftgn}, and Adapt \cite{adapt}, demonstrated the need for structured and disentangled representations of agents and environments. More recent world model approaches, such as GAIA-1 and GAIA-2 \cite{gaia1,gaia2}, and FIERY \cite{fiery}, further highlight that object- and scene-centric modeling is essential for robust prediction and decision making. These applications further motivate the pursuit of temporally consistent object-centric learning frameworks.

\section{The Generative Revolution: A Diffusion Model Primer}
\label{sec:related_diffusion}

In parallel to the developments in OCL, the field of generative modeling has been transformed by the success of diffusion models. These models emerged as the state-of-the-art paradigm, replacing earlier approaches like Variational Autoencoders (VAEs) \cite{kingma2013auto} and Generative Adversarial Networks (GANs) \cite{goodfellow2014generative}. While GANs, particularly advanced architectures like StyleGAN \cite{karras2019style}, could produce exceptionally sharp images, they were often difficult to train. Diffusion models offered a new path, providing comparable or superior visual quality with the benefit of stable and reliable training \cite{dhariwal2021diffusion}.

\subsection{High-Fidelity Image Synthesis}
\label{ssec:related_diffusion_image}

Building on the technical foundations introduced in Chapter~2, diffusion models have rapidly established themselves as the leading paradigm for high-fidelity image synthesis. A central reason for this success is their ability to combine strong generative performance with stable and reliable training, in contrast to earlier approaches such as VAEs~\cite{kingma2013auto} and GANs~\cite{goodfellow2014generative}. This section reviews key advances in the diffusion-based literature that have driven improvements in image quality, scalability, and controllability.

Early work demonstrated that diffusion models could match or exceed the visual fidelity of GANs. Dhariwal and Nichol~\cite{dhariwal2021diffusion} showed that classifier guidance and architectural refinements allowed diffusion models to achieve state-of-the-art performance on standard image benchmarks. Subsequent research emphasized efficiency and scalability. Latent Diffusion Models (LDMs)~\cite{rombach2022high} introduced perceptual compression through a pretrained VAE, reducing the computational cost of denoising while enabling synthesis at high resolution. This formulation provided the foundation for widely used systems such as Stable Diffusion.

A major line of work has focused on conditional generation. Cross-attention mechanisms allow diffusion models to integrate external signals such as text, layout, or semantic maps, enabling controllable synthesis at scale. This direction led to breakthroughs in text-to-image generation, including Imagen~\cite{saharia2022photorealistic} and Stable Diffusion~\cite{rombach2022high}, which demonstrated unprecedented photorealism and diversity guided by natural language descriptions. Larger-scale successors such as SDXL~\cite{podell2023sdxl} extended this framework with deeper backbones and multiple encoders for more nuanced conditioning.

Architectural innovations have further expanded the capacity of diffusion models. Diffusion Transformers (DiT)~\cite{peebles2023scalable} replaced the conventional U-Net with a pure Transformer backbone, highlighting the scalability of self-attention for modeling long-range dependencies in visual data. Hybrid approaches such as SD3.5~\cite{esser2024sd35} and FLUX~\cite{black2024flux} build on this trend by combining convolutional and transformer components to balance efficiency with expressivity.

Finally, efficiency has become a central research focus. Approaches such as Rectified Flows~\cite{liu2022flow}, consistency models~\cite{song2023consistency}, and distillation techniques~\cite{salimans2022progressive} aim to reduce the number of denoising steps required, significantly accelerating generation while maintaining high visual fidelity. These developments illustrate the pace of progress and establish diffusion models as the dominant framework for controllable, high-quality image synthesis.

\subsection{Extending Synthesis to Video}
\label{ssec:related_diffusion_video}

Video synthesis has been studied under a variety of formulations, with one of the earliest problems being \emph{future frame prediction}. In this setting, the task is to extrapolate the temporal evolution of a scene from an observed context. Models such as PredNet \cite{lotter2016prednet}, Stochastic Video Generation (SVG) \cite{denton2018stochastic}, SLAMP \cite{slamp}, SLAMP3D \cite{slamp3d}, and related work \cite{thesis} attempted to directly predict future frames in both autonomous driving and general video domains. While these approaches were important first steps, they often struggled to maintain long-term temporal consistency and object permanence.

The success of image diffusion models naturally led to their extension to video generation, addressing many of these limitations. Foundational work like Video Diffusion Models (VDMs) \cite{ho2022video} demonstrated that the denoising framework could be applied to sequential data. Subsequent large-scale models such as Imagen Video \cite{ho2022imagen}, Make-A-Video \cite{singer2022make}, and Phenaki \cite{villegas2022phenaki} showcased the ability to generate short, high-definition, and coherent video clips from text prompts.

These models typically adapt image generation architectures by incorporating a temporal dimension. This is often achieved by modifying the U-Net architecture to include temporal attention layers or by using 3D convolutions instead of 2D ones. Such temporal modules allow the model to correlate information across frames, ensuring that the generated video remains consistent over time. While these models have achieved impressive synthesis quality, a common characteristic is that they treat the video as a single, holistic output. They do not maintain an explicit, factorized representation of the objects within the scene. This makes fine-grained control—such as editing a single object without affecting the rest of the video—particularly challenging. Recent efforts have introduced control mechanisms through keyframes, masks, or motion guidance, as seen in works like Tune-A-Video \cite{wu2023tune} and ControlVideo \cite{zhang2023controlvideo}. However, these methods remain orthogonal to an object-centric approach and do not operate on learned, disentangled object representations.

\section{The Synthesis of Structure and Photorealism: OCL Meets Diffusion}
\label{sec:related_intersection}

Given the complementary strengths of OCL (structured representation) and diffusion models (photorealistic synthesis), combining them is a logical and powerful research direction. The goal is to build a hybrid model where an OCL encoder discovers objects and a diffusion decoder generates high-fidelity images conditioned on these object representations. However, early attempts to create such a model revealed a fundamental technical challenge: whether to initialize components from pretrained models or to train them entirely from scratch.

\subsection{Challenges in Conditioning Pretrained Models}
\label{ssec:related_dilemma}

The first path, taken by methods like Latent Slot Diffusion (LSD) \cite{jiang2023object} and GLASS \cite{singh2024guided}, was to use a large, publicly available pretrained diffusion model (e.g., Stable Diffusion) as the decoder. The motivation was to leverage the immense generative power and world knowledge encoded in these models without the prohibitive cost of retraining them. This approach, however, encountered a significant obstacle: a \textbf{text-conditioning bias}. The cross-attention layers in these pretrained models were designed and optimized to receive text embeddings from a specific encoder like CLIP. When fed object-slot representations instead, a semantic misalignment occurred. The diffusion model struggled to interpret this out-of-domain conditioning, often resulting in degraded generative quality and a failure to ground the generation in the provided slots properly.

The second path, taken by SlotDiffusion \cite{wu2023slotdiffusion}, was to avoid this bias by building and training the entire system from the ground up on a target dataset. This involved training the slot encoder, the VAE, and the diffusion U-Net together. While this approach ensures that the conditioning mechanism is perfectly aligned with the slot representations by design, it has severe drawbacks. Training a diffusion model from scratch is a computationally massive undertaking. More importantly, this approach forgoes the powerful, general-purpose prior learned by large-scale pretrained models. Therefore, the resulting generative capacity is limited, making the model less effective at handling the diversity and complexity of real-world images.

This challenge raised a central research question: how can pretrained diffusion models be adapted to utilize slot-based conditioning while effectively retaining their powerful generative priors? In this thesis, we address this question with the \textbf{SlotAdapt} framework \cite{akan2025slot-guided}. Instead of forcing slots into a text-centric framework or training from scratch, SlotAdapt introduces a novel adaptation strategy, which inserts lightweight \textbf{adapter layers} \cite{mou2024t2i-adapter} into the frozen diffusion model, new cross-attention modules dedicated solely to receiving slot conditioning. SlotAdapt introduces a \textbf{register token} \cite{darcet2024vision} fed to the original text-conditioning layers to handle global scene information. This design resolves the conditioning dilemma, successfully combining the power of pretrained models with the structured control of OCL. The specific architecture and training strategy of the SlotAdapt framework, designed to resolve this dilemma, will be detailed in Chapter~\ref{chap:slotadapt}.

\section{Compositional Generation and Editing}
\label{sec:related_compositional}

A primary motivation for developing object-centric generative models is to enable compositional control over the synthesis process. If a model represents a scene as a collection of discrete entities, it should be possible to edit the scene by manipulating these entities.

\subsection{Compositional Control in Images}
\label{ssec:related_compositional_image}

The ability to perform structured edits like adding, removing, or replacing objects in an image is a key benchmark for compositional understanding. The framework developed in this thesis, SlotAdapt, is the first to demonstrate this capability with high fidelity on complex, real-world datasets like COCO using unsupervised object representations \cite{akan2025slot-guided}. By operating directly on the learned object slots, our method allows for zero-shot compositional editing. For instance, an object can be removed from a scene by deactivating its corresponding slot, or an object from one image can be placed into another by transferring its slot, demonstrating accurate compositional control over the generative process.

\subsection{Compositional Control in Video Generation}
\label{ssec:related_compositional_video}

While compositional editing for images is a significant achievement, extending this capability to video is a far more complex frontier. It requires manipulating an object's appearance and ensuring that its motion and interactions remain coherent over time. At the time of this work, there is a notable gap in the literature for methods that allow for compositional video editing (e.g., object replacement, deletion) by directly manipulating learned, unsupervised object representations. The second significant contribution of this thesis, a novel framework for compositional video synthesis \cite{akan2025compositional}, directly addresses this gap. By learning temporally consistent video slots, our method is the first to enable such structured editing in the video domain, treating a video not as an immutable sequence of pixels, but as a dynamic and manipulable collection of persistent objects. Chapter~\ref{ch:video} will present the whole methodology of our temporal framework that enables this novel capability.

\section{Thesis Contributions}
\label{sec:related_summary}

This review of the literature has highlighted two critical challenges at the intersection of object-centric learning and generative modeling. First, the field faced a dilemma in static images: pretrained diffusion models suffered from conditioning mismatch, while training from scratch limited generative capacity. Second, in video, methods were either strong at unsupervised segmentation or at holistic generation, but not both, and none supported compositional editing with learned object representations.

This thesis addresses both challenges by introducing solutions for image-based conditioning and temporal compositional generation.
\begin{enumerate}
    \item For images, we propose \textbf{SlotAdapt}, a framework that resolves the conditioning dilemma by efficiently adapting a pretrained diffusion model for object-centric control. This enables state-of-the-art object discovery and, for the first time, high-fidelity compositional editing on complex real-world scenes.
    \item For video, we introduce a framework for \textbf{compositional video synthesis} that extends these principles to the temporal domain. This is the first work to unify high-performance unsupervised video object segmentation with high-quality generative modeling, enabling compositional editing by manipulating learned, temporally-aware object slots.
\end{enumerate}
\newpage
\chapter[Adapting Pretrained Diffusion Models for Object-Centric Learning]{SlotAdapt: Adapting Pretrained Diffusion Models for Object-Centric Learning}
\label{chap:slotadapt}

\section{Introduction}
\label{sec:slotadapt_intro}

As established in Chapter 3, integrating object-centric learning with state-of-the-art generative models presents a significant challenge. This chapter introduces \textbf{SlotAdapt}, a novel framework designed to resolve this challenge through a principled and efficient adaptation strategy.

The central hypothesis of this work is that large pretrained diffusion models can be leveraged for object-centric tasks without the need for expensive retraining. SlotAdapt adapts the frozen model to incorporate conditioning information from object slots. This is achieved through three key innovations that operate together:
\begin{enumerate}
    \item \textbf{Adapter-based conditioning} introduces new, lightweight pathways for object-specific information to guide the generative process. The adapters avoid misalignment between slot representations and text-trained attention pathways by bypassing the model's original text-conditioning layers for object semantics.
    \item A \textbf{register token} mechanism disentangles global scene context from object-specific representations. This allows the model to handle background, lighting, and style information separately from foreground objects, making individual slots more focused and effective.
    \item A \textbf{self-supervised attention guidance} loss enforces consistency between the encoder's perception of objects and the decoder's generation of them. This improves segmentation quality and helps resolve the part-whole hierarchy problem without external labels, guiding slots to bind to whole objects rather than fragments.
\end{enumerate}
This chapter provides a detailed technical exposition of SlotAdapt, from its object-centric encoder to the adapted diffusion decoder. We then present a comprehensive experimental evaluation, demonstrating how this approach not only overcomes the limitations of prior work but also unlocks new capabilities in high-fidelity compositional generation and editing on complex, real-world datasets.

\section{The SlotAdapt Architecture}
\label{sec:slotadapt_arch}

The SlotAdapt framework is designed to integrate a slot-based object encoder with a pretrained latent diffusion model decoder in an effective and efficient way. The central principle is to modify the decoder just enough to make it compatible with object-centric representations, while keeping the vast majority of its powerful, pretrained weights frozen. This preserves the rich generative prior learned by the base model while enabling fine-grained, object-level control.

\subsection{Overall Pipeline}
\label{ssec:slotadapt_pipeline}

\begin{figure}[h!]
    \centering
    \includegraphics[width=\textwidth]{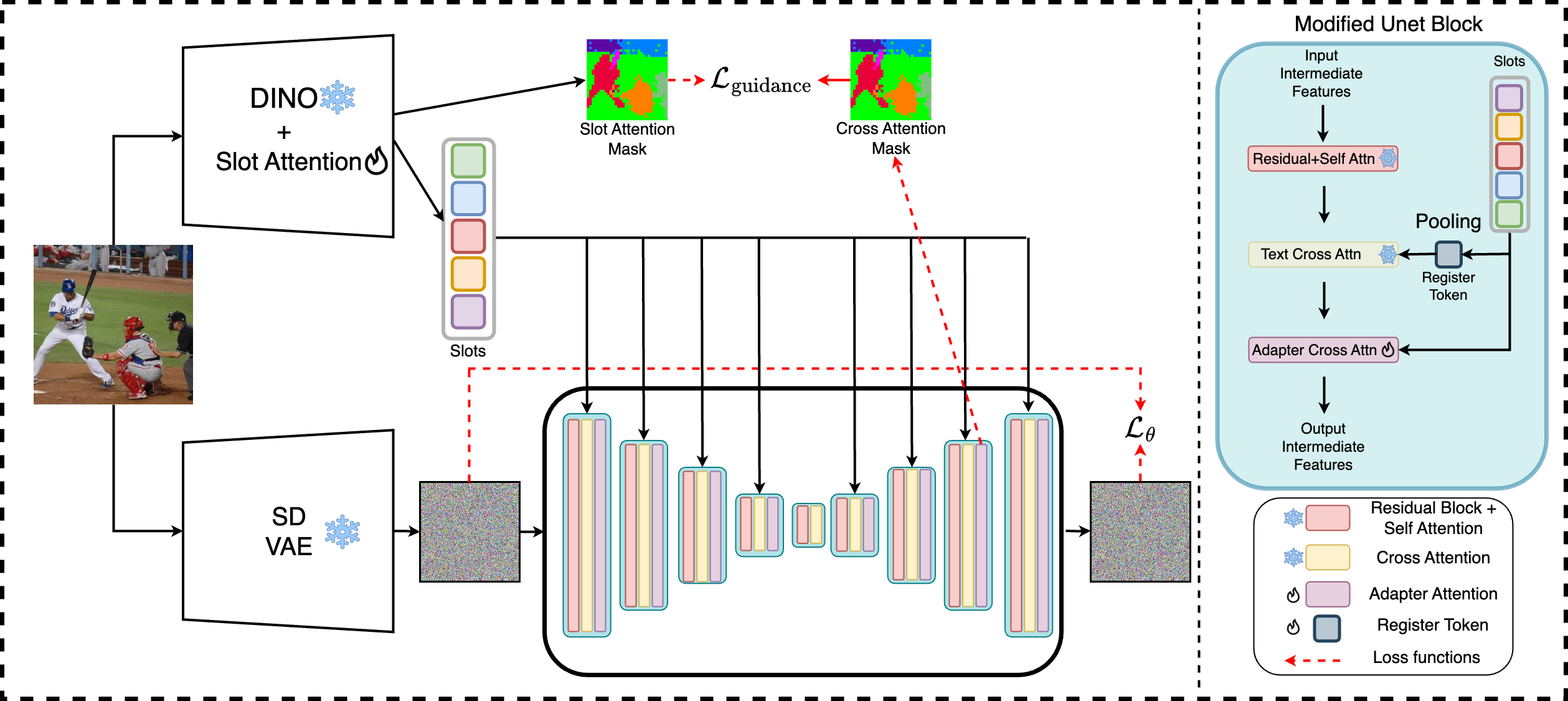} 
    \caption[The SlotAdapt Architecture]{The SlotAdapt Architecture. An input image is processed by a DINOv2 and Slot Attention encoder to produce object slots. These slots, and a pooled register token, condition a modified Stable Diffusion U-Net. New adapter layers (purple) receive slot conditioning, while the original cross-attention layers (yellow) receive the register token. A guidance loss aligns the slot attention masks with the adapter cross-attention masks to improve segmentation.}
    \label{fig:slotadapt_architecture}
\end{figure}

The end-to-end pipeline of SlotAdapt is illustrated in Figure \ref{fig:slotadapt_architecture}. The process begins with an input image fed into a two-stage object-centric encoder. First, a frozen visual backbone extracts a rich, patch-based feature map. Second, a trainable Slot Attention module processes this map to produce a set of $K$ object-centric slot vectors. These slots and a derived register token are then used to condition the generative decoder. The decoder, a pretrained Latent Diffusion Model with frozen weights, uses these representations to denoise a latent vector and reconstruct the original image.

The entire model is trained end-to-end by minimizing a reconstruction objective, augmented by the attention guidance loss. Critically, the only trainable components are the Slot Attention module and the lightweight adapter layers added to the decoder, making the training process more efficient than retraining a complete diffusion model.

\subsection{Object-Centric Encoder}
\label{ssec:slotadapt_encoder}

The encoder decomposes the input image into a set of structured, object-centric representations. This process has two stages.

First, we use a powerful, pretrained vision transformer, specifically DINOv2 \cite{oquab2023dinov2}, as a frozen visual backbone. Given an input image $\mathbf{x} \in \mathbb{R}^{H \times W \times 3}$, the backbone acts as a feature extractor, producing a spatially smaller grid of feature vectors $\mathbf{f} \in \mathbb{R}^{h \cdot w \times d}$. DINOv2, when used as a backbone, due to its self-supervised training objective, encourages the emergence of highly semantic features that correspond well to object boundaries, making it an ideal foundation for unsupervised object discovery~\cite{seitzer2023bridging}.

Second, the standard Slot Attention mechanism \cite{locatello2020object} is applied to these feature vectors. As introduced in Section~\ref{sec:slot_attention_background}, this module uses a competitive, iterative attention process (typically three iterations) to dynamically bind a set of $K$ learnable vectors, or "slots," to distinct objects or entities in the scene. The output of this stage is a set of slot representations $\bS \in \mathbb{R}^{K \times d}$, where each slot vector ideally contains the semantic and appearance information for a single object in the input image.

\subsection{Slot-Conditioned Diffusion Decoder}
\label{ssec:slotadapt_decoder}

The core innovation of SlotAdapt lies in its decoder, a pretrained Stable Diffusion U-Net model adapted for slot-based conditioning. The key idea is to introduce new pathways for the slot information while leaving the original, powerful weights of the model untouched. This is achieved through a dual-conditioning strategy that separates object-level information from global scene context.

\subsubsection{Adapter-Based Conditioning for Object Semantics}
To circumvent the text-conditioning bias of the original model, we do not use its existing cross-attention layers for the object slots. Instead, inspired by recent work in efficient model adaptation \cite{mou2024t2i-adapter}, we inject new, lightweight cross-attention layers, which we call \textbf{adapters}, into the frozen U-Net. Specifically, an adapter layer is inserted after each existing cross-attention layer in all downsampling and upsampling blocks of the U-Net. These adapter layers are dedicated exclusively to receiving the learned object slots $\bS$ as their conditioning context (key and value inputs), while the U-Net's spatial features serve as the query.

During training, only the weights of these small adapter layers are updated. This allows them to learn the mapping between object-slot semantics and the internal spatial representations of the diffusion model without interfering with or corrupting its vast pretrained knowledge. This minimal, targeted modification enables slots to focus on object semantics without interference from the text-conditioning pathway.

\subsubsection{The Register Token for Global Context}
While the adapters handle object-specific information, the model still needs to represent global scene properties like background, overall lighting, and artistic style. Therefore, we introduce a register token $\br$, a concept similar to that explored in vision transformers \cite{darcet2024vision}, to handle this global context explicitly. This single token is created by computing the mean of the generated slot vectors $\bS$, effectively creating a summary representation of the entire scene's content.

This register token is then fed as input to the U-Net's \textit{original}, text-trained cross-attention layers. This strategy effectively repurposes the powerful, existing conditioning mechanism to handle global information. By delegating the global context to this dedicated token, we allow the individual slots and their corresponding adapter layers to concentrate solely on representing and generating their specific objects. This disentanglement is important for robust object discovery and enables clean compositional editing. The denoising loss we optimize for reconstruction is:

\begin{ceqn}
\begin{equation}
\mathcal{L}_{\theta} = \Vert \mathbf{\mathbf{\epsilon}}_t - \mathbf{\epsilon}_{\btheta}(\bx_t, t, \bS, \br) \Vert_{2}^{2}
\end{equation}
\end{ceqn}

\subsection{Self-Supervised Attention Guidance}
\label{ssec:slotadapt_guidance}

A common challenge in OCL is the part-whole hierarchy problem, where slots may bind to salient parts of an object rather than the whole entity~\cite{hinton1979some}. To address this without external supervision, we introduce a self-supervisory signal by enforcing consistency between the encoder's and decoder's attention mechanisms.

Formally, the Slot Attention encoder produces an attention mask $\bA_{\text{\tiny SA}}$. At the same time, the cross-attention adapters in the diffusion model decoder yield a corresponding attention mask $\bA_{\text{\tiny DM}}$:

\begin{ceqn}
\begin{equation}
\label{eq:attention_mask}
\bA_{\text{\tiny SA}} = \mathrm{Softmax}\!\left(\frac{k_{\text{\tiny SA}}(\bff_{\text{\tiny SA}})\, q_{\text{\tiny SA}}(\bS)^\top}{\sqrt{D}}\right),
\quad
\bA_{\text{\tiny DM}} = \mathrm{Softmax}\!\left(\frac{k_{\text{\tiny DM}}(\bS)\, q_{\text{\tiny DM}}(\bff_{\text{\tiny DM}})^\top}{\sqrt{D}}\right)
\end{equation}
\end{ceqn}
where $q_{\text{\tiny SA}}$, $k_{\text{\tiny SA}}$, $q_{\text{\tiny DM}}$, and $k_{\text{\tiny DM}}$ are learnable linear functions, and $\bff_{\text{\tiny SA}}$, $\bff_{\text{\tiny DM}}$ denote image features, and $D$ is the key/query feature dimension used for scaling. Softmax is applied row-wise (over the last dimension). The cross-attention layer in the adapter employs a multi-head structure, producing multiple attention masks. We average these masks across heads and use the result as the diffusion attention mask $\bA_{\text{\tiny DM}}$.

In practice, the Slot Attention encoder produces $\bA_{SA} \in \mathbb{R}^{(h \cdot w) \times K}$, while the adapter’s cross-attention mechanism yields $\bA_{DM} \in \mathbb{R}^{K \times (h' \cdot w')}$. In an ideal scenario, these two masks should be transposed from each other, as they encode the same underlying object–pixel relationships.

Having defined the attention masks in Eq.~\ref{eq:attention_mask}, we now describe how they are aligned through our guidance loss. We extract the decoder attention mask $\bA_{DM}$ from the third upsampling block of the U-Net. We then formulate a guidance loss to enforce the alignment between the two masks:
\begin{ceqn}
\begin{equation}
    \mathcal{L}_{\text{guidance}} = \text{BCE}(\bA_{SA}, \bA_{DM}^{\top})
    \label{eq:slotadapt_guidance}
\end{equation}
\end{ceqn}
where BCE is the binary cross-entropy loss, since the two masks differ in spatial resolution, we resize the decoder mask to match the encoder mask before applying BCE. This loss encourages the two attention maps to converge, effectively using the spatial knowledge in the pretrained diffusion model to guide the slot attention module towards producing more coherent, object-level segmentations. The total training objective is a weighted sum of the standard diffusion reconstruction loss $\mathcal{L}_{\theta}$ and this new guidance loss:
\begin{ceqn}
\begin{equation}
    \mathcal{L} = \mathcal{L}_{\theta} + \lambda \mathcal{L}_{\text{guidance}}
    \label{eq:slotadapt_total_loss}
\end{equation}
\end{ceqn}

where $\lambda$ is a weighting hyperparameter that balances reconstruction and guidance terms.

\section{Experimental Evaluation}
\label{sec:slotadapt_exp}

This section presents a thorough empirical evaluation of the SlotAdapt framework. We first detail the experimental setup, including datasets, baselines, and evaluation protocols. We then present the main results for unsupervised segmentation and compositional generation, supported by a detailed qualitative analysis. Finally, we conduct a series of comprehensive ablation studies to dissect the framework and validate our key architectural and methodological contributions.

\subsection{Setup}
\label{ssec:slotadapt_setup}

\paragraph{Datasets}
Our evaluation covers both synthetic and real-world scenarios. We use the synthetic \textbf{MOVi-E} dataset~\cite{greff2022kubric}, which features complex scenes with up to 23 objects. For real-world evaluation, we use two standard benchmarks: \textbf{PASCAL VOC 2012}~\cite{everingham2010pascalvoc} and \textbf{MS COCO 2017}~\cite{lin2014coco}. These datasets are challenging due to their multi-object nature, varied scenes, and large number of object classes (20 and 80, respectively).

\paragraph{Baselines}
We compare SlotAdapt against state-of-the-art unsupervised object-centric methods. On real-world datasets, we include Slot Attention (SA), SLATE, DINOSAUR, Latent Slot Diffusion (LSD), and SlotDiffusion. For fairness, all methods use a frozen DINOv2 as the visual encoder.

\paragraph{Metrics}
For the unsupervised object segmentation task, we use standard metrics from prior work: Foreground Adjusted Rand Index (FG-ARI), Mean Best Overlap (mBO), and Mean Intersection over Union (mIoU). We report mBO and mIoU at both the instance level ($mBO^i$, $mIoU^i$) and class level ($mBO^c$, $mIoU^c$) to assess whether distinct objects of the same class are correctly separated. For generation quality, we use the Fréchet Inception Distance (FID) \cite{heusel2017fid} and Kernel Inception Distance (KID) \cite{binkowski2018kid}.

\subsection{Main Results}
\label{ssec:slotadapt_results}

\begin{table}[t]
    \centering
    \caption[Segmentation Performance on Real-world Datasets]{\textbf{Unsupervised object segmentation on real-world datasets.} We compare SlotAdapt with state-of-the-art methods on VOC (left) and COCO (right). We present two versions of our method, with and without guidance loss.}
    \label{table:real-world-seg-quan}

    \begin{minipage}[t]{0.485\textwidth}
        \centering
        \scriptsize
        \setlength{\tabcolsep}{3.5pt}
        \begin{tabular}{lccc}
            \toprule
            \textbf{VOC} & FG-ARI & mBO$^i$ & mBO$^c$ \\
            \midrule
            SA + DINO ViT & 12.3 & 24.6 & 24.9 \\
            SLATE + DINO ViT & 15.6 & 35.9 & 41.5 \\
            DINOSAUR & 23.2 & 43.6 & 50.8 \\
            LSD & 18.7 & 40.5 & 43.5 \\
            SlotDiffusion & 17.8 & 50.4 & \textbf{55.3} \\
            \midrule
            Ours & 28.8 & \textbf{51.6} & 52.0 \\ 
            Ours + Guidance & \textbf{29.6} & 51.5 & 51.9 \\
            \bottomrule
        \end{tabular}
        \caption*{\textbf{(a)} VOC}
    \end{minipage}
    \hfill
    \begin{minipage}[t]{0.485\textwidth}
        \centering
        \scriptsize
        \setlength{\tabcolsep}{3.5pt}
        \begin{tabular}{lccc}
            \toprule
            \textbf{COCO} & FG-ARI & mBO$^i$ & mBO$^c$ \\
            \midrule
            SA + DINO ViT & 21.4 & 17.2 & 19.2 \\
            SLATE + DINO ViT & 32.5 & 29.1 & 33.6 \\
            DINOSAUR & 34.3 & 32.3 & 38.8 \\
            LSD & 33.8 & 27.0 & 30.5 \\
            SlotDiffusion & 37.2 & 31.0 & 35.0 \\
            \midrule
            Ours & \textbf{42.3} & 31.5 & 34.8 \\
            Ours + Guidance & 41.4 & \textbf{35.1} & \textbf{39.2} \\
            \bottomrule
        \end{tabular}
        \caption*{\textbf{(b)} COCO}
    \end{minipage}
\end{table}

\paragraph{Unsupervised Object Segmentation}
Table \ref{table:real-world-seg-quan} presents the main segmentation results on VOC and COCO. SlotAdapt consistently and significantly outperforms all baseline methods across nearly all metrics. On the challenging COCO dataset, our method with guidance achieves an FG-ARI of 41.4 and an instance-level mBO of 35.1, marking a substantial improvement over the next best methods, SlotDiffusion and DINOSAUR. The improvement in instance-level metrics is particularly noteworthy, as this metric specifically penalizes models that incorrectly merge multiple instances of the same class. Our strong performance indicates a superior ability to differentiate between individual objects, a critical prerequisite for meaningful compositional control.

\paragraph{Image Reconstruction and Compositional Editing}
Table \ref{tab:model_eval} provides a quantitative assessment of generation quality for standard reconstruction and the more challenging task of compositional generation. For reconstruction, SlotAdapt consistently outperforms competing approaches, showing that our adaptation strategy effectively preserves the generative fidelity of the pretrained model. More importantly, SlotAdapt demonstrates clear advantages in compositional generation, producing coherent and realistic scenes even when the underlying slot representations are manipulated for editing. This highlights the framework’s ability to maintain both visual quality and semantic consistency during complex editing operations.

\begin{table}[t]
    \centering
    \small
    \caption[Reconstruction and Compositional Generation Performance]{\textbf{Model Evaluation} Comparison of FID and KID scores for reconstruction (left) and compositional generation (right) across different methods.}
    \label{tab:model_eval}
    \setlength{\tabcolsep}{5pt}
    \begin{minipage}[t]{0.45\textwidth}
        \begin{tabular}{lc|c}
        \toprule
        Method & FID & KID$\times$1000 \\
        \midrule
        LSD & 35.537 & 19.086 \\
        SlotDiffusion & 19.448 & 5.852 \\
        Ours & \textbf{10.857} & \textbf{0.388} \\
        \bottomrule
        \end{tabular}
    \end{minipage}%
    \hspace{0.5cm} %
    \begin{minipage}[t]{0.45\textwidth}
        \begin{tabular}{lc|c}
        \toprule
        Method & FID & KID$\times$1000 \\
        \midrule
        LSD & 167.232 & 103.482 \\
        SlotDiffusion & 64.213 & 57.309 \\
        Ours & \textbf{40.568} & \textbf{34.381} \\
        \bottomrule
        \end{tabular}
    \end{minipage}
\end{table}

\paragraph{Qualitative Analysis}

\begin{figure}[h!]
    \centering
    \includegraphics[width=\textwidth]{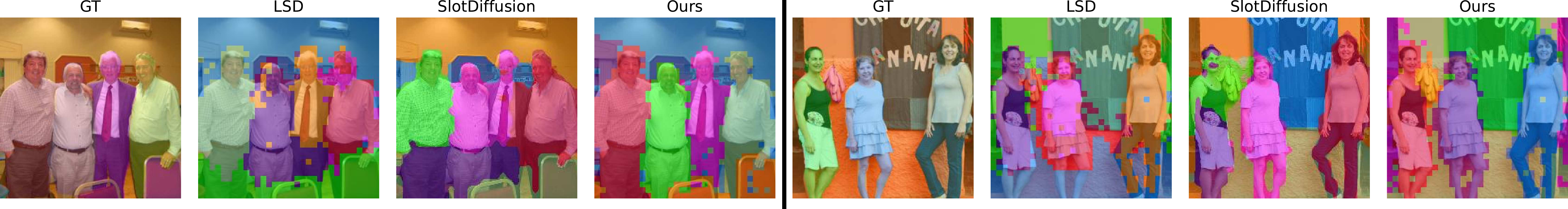}
    \caption[Qualitative segmentation comparison of SlotAdapt with baselines]{Qualitative comparisons with other methods on COCO. SlotAdapt can more effectively differentiate between object instances of the same class compared to other methods.}
    \label{fig:seg_compare}
\end{figure}

\begin{figure}[h!]
    \centering
    \includegraphics[width=\textwidth]{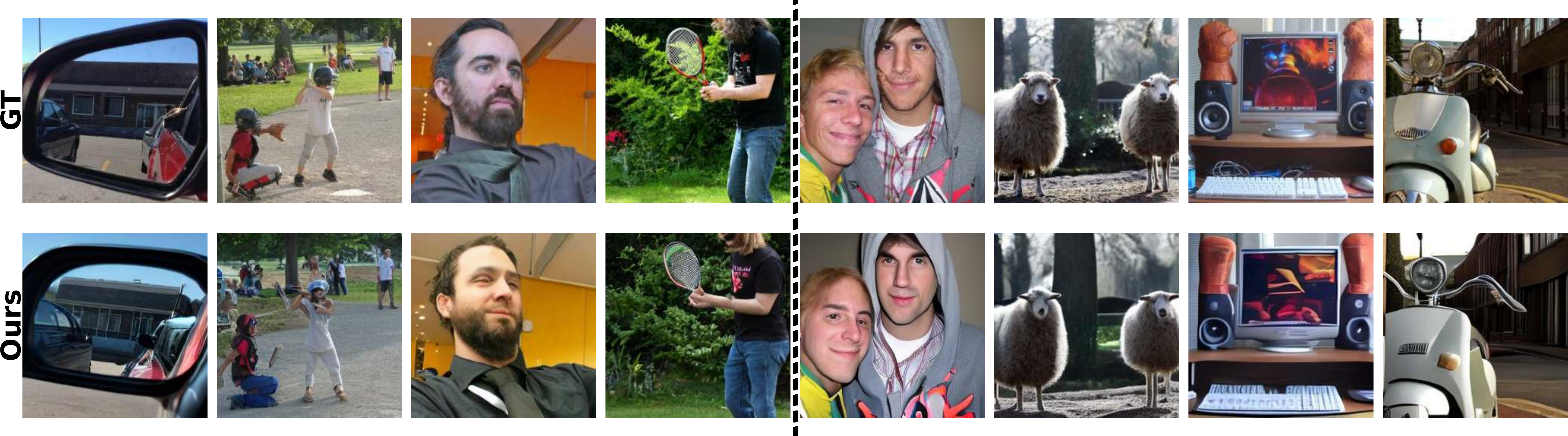}
    \caption[High-fidelity reconstruction results from SlotAdapt]{Generation Results. Sample images reconstructed by SlotAdapt on COCO (left) and VOC (right). SlotAdapt generates reconstructions highly faithful to the original input images.}
    \label{fig:gen}
\end{figure}

A strong qualitative performance supports the quantitative results. The high fidelity of our reconstructions is evident in Figure \ref{fig:gen}, where SlotAdapt accurately reproduces complex textures, lighting, and fine details from the original images. Figure \ref{fig:seg_compare} compares segmentation quality, where SlotAdapt produces fewer artifacts and more coherent object masks than LSD and SlotDiffusion.

\begin{figure}[h!]
    \centering
    \includegraphics[width=\textwidth]{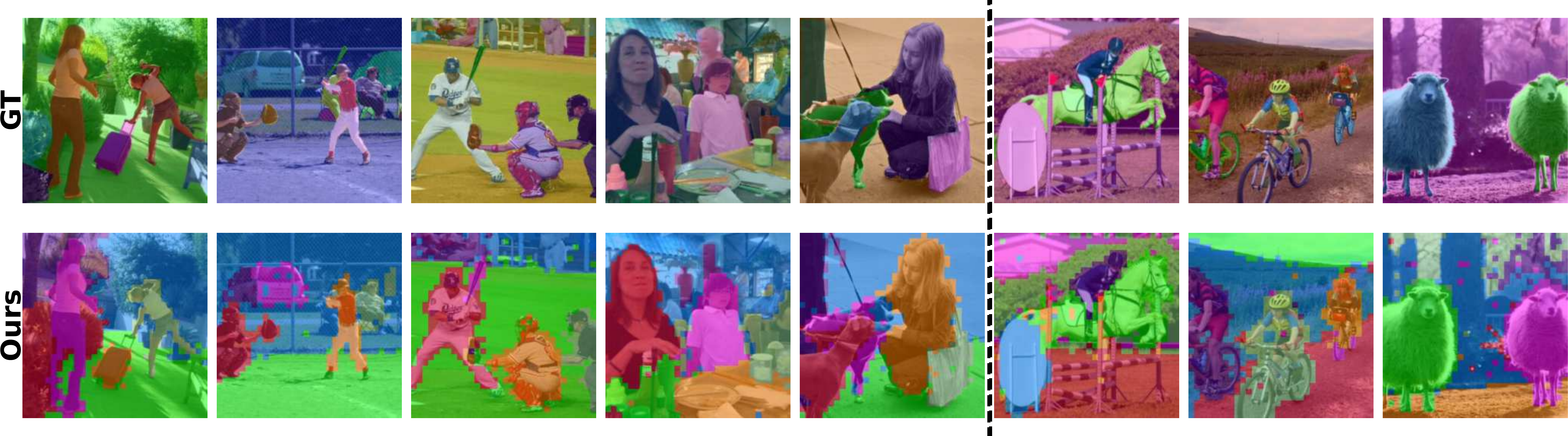}
    \caption[Unsupervised object segmentation results for SlotAdapt]{Unsupervised Object Segmentation. Visualizations of predicted segments on COCO (left) and VOC (right). SlotAdapt accurately binds distinct instances belonging to the same class.}
    \label{fig:seg}
\end{figure}

\begin{figure}[h!]
    \centering
    \includegraphics[width=\textwidth]{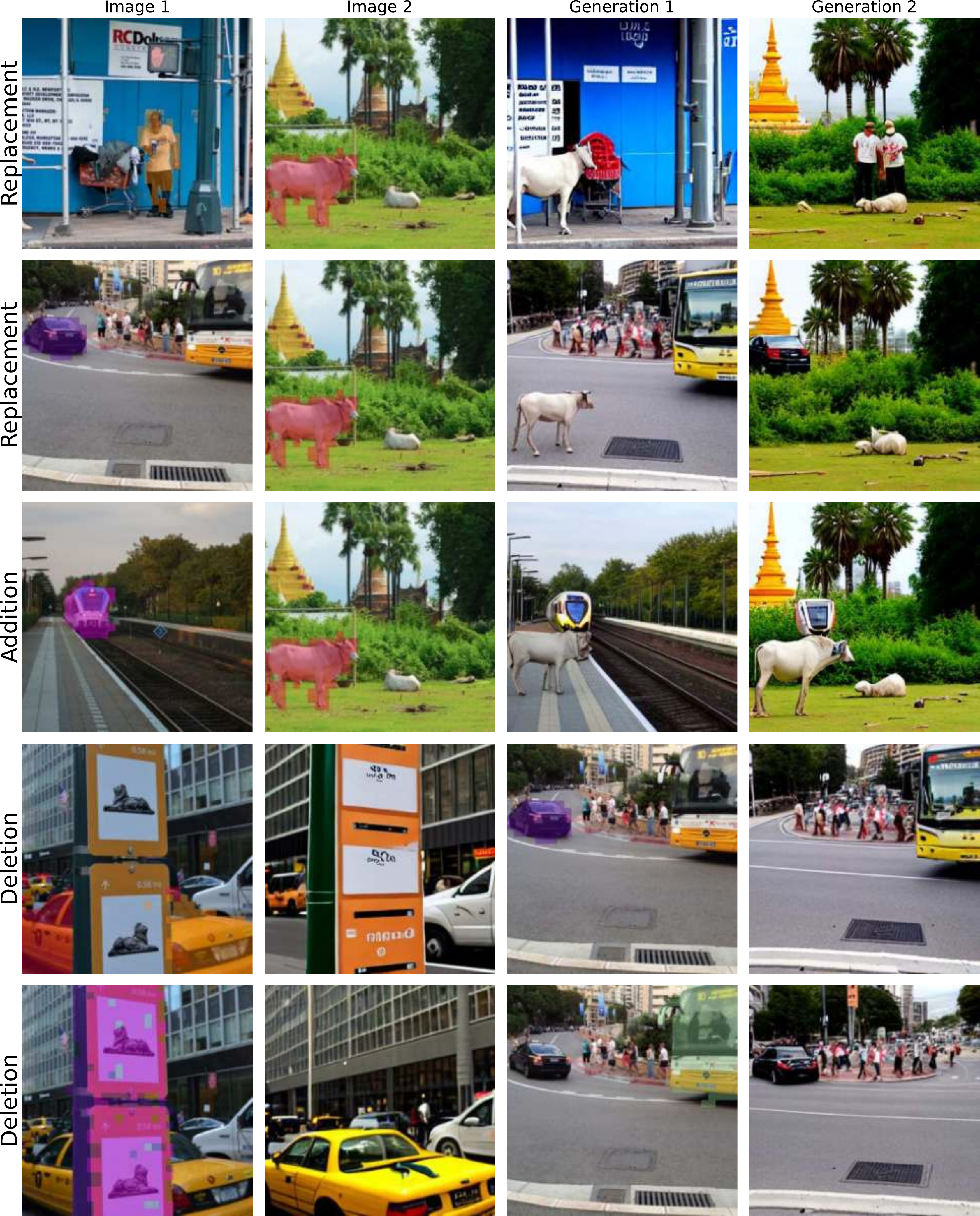}
    \caption[Compositional editing examples with SlotAdapt]{Compositional Editing. We demonstrate object removal, replacement, and addition edits on COCO images by using slots. The edits are successful and result in realistic and coherent scenes.}
    \label{fig:comp_gen}
\end{figure}

Furthermore, Figure \ref{fig:seg} highlights our model's superior segmentation. These examples show that SlotAdapt produces more coherent masks that better align with whole objects, avoiding the fragmented or merged segments often produced by other methods. This improved segmentation directly enables the advanced compositional editing capabilities shown in Figure \ref{fig:comp_gen}. By manipulating the learned slots, we can perform targeted, zero-shot edits—such as removing a stop sign, replacing a person with a cow, or adding an elephant to a new scene—with realistic and coherent results. To our knowledge, this is the first unsupervised slot-based approach to demonstrate high-fidelity compositional editing on COCO.

\begin{figure}[!ht]
\centering
\includegraphics[width=.9\textwidth]{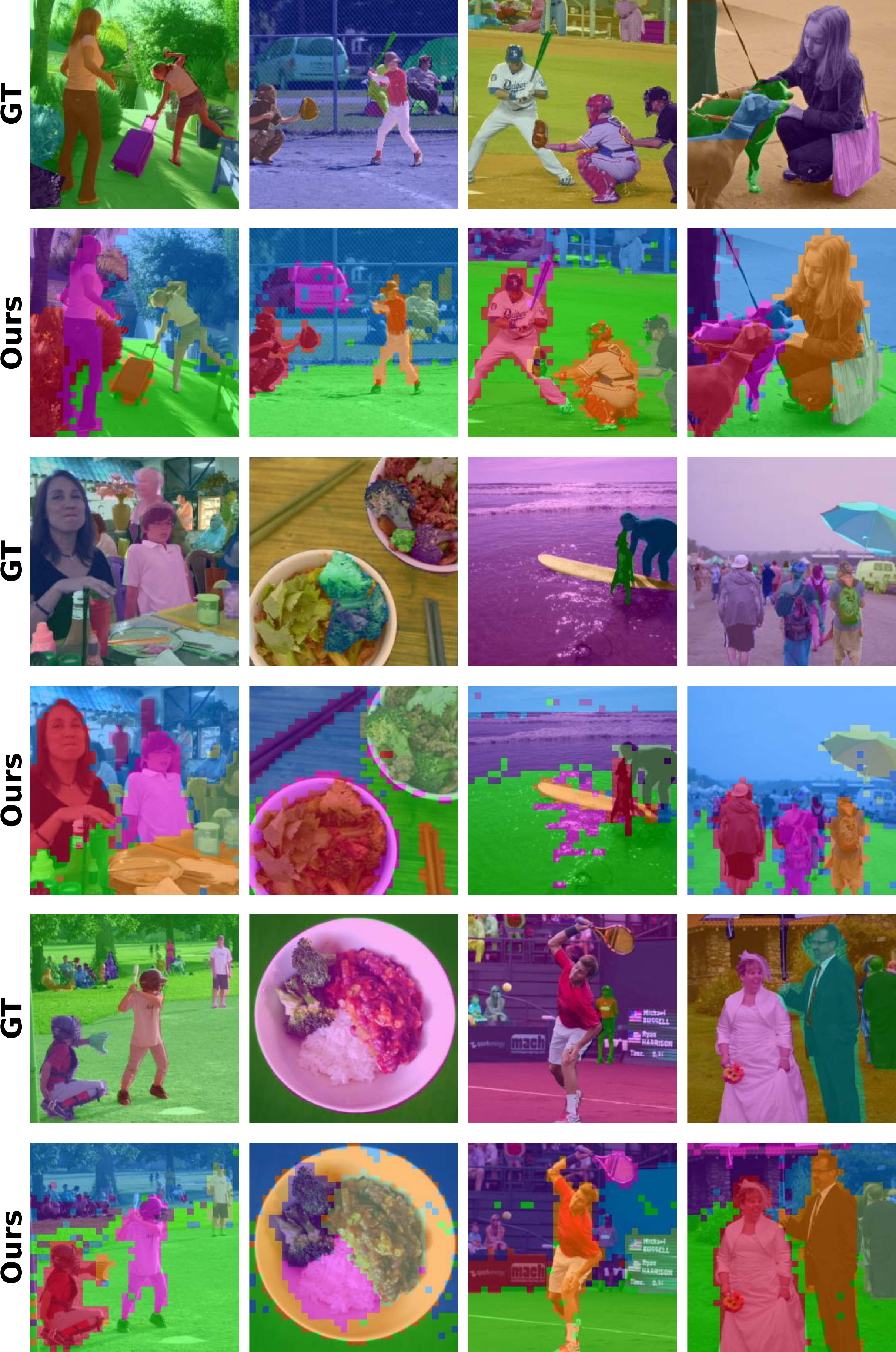}
\vskip -0.1in
\caption[Qualitative segmentation on COCO]{Qualitative comparisons on COCO}
\label{fig:supp_coco_seg}
\end{figure}

\begin{figure}[!ht]
\centering
\includegraphics[width=.9\textwidth]{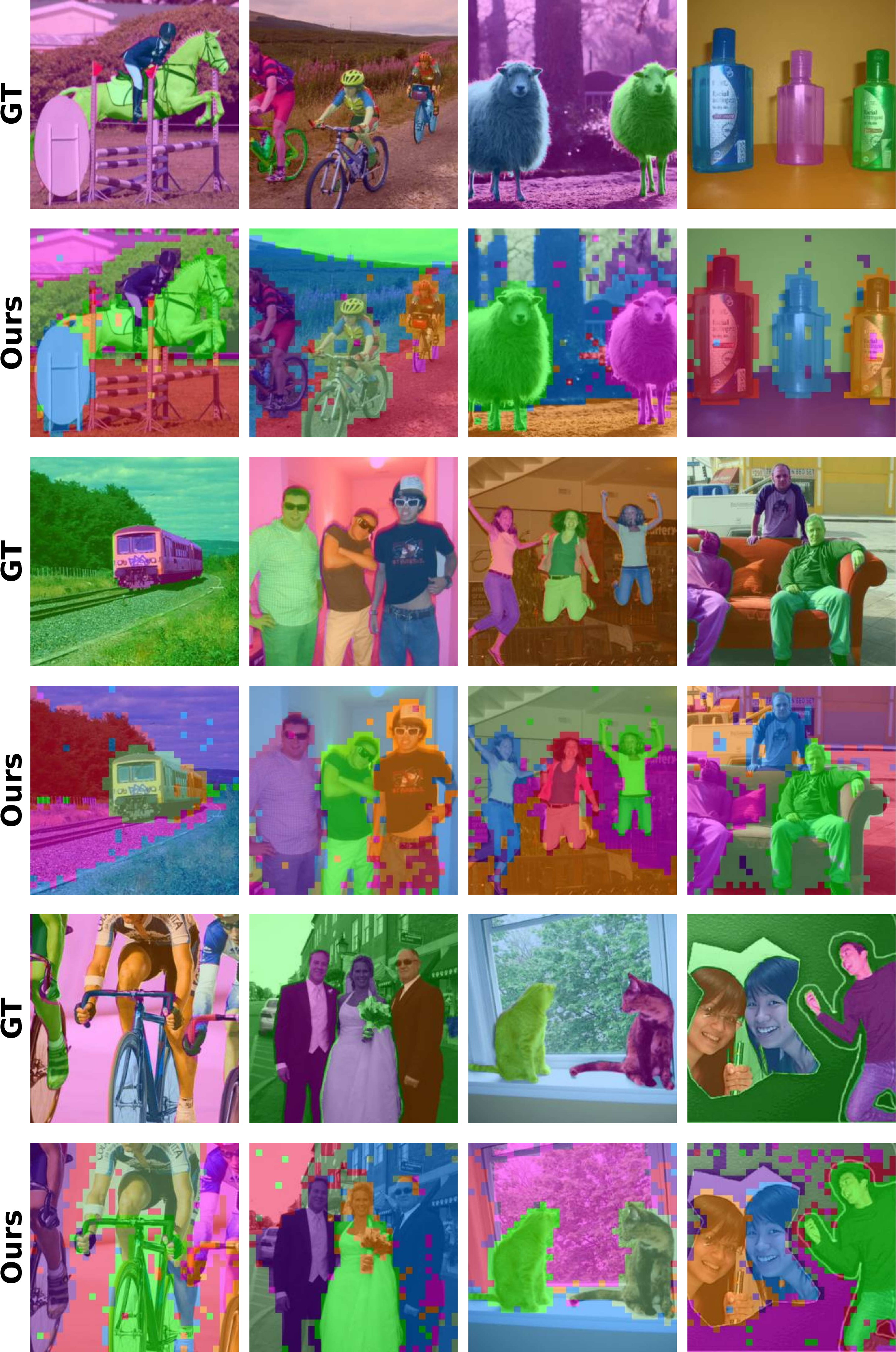}
\vskip -0.1in
\caption[Qualitative segmentation on VOC]{Qualitative comparisons on VOC}
\label{fig:supp_voc_seg}
\end{figure}

\subsection{Ablation Studies and Deeper Analysis}
\label{ssec:slotadapt_ablations}
We conduct ablation studies to evaluate how individual components, hyperparameters, and design decisions contribute to the overall performance of the framework.

\begin{table}[t]
    \caption[Effect of different guidance strategies on Segmentation Performance]{
    \textbf{Evaluation of guidance strategies.} We present the segmentation performance on COCO for different guidance strategies. Joint guidance gives the best scores and significantly improves over no guidance option.
    \label{tab:guidance-ablation}
    }
    \centering
        \centering
        \small
        \setlength{\tabcolsep}{5pt}
        \begin{tabular}{lccccc}
        \toprule
        \textbf{} & FG-ARI & mBO$^i$ & mBO$^c$ & mIoU$^i$ & mIoU$^c$ \\
        \midrule
        No guidance & 42.3 & 31.5 & 34.8 & 31.7 & 38.5 \\
        \midrule
        Slot Guidance & 41.2 & 33.4 & 36.9 & 33.1 & 37.9 \\
        DM Guidance & 42.0 & 31.2 & 34.6 & 32.0 & 38.4 \\
        Joint Guidance & 41.4 & \textbf{35.1} & \textbf{39.2} & \textbf{36.1} & \textbf{41.4} \\
        Multiplication Guidance & \textbf{43.3} & 31.9 & 35.3 & 31.7 & 36.4  \\
        \bottomrule
        \end{tabular}
\end{table}

\begin{table}[!h]
\centering
\caption[Architectural ablations on MOVi-E]{\textbf{Architectural ablations on MOVi-E.} We examine the effects of architectural choices on segmentation and representation performance. We present block combinations on the left and register token choice on the right. Up, Down and Mid refer to all upsampling blocks, all downsampling blocks and middle block in the diffusion model.}
\label{tab:combined_ablation}
\resizebox{\textwidth}{!}{%
\begin{tabular}{@{}lccccc|ccc@{}}
\toprule
& \multicolumn{5}{c}{Conditioning Blocks}& \multicolumn{3}{c}{Register Token} \\
\cmidrule(lr){2-6} \cmidrule(lr){7-9}
& Up+Down+Mid & Only Up & Only Down & Up+Mid & Up+Down & No Token & Slot Pooling & Feature Pooling \\
\midrule
\multicolumn{9}{c}{Segmentation (\%)} \\
\midrule
FG-ARI (\textuparrow) & 56.89 & 57.38 & \textbf{57.93} & 57.39 & 56.45 & 54.38 & 56.27 & \textbf{57.18} \\
mBO (\textuparrow) & 39.59 & 43.05 & 40.20 & 39.96 & \textbf{43.38} & 40.07 & 41.65 & \textbf{43.98} \\
mIoU (\textuparrow) & 37.75 & 41.53 & 38.77 & 38.83 & \textbf{41.86} & 40.07 & \textbf{40.10} & 39.83 \\
\midrule
\multicolumn{9}{c}{Representation} \\
\midrule
Category (\textuparrow) & 43.92 & 43.82 & \textbf{45.88} & 41.54 & 43.91 & 42.42 & \textbf{43.54} & 42.63 \\
Position (\textdownarrow) & 1.92 & 1.82 & \textbf{1.61} & 1.92 & 1.72 & 1.89 & \textbf{1.75} & 1.78 \\
3D B-Box (\textdownarrow) & 3.94 & 3.78 & \textbf{3.48} & 3.95 & 3.75 & 3.83 & \textbf{3.77} & 3.78 \\
\bottomrule
\end{tabular}%
} 
\end{table}

\begin{figure}[h!]
    \centering
    \includegraphics[width=0.8\textwidth]{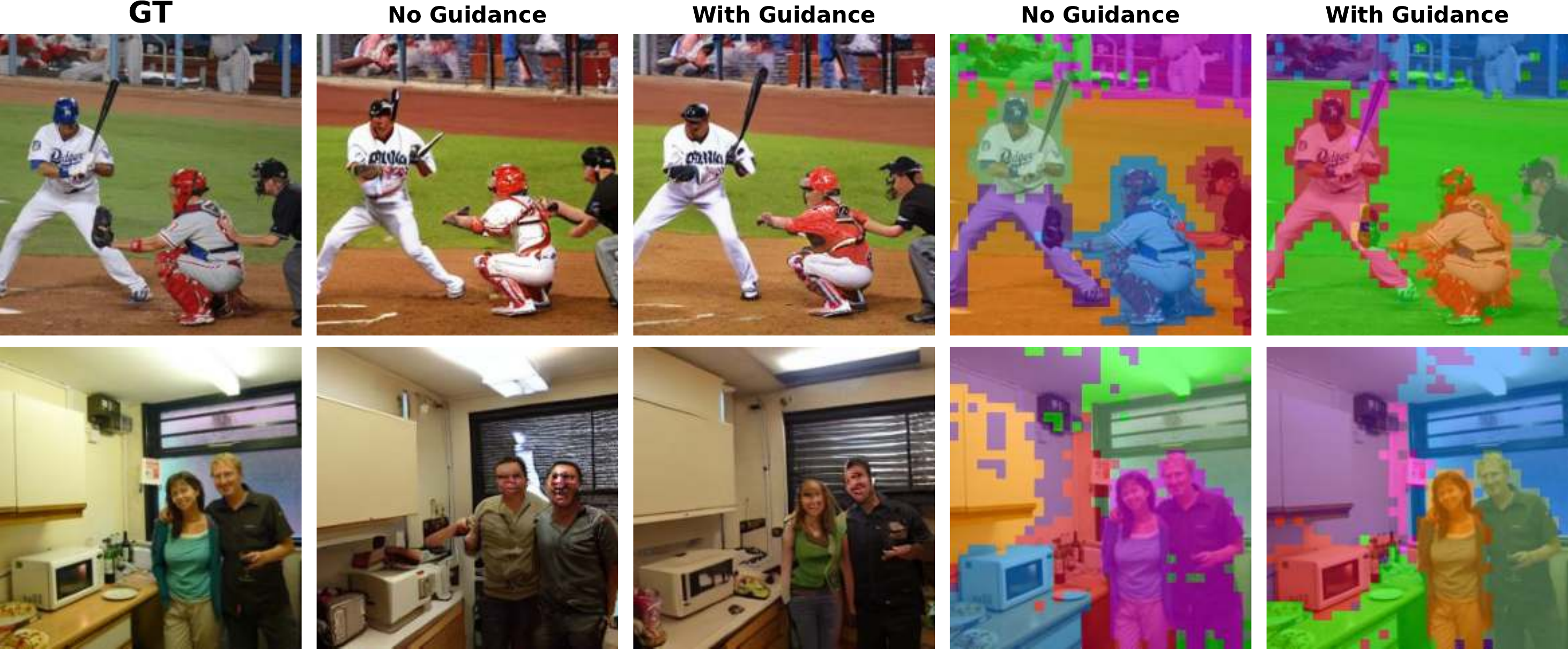}
    \caption[Qualitative comparison of generation with and without attention guidance]{Qualitative comparison: with vs. without guidance. We visualize generated images and predicted segments on the COCO dataset. Guidance improves segmentation quality and alignment with whole objects.}
    \label{fig:guidance_compare}
\end{figure}

\paragraph{Effectiveness of Attention Guidance}
Table \ref{tab:guidance-ablation} shows the impact of our self-supervised guidance loss on COCO. The ``joint guidance" strategy, where encoder and decoder attention maps are refined together, consistently yields the strongest results across segmentation metrics. This indicates that enforcing consistency between the perception and generation modules provides a powerful self-supervisory signal. The effect is also visible in Figure \ref{fig:guidance_compare}, where guidance reduces part–whole confusion and produces masks that more reliably capture entire objects.

\paragraph{Impact of Architectural Choices}
Table \ref{tab:combined_ablation} investigates two key architectural decisions on the MOVi-E dataset. The right side shows that including a \textbf{register token} provides a consistent and significant performance benefit across all metrics compared to using no token. This validates our hypothesis that explicitly dedicating a representation to global context allows the individual object slots to become more specialized. The left side shows that applying \textbf{adapter conditioning} to both the upsampling and downsampling blocks yields the best overall performance, suggesting that injecting object-level information at multiple stages of the generative process is most effective.

\begin{table*}[t]
\begin{small}
    \caption[Performance on MOVi-E Dataset]{\textbf{Comparative evaluation on MOVi-E:}
    (Left) Segmentation results,
    (Right) Representation assessment: We evaluate slots through predictive probing. Spatial attributes (position, 3D bounding box) are assessed via MSE (mean squared error), while categorical predictions are assessed by classification accuracy.}
    \label{tab:movie}
    \centering
    \begin{adjustbox}{max width=\textwidth}
    \begin{tabular}{lcccc}
    \toprule
    \scriptsize{\textbf{Segmentation}}          & SLATE        & SLATE$^+$      & LSD & Ours   \\
    \midrule
    mBO ($\uparrow$)         & 30.17 & 22.17  & 38.96 & \textbf{43.38} \\
    mIoU ($\uparrow$)        & 28.59 & 20.63  & 37.64 & \textbf{41.85} \\
    FG-ARI ($\uparrow$)      & 46.06 & 45.25  & 52.17 & \textbf{56.45} \\
    \bottomrule
    \end{tabular}
        \begin{tabular}{lcccc}
    \toprule
    \scriptsize{\textbf{Representation}}           & SLATE        & SLATE$^+$      & LSD & Ours   \\
    \midrule
    Position ($\downarrow$)  & 2.09 & 2.15   & 1.85 & \textbf{1.77}  \\
    3D B-Box ($\downarrow$)   & 3.36 & 3.37  & \textbf{2.94} & 3.75    \\
    Category ($\uparrow$)    & 38.93 & 38.00 & 42.96 & \textbf{43.92} \\
    \bottomrule
    \end{tabular}
    \end{adjustbox}   
\end{small}
\end{table*}

\paragraph{Performance on Complex Synthetic Data}
To test the limits of our model's discovery and representation capabilities, we evaluated it on the synthetic MOVi-E dataset, with results shown in Table \ref{tab:movie}. SlotAdapt demonstrates significant improvements in segmentation, with an approximate 10\% enhancement in object discovery and segmentation accuracy. More importantly, we assessed the quality of the learned slots for downstream tasks by training a simple 2-layer MLP to predict object properties from them. The strong performance on predicting object category, 3D position, and bounding boxes confirms that the learned slots are not just useful for reconstruction but capture meaningful, disentangled semantic and spatial information about the objects.

\begin{table}[t]   
   \centering
   \small
   \setlength{\tabcolsep}{5pt}
   \arrayrulecolor{black}
   \caption[Impact of Slot Count on Performance]{\textbf{Impact of Slot Count on Performance:} We analyze the effect of varying the number of slots on the COCO dataset.
   }
   \label{tab:slot_num}
   \begin{tabular}{lccccccc}
       \toprule
       \multirow{2}{*}{Method} & 
       \multirow{2}{*}{FG-ARI} & 
       \multicolumn{2}{c}{Instance} & 
       \multicolumn{2}{c}{Semantic} & 
       \multirow{2}{*}{FID} & 
       \multirow{2}{*}{KID} \\
       \cmidrule(lr){3-4} \cmidrule(lr){5-6}
       & & mBO & mIoU & mBO & mIoU & & \\
       \midrule
       
       80 slots & 18.1 & 23.2 & 26.4 & 28.6 & 32.9 & 114.236 & 64.610 \\
       7 slots & 41.4 & 35.1 & 36.1 & 39.2 & 41.4 & 10.857 & 0.388 \\
       
       \bottomrule
   \end{tabular}
\end{table}

\paragraph{Sensitivity to Hyperparameters}
We further analyze the model's sensitivity to key hyperparameters.
\begin{itemize}
    \item \textbf{Slot Count:} As shown in Table \ref{tab:slot_num}, deviating from the convention of matching slot count to the maximum number of objects (7 for COCO) by dramatically increasing it to 80 led to a severe degradation in performance across all metrics. This suggests that an excessive number of slots creates an overly competitive and redundant representation space that the model struggles to optimize, confirming that a concise slot count is crucial.
    \item \textbf{Guidance Loss Timing:} We hypothesized that the guidance loss would be more effective after stabilizing the slot and adapter representations. Table \ref{tab:guidance_start} confirms this: starting the guidance loss after 40K training iterations yields significantly better segmentation performance than applying it from the beginning. This "warm-up" period allows the attention maps to become meaningful before they are used as a supervisory signal.
    \item \textbf{Classifier-Free Guidance (CFG) Scale:} The CFG scale is a critical inference-time parameter that controls the strength of conditioning. We performed a sweep of CFG values, with results in Table \ref{tab:cfg_evals}. We found that a small CFG value of 1.3 yielded the optimal balance, achieving the best FID and KID scores.
\end{itemize}

\begin{table}[ht]
   \caption[Impact of Guidance Loss Timing]{
    \textbf{Impact of Guidance Loss Timing:} We compare the effect of applying the guidance loss from the start of training versus introducing it after the model has learned initial representations. Results are presented on the COCO dataset.
    }
   \label{tab:guidance_start}
   
   \centering
   \small
   \setlength{\tabcolsep}{5pt}
   \arrayrulecolor{black}
   
   \begin{tabular}{lcccccc}
       \toprule
       \multirow{2}{*}{Method} & 
       \multirow{2}{*}{FG-ARI} & 
       \multicolumn{2}{c}{Instance} & 
       \multicolumn{2}{c}{Semantic} \\
       \cmidrule(lr){3-4} \cmidrule(lr){5-6}
       & & mBO & mIoU & mBO & mIoU \\
       \midrule
       
       Start from 0 & 37.85 & 32.65 & 33.99 & 36.059 & 39.254 \\
       Original (start after 40K) & 41.4 & 35.1 & 36.1 & 39.2 & 41.4 \\
       
       \bottomrule
   \end{tabular}
\end{table}

\begin{table}[t]
    \caption[Impact of CFG Value on Generation Quality]{\textbf{Impact of CFG Value on Generation Quality:} We evaluate the effect of different CFG values on generation quality using the COCO dataset.
    }
    \label{tab:cfg_evals}
    \centering
    \small
    \setlength{\tabcolsep}{5pt}
    
    \arrayrulecolor{black}
    
    \begin{tabular}{lcc}
        \toprule
        CFG Value & FID & KID$\times$1000 \\
        \midrule
        
        7.5 & 21.350 & 6.271 \\
        5.0 & 17.558 & 4.236 \\
        2.5 & 13.734 & 2.151 \\
        2.0 & 12.427 & 1.424 \\
        1.5 & 11.041 & 0.590 \\
        1.4 & 10.880 & 0.459 \\
        1.3 & \textbf{10.857} & \textbf{0.388} \\
        1.2 & 11.057 & 0.492 \\
        
        \bottomrule
    \end{tabular}
\end{table}

\begin{table}[ht!]
   \caption[Impact of Global Information Capturing Strategies]{
    \textbf{Impact of Global Information Capturing Strategies:} We compare slot averaging and the use of an additional slot token for capturing global information. Results are presented on the COCO dataset.
    }
   \label{tab:additional_token}
   
   \centering
   \small
   \setlength{\tabcolsep}{4.5pt}
   \arrayrulecolor{black}
   
   \begin{tabular}{lccccccc}
       \toprule
       \multirow{2}{*}{Method} & 
       \multirow{2}{*}{FG-ARI} & 
       \multicolumn{2}{c}{Instance} & 
       \multicolumn{2}{c}{Semantic} & 
       \multirow{2}{*}{FID} & 
       \multirow{2}{*}{KID} \\
       \cmidrule(lr){3-4} \cmidrule(lr){5-6}
       & & mBO & mIoU & mBO & mIoU & & \\
       \midrule
       
       Additional Slot Token & 43.8 & 31.9 & 32.4 & 35.5 & 37.3 & 11.212 & 0.431 \\
       Slot Average Token & 42.3 & 31.5 & 34.8 & 34.8 & 38.5 & 10.857 & 0.388 \\
       
       \bottomrule
   \end{tabular}
\end{table}

\begin{table}[t]
   \caption[Impact of a Better Segmentation Method]{
   \textbf{Impact of a Better Segmentation Method:} We evaluate the effect of a better segmentation model on performance by replacing Slot Attention with BOQ-SA, an improved version. Results are presented on the COCO dataset.
   }
   \label{tab:boqsa}
   \centering
   \small
   \setlength{\tabcolsep}{5pt}
   \arrayrulecolor{black}
   
   \begin{tabular}{lcccccc}
       \toprule
       Method & 
       FG-ARI & 
       mBO$^i$ & 
       mIoU$^i$ & 
       mBO$^c$ & 
       mIoU$^c$ \\
       \midrule
       
       Slot Attention & 42.3 & 31.5 & 31.7 & 34.8 & 38.5 \\
       BOQ-SA & 42.2 & 31.2 & 32.551 & 35.266 & 37.82 \\
       
       \bottomrule
   \end{tabular}
\end{table}

\paragraph{Analysis of Alternative Components}
Finally, we explored alternatives for core components of our architecture.
\begin{itemize}
    \item \textbf{Register Token Strategy:} We experimented with an alternative to our slot-pooling register token: adding an extra, dedicated "global slot" to the Slot Attention module. The results in Table \ref{tab:additional_token} reveal an interesting trade-off: the dedicated global slot improved foreground segmentation (FG-ARI) but led to worse semantic segmentation and generation quality (FID/KID). We hypothesize this is due to the competitive nature of slot attention; the global slot must compete with object slots, limiting its ability to capture holistic information.
    \item \textbf{Improved Slot Encoder:} To see if our model's performance was limited by the encoder, we replaced the standard Slot Attention module with BOQ-SA, an improved version with an optimized initialization scheme. As seen in Table \ref{tab:boqsa}, this resulted in only very modest changes in performance. This suggests that combining a powerful DINOv2 backbone and our attention guidance mechanism already provides a sufficiently strong object discovery signal.
\end{itemize}

\subsection{Discussion of Limitations}
While SlotAdapt represents a significant advance, it is important to acknowledge its limitations. First, while the compositional edits are generally high-fidelity, the reconstructed background can sometimes exhibit subtle changes compared to the original image, as the diffusion model reimagines the occluded regions. Second, like most OCL methods, SlotAdapt can still be susceptible to under- and over-segmentation in highly cluttered or ambiguous scenes. Future work could explore more sophisticated slot-management techniques, such as dynamic merging and splitting of slots, to address these cases. Finally, the framework's reliance on a pretrained diffusion model means its effectiveness can be domain-limited by the training data of that base model.

\section{Chapter Summary}
\label{sec:slotadapt_summary}

This chapter introduced SlotAdapt, a framework that integrates Slot Attention with pretrained diffusion models through three components: lightweight adapters for slot conditioning, a register token for global context, and a self-supervised attention guidance loss. Empirically, SlotAdapt achieves state-of-the-art unsupervised object discovery on COCO and supports zero-shot compositional editing from unsupervised representations. To our knowledge, by addressing conditioning misalignment, SlotAdapt is the first unsupervised slot-based framework to enable high-fidelity compositional editing on complex real-world datasets. These results establish a foundation for extending object-centric conditioning to temporal domains.

Beyond its immediate results, SlotAdapt highlights the broader principle that large pretrained diffusion models can be adapted to structured representations with minimal additional training. This efficient adaptation strategy provides a robust foundation for scaling object-centric learning to more complex generative tasks. Having established a successful framework for object-centric image synthesis, we are now prepared to extend these ideas to the video domain. The next chapter builds on these principles to address the challenges of object-centric video synthesis and compositional editing.
\newpage
\chapter[Extending Compositionality to the Temporal Domain]{Extending Compositionality to the Temporal Domain: Object-Centric Video Synthesis}
\label{ch:video}

\section{Introduction}
\label{sec:video_intro}

In the earlier chapters, we progressively built toward a central goal: enabling machines to perceive and generate visual scenes in a manner that reflects the structured, object-centric nature of the world. In Chapter ~\ref{chap:slotadapt}, we introduced SlotAdapt, a framework that addresses the core challenge of adapting powerful pretrained diffusion models for object-centric \textit{image} generation. By demonstrating high-fidelity, compositional control over static scenes, SlotAdapt established a robust methodology for uniting structured latent representations with state-of-the-art synthesis. In this chapter, we extend this foundation into the temporal domain, addressing the more complex challenge of modeling compositionality in video.

The real world is not just a set of separate snapshots; it is a continuous flow of events shaped by temporal structure. Humans have a natural ability to parse this dynamic stream into a coherent mental model of persistent, interacting entities \cite{spelke2007core, ullman2017mind}. We do not merely see a person in one frame and a person in the next; we perceive the \textit{same} person moving through space. This ability to maintain object files, tracking identities through occlusion, and understanding motion is fundamental to our intelligence \cite{kahneman1992reviewing}. For artificial systems, however, modeling this temporal dimension remains a formidable challenge.

Transitioning from images to video introduces a host of complexities that static models are unable to handle. A generative video model must not only discover the objects in each frame but also solve the temporal binding problem: ensuring the representation for a specific object remains stable as it moves, deforms, and interacts with its environment. This task is particularly challenging under real-world conditions such as motion blur, abrupt lighting changes, and frequent occlusions. Early attempts in temporal object-centric learning often relied on auxiliary supervisory signals like optical flow or depth maps to enforce this consistency \cite{kipf2021conditional, elsayed2022savi++}. While helpful, such methods are inherently fragile, as these external cues are often noisy and unreliable in the very scenarios where tracking is most difficult.

In parallel, the field of generative modeling has seen the rise of large-scale text-to-video diffusion models capable of producing high-quality video \cite{ho2022imagen, singer2022make, villegas2022phenaki}. Yet, these models typically operate on holistic, entangled scene representations. These models treat video as a monolithic block of pixels and lack explicit object-level structure, which limits compositional reasoning and fine-grained control. This limits their ability to support tasks that require a disentangled understanding of individual entities within a scene.

This chapter directly addresses this open problem at the intersection of object-centric learning and generative video synthesis. We propose a fully self-supervised framework that combines deep representation learning with high-quality synthesis capabilities. We extend the core principles of SlotAdapt, introduced in Chapter~\ref{chap:slotadapt}, adapting its architecture to learn temporally coherent slots that encode object identity, motion, and interaction, all without relying on handcrafted external signals. By conditioning a pretrained diffusion model on these learned temporal features, we achieve video reconstructions that are both photorealistic and semantically grounded. To our knowledge, this is the first approach to combine object-centric representation learning with high-quality generative capabilities for real-world videos. This unified architecture enables direct and flexible compositional editing—such as object insertion, deletion, and replacement—by manipulating learned, unsupervised object representations, all while maintaining temporal coherence within a self-supervised framework \cite{akan2025compositional}.

\section{From Images to Video: The Adapted Framework}
\label{sec:video_methodology}

\begin{figure*}[!htbp]
    \centering
    \includegraphics[width=0.95\textwidth]{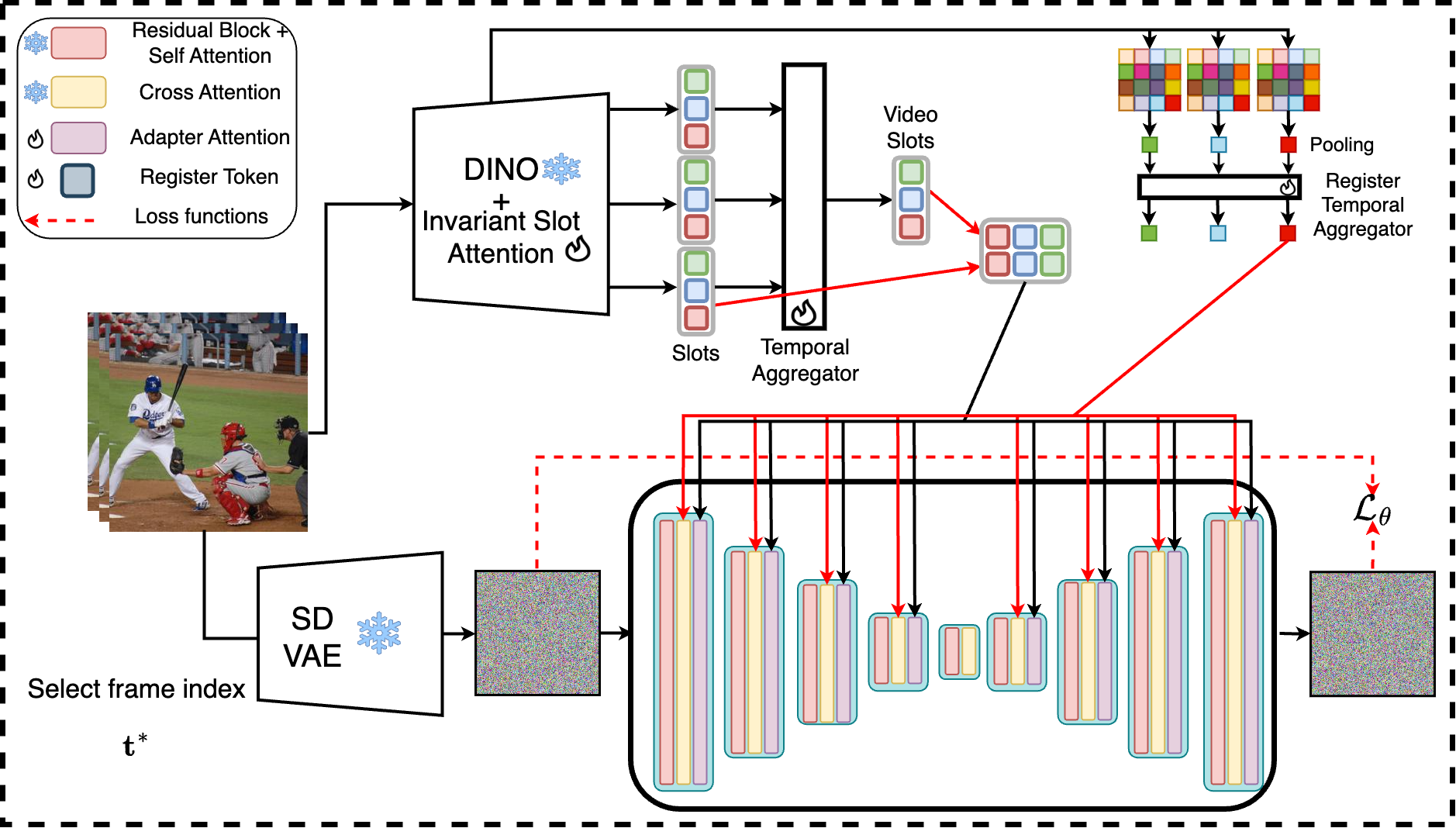}
    \caption[The SlotAdapt Video Architecture]{
        \textbf{Architecture Block Diagram.}
        We extract object-centric and temporally consistent information from input video frames using a visual backbone composed of DINOv2 and Invariant Slot Attention (ISA). The ISA mechanism generates slots for each frame, which are then aggregated temporally using a Transformer-based temporal aggregator to produce enriched, temporally-aware video slots. Concurrently, global context information is summarized by average pooling frame-level features and further processed through a separate temporal aggregator to produce global scene tokens. A pretrained Stable Diffusion Variational Autoencoder (VAE) encodes a randomly selected frame into latent space, and Gaussian noise is subsequently added. The diffusion model is explicitly conditioned on both the temporally aggregated video slots and the slots from the selected frame (shown here as the last frame for visualization purposes, though in practice this could be any frame) via additional adapter attention layers, and on the global scene token using the diffusion model's native cross-attention layers. During training, the model learns to predict the added noise, with the diffusion loss ($\mathcal{L}_{\theta}$) measuring the prediction error. This training strategy ensures temporally coherent, object-centric video synthesis and intuitive compositional editing capabilities across video frames.
    }
    \label{fig:video_method_diagram}
\end{figure*}

Adapting the successful image-based framework from Chapter~\ref{chap:slotadapt} to the dynamic and complex domain of video required more than a simple extension; it required a fundamental redesign of the model's core components to handle the critical dimension of time. While the main idea of using a pretrained diffusion model with lightweight adapters is preserved for efficiency, both the object-centric encoder and the training strategy were re-engineered to explicitly model temporal continuity, motion, and object permanence. This section explains the adapted framework, which consists of two main components: a temporal object-centric encoder designed to discover and track entities across video frames, and a slot-conditioned diffusion decoder capable of synthesizing photorealistic and temporally coherent video content from these learned representations. A block diagram illustrating the complete architecture is presented in Figure~\ref{fig:video_method_diagram}.

\subsection{Object-Centric Temporal Encoder: Discovering and Tracking Entities}
\label{sec:video_encoder}

The primary goal of the encoder is to perceive a raw video clip and decompose it into a set of structured, object-centric representations that remain consistent over time. This process begins by establishing a rich, per-frame feature extraction and then employs specialized mechanisms to group these features into object-specific slots and enrich them with temporal context.

Given an input video clip composed of $L$ frames, we first process each frame individually through a frozen DINOv2 visual backbone \cite{oquab2023dinov2}. This is consistent with the image model and provides a powerful set of patch-level feature vectors, $\mathbf{f}_t \in \mathbb{R}^{N \times d}$ for each frame $t$, where $N$ is the number of patches and $d$ is the feature dimension. These features, which are known to be highly semantic and correspond well to object boundaries, form the basis for our object discovery process. The central challenge, however, is to ensure that the representation for a given object remains stable even as its position, scale, and orientation change from one frame to the next. To solve this, we introduce two key architectural innovations.

\subsubsection{Disentangling Identity from Pose with Invariant Slot Attention (ISA)}
A significant limitation of standard Slot Attention for video is that the learned slot representations often entangle an object's intrinsic identity with its transient spatial properties, or ``pose." This is problematic for tracking, as the slot for a moving object would need to change substantially in every frame, making it difficult to maintain a consistent identity.

To overcome this, we replace the standard Slot Attention module with \textbf{Invariant Slot Attention (ISA)} \cite{biza2023invariant}. ISA is a more sophisticated mechanism specifically designed to learn a \textit{disentangled} representation by explicitly separating an object’s identity from its pose. The core idea is to augment each of the $K$ slots with its own learnable pose parameters: a position $\mathbf{s}^j_{p,t} \in \mathbb{R}^2$ and a scale $\mathbf{s}^j_{s,t} \in \mathbb{R}^2$. These parameters allow the model to establish a canonical, object-centric frame of reference for each slot.

The attention process then operates within this relative frame, making the computation invariant to the object's global position and scale in the image. This allows the core slot vector, $\mathbf{s}^j_t$, to focus exclusively on learning the intrinsic, pose-invariant properties of the object it represents—its identity, shape, and appearance.

The mathematical procedure for a single time step $t$ unfolds as follows. Starting from the absolute coordinate grid of the image patches, which is shared for each frame, $\mathbf{G}_{\text{abs}, t}$, a relative coordinate grid is computed for each slot $j$:
\begin{ceqn}\begin{equation}
\mathbf{G}^j_{\text{rel}, t} := (\mathbf{G}_{\text{abs}, t} - \mathbf{s}^j_{p,t}) \oslash \mathbf{s}^j_{s,t} \in \mathbb{R}^{N \times 2}
\label{eq:relative_grid}
\end{equation}\end{ceqn}
where $\oslash$ denotes element-wise division. This step normalizes the spatial features with respect to the slot's estimated pose. Next, unnormalized attention scores, or logits, are calculated. These scores incorporate not only the visual features $\mathbf{f}_t$ but also the pose-invariant spatial information from the relative grid:
\begin{ceqn}\begin{equation}
\mathbf{m}_t^j := \dfrac{1}{\sqrt{d}}\ p\left(k\left(\mathbf{f}_t\right) + g(\mathbf{G}^j_{\text{rel}, t})\right) q(\mathbf{s}^j_t) \in \mathbb{R}^{N}
\end{equation}\end{ceqn}
where $q, k, p,$ and $g$ are learnable linear projections. These logits form the rows of an attention matrix $\mathbf{M}_t$. A softmax function is then applied column-wise, forcing the slots to compete for ownership of each image patch and producing the final attention weights $\mathbf{A}_t$:
\begin{ceqn}\begin{equation}
\mathbf{A}_t := \underset{j=1,\dots,K}{\mathrm{softmax}} \left(\mathbf{M}_t\right) \in \mathbb{R}^{K\times N}
\end{equation}\end{ceqn}
Crucially, the pose parameters themselves are updated based on the computed attention weights $\mathbf{a}_t^j$ (the $j$-th row of $\mathbf{A}_t$). The position is calculated as the weighted mean of the absolute coordinates, and the scale is the weighted standard deviation:
\begin{ceqn}\begin{align}
\mathbf{s}^j_{p,t} &:= \dfrac{\mathbf{G}_{\text{abs}, t}^T \mathbf{a}_t^j}{\sum_{i=1}^N \mathbf{a}_t^j[i]} \in \mathbb{R}^2 \\
\mathbf{s}^j_{s,t} &:= \sqrt{\dfrac{(\mathbf{G}_{\text{abs}, t}^T - \mathbf{s}^j_p \mathbf{1}_N)^2 \mathbf{a}_t^j}{\sum_{i=1}^N \mathbf{a}_t^j[i]}} \in \mathbb{R}^2
\end{align}\end{ceqn}
Finally, the slot identity vector $\mathbf{s}_t^j$ is updated using a Gated Recurrent Unit (GRU), which integrates the aggregated feature information. This entire process is repeated for several iterations, allowing the model to converge on a stable, disentangled representation where $\mathbf{s}_t^j$ encodes "what" the object is, and $(\mathbf{s}^j_{p,t}, \mathbf{s}^j_{s,t})$ encode ``where" and ``how large" it is.

\subsubsection{Temporal Aggregation of Object-Centric and Scene-Level Representations}

While Invariant Slot Attention provides a powerful mechanism for discovering robust, pose-invariant object representations within a single frame, a video is by nature a sequence of related events. To understand a video, the model must capture relationships and motion \textit{across} frames. It is not enough to know that a car exists in frame 1 and frame 2; the model must infer that it is the \textit{same} car and that it has moved. To achieve this crucial temporal binding, our framework uses two parallel processing streams that separately gather context for individual objects and the global scene.

\textbf{Object-Level Temporal Context.} The primary focus of our encoder is to model the dynamics of individual objects. This is achieved with a \textbf{Transformer-based temporal aggregator}. After ISA has produced a set of $K$ slots for each of the $L$ frames in a sliding video window, we denote the slots for frame $t$ as $\mathbf{S}_t = \{ \mathbf{s}^j_t \}_{j=1}^K$, where each $\mathbf{s}^j_t$ corresponds to one object slot. Collecting across frames yields the sequence $\{\mathbf{S}_1, \mathbf{S}_2, \dots, \mathbf{S}_L\}$, which is then concatenated into a single long sequence and passed as input to a standard Transformer encoder architecture \cite{vaswani2017attention}, augmented with learnable temporal positional embeddings to inform the model of the sequential order of the frames. The self-attention mechanism within this Transformer allows every slot to attend to every other slot across the entire temporal window. This process enables the model to learn complex temporal relationships, such as an object's trajectory or its interactions with other entities in the scene. The output of this aggregator is a set of temporally enriched slot representations, denoted $\tilde{\mathbf{S}}$.

\begin{ceqn}\begin{equation}
\tilde{\mathbf{S}}_{1:L} = \mathrm{Transformer}(\mathbf{S}_{1:L}), \quad \mathbf{S}_{1:L} = \mathrm{concat}(\mathbf{S}_1, \dots, \mathbf{S}_L).
\end{equation}\end{ceqn}

To form the final ``video slot" that is used for conditioning the decoder, the original, per-frame slot is concatenated with its corresponding temporally-enriched version:
\begin{ceqn}\begin{equation}
\tilde{\bS}_t^+ = \text{concat}(\bS_t, \tilde{\bS}_t)
\end{equation}\end{ceqn}
This final augmented representation, $\tilde{\bS}_t^+$, contains both the fine-grained, pose-invariant details of the object in the current frame and the broader dynamic context of its motion and interactions over time.

\textbf{Scene-Level Temporal Context.} In parallel with processing object-specific information, the model must also capture the global context of the scene. To achieve this, we build on the \textbf{register token} mechanism introduced in Chapter~\ref{chap:slotadapt}. The register token served as an additional latent that aggregated global image information not captured by the object slots, ensuring that background structure and holistic cues were preserved during image generation. Extending this idea to video, we compute for each frame $t$ a context vector $\mathbf{r}_t$ by average-pooling the DINO feature vectors of that frame. This vector encapsulates scene-level properties such as the background, overall lighting, and, critically, the collective pose configuration of the objects within the frame. As will be empirically validated in a later section, incorporating this pose information into the register token is essential for enabling the generative decoder to reconstruct the scene from the pose-invariant object slots accurately.

Just like the object slots, this sequence of per-frame context vectors is processed by its own separate Transformer encoder to yield temporally-aware global scene tokens $\tilde{\mathbf{r}}_t \in \mathbb{R}^d$:
\begin{ceqn}\begin{equation}
\label{eq:register_aggregator_final_chap5}
\mathbf{r}_t = \frac{1}{N} \sum_{i=1}^{N} \mathbf{f}_{t,i}, \quad \tilde{\mathbf{r}}_t = \mathrm{Transformer}(\mathbf{r}_1, \dots, \mathbf{r}_L)_t.
\end{equation}\end{ceqn}
These global tokens, $\tilde{\mathbf{r}}_t$, summarize high-level dynamics and semantic context, providing the necessary grounding for the object-specific information contained within the primary video slots.

At the conclusion of the encoding stage, the model has successfully decomposed the video clip into two distinct types of representations for any given frame $t^\star$: a set of primary, object-centric video slots, $\tilde{\mathbf{S}}_{t^\star}^{\mathbf{+}}$, and a complementary, global context token, $\tilde{\mathbf{r}}_{t^\star}$.

\subsection{Slot-Conditioned Diffusion Decoding and Training Strategy}
\label{sec:video_decoder_final_chap5}

The decoding stage is responsible for synthesizing a photorealistic frame by interpreting the representations generated by the temporal encoder. While the encoder was fundamentally redesigned for temporal dynamics, the decoder architecture deliberately leverages the efficiency and power of the SlotAdapt design from Chapter~\ref{chap:slotadapt}. We retain the pretrained Stable Diffusion v1.5 model as our generative backbone, keeping its extensive pretrained parameters frozen to preserve its rich, learned prior of the visual world.

The process begins with a single video frame, denoted as $\mathbf{v}_{t^\star}$, randomly selected for the current training iteration. Here, $\mathbf{v}_{t^\star}$ refers to the frame in pixel space. To move the computationally intensive diffusion process into a more manageable domain, this image is passed through the encoder of a pretrained Variational Autoencoder (VAE). The encoder maps $\mathbf{v}_{t^\star}$ from the high-dimensional pixel space to a compressed latent representation, denoted as $\mathbf{x}_{t^\star} \in \mathbb{R}^{h \times w \times c}$, where $h$ and $w$ are spatially smaller than the original image dimensions. This clean latent, $\mathbf{x}_{t^\star}$, serves as the ground truth that the model will learn to recover.

Following the standard diffusion model paradigm, this latent representation is subjected to a fixed forward noising process over a sequence of $T$ timesteps. At any given timestep $\tau$, a predefined amount of Gaussian noise is added to the latent, producing a noisy version $\mathbf{x}_\tau$. The core task of the diffusion model's U-Net architecture is to learn the reverse of this process: given a noisy latent $\mathbf{x}_\tau$ at a random timestep $\tau$, the model must predict the original noise $\boldsymbol{\epsilon}$ that was added to create it. By successfully predicting and removing this noise iteratively, the model can reverse the diffusion process to generate a clean latent from pure noise.

A central component of our methodology is the highly efficient \textbf{1-frame training strategy}. The effectiveness of this strategy lies in how it guides the denoising process. The model performs the noise prediction for the single frame's latent $\mathbf{x}_\tau$ while being conditioned on the temporally-aggregated representations, $\tilde{\mathbf{S}}_{t^\star}^{\mathbf{+}}$ and $\tilde{\mathbf{r}}_{t^\star}$, which contain context from the \textit{entire} video clip. This design compels the model to develop an implicit understanding of motion and temporal causality. For instance, to correctly generate a person's arm in mid-throw at frame $t^\star$, the model must leverage the information within the video slot that encodes the arm's position and velocity from the preceding frames. Accurate frame synthesis requires leveraging temporal context from both past and future frames, which is encapsulated in the aggregated representations. This strategy effectively teaches a pretrained image generator to understand and render motion-consistent content, achieving high efficiency without the need to train a full video diffusion model from scratch.

The training objective is therefore to minimize the mean squared error between the true noise and the predicted noise, formulated as the standard diffusion loss, conditioned on our temporally-aware representations:
\begin{ceqn}
\begin{equation}
\mathcal{L}(\btheta) =
\mathbb{E}_{\mathbf{x} \sim p(\mathbf{X}),\;
\boldsymbol{\epsilon}_\tau \sim \mathcal{N}(0, \mathbf{I}),\;
\tau \sim \mathcal{U}\{1,\dots,T\}}
\left[
\left\| \boldsymbol{\epsilon}_\tau -
\epsilon_{\btheta}(\mathbf{x}_\tau, \tau, \tilde{\bS}_{t^*}^+, \tilde{\br}_{t^*}^+)
\right\|_2^2
\right]
\end{equation}
\end{ceqn}

Here, $\mathbf{x} \sim p(\mathbf{x})$ is a clean video latent sampled from the data distribution, and $\mathbf{x}_{t^\star}$ denotes the initial clean latent representation of the target frame. The noise term $\boldsymbol{\epsilon}_\tau \sim \mathcal{N}(0, \mathbf{I})$ is independently sampled for each $\tau \sim \mathcal{U}(1, T)$, and $\mathbf{x}_\tau$ is the corresponding noisy latent obtained by applying $\boldsymbol{\epsilon}_\tau$ to $\mathbf{x}_{t^\star}$ according to the diffusion schedule. The neural network $\epsilon_{\btheta}$ receives $\mathbf{x}_\tau$, the timestep $\tau$, our augmented video slots $\tilde{\bS}_{t^*}^+$, and our temporally-aware register token $\tilde{\br}_{t^*}^+$ as input to predict the added noise. This formulation enables robust training and allows the model to achieve state-of-the-art results in compositional video synthesis without excessive computational resources.

\subsection{Conditioning the Decoder with Disentangled Representations}
\label{subsec:isa_adaptation}

The effectiveness of our object-centric video generation framework relies on a careful synergy between the encoder’s representations and the conditioning mechanism of the decoder. On the encoder side, Invariant Slot Attention (ISA) disentangles object identity from pose, producing slots that are stable across motion and deformation. This separation is crucial for robust tracking but introduces a representational gap: the diffusion-based decoder must reconstruct concrete visual scenes, which requires explicit spatial grounding. Without mechanisms to reintroduce position, scale, and orientation, the slots would inevitably leak pose information or fail to yield coherent reconstructions.

In the original ISA design, this problem was addressed with a spatial broadcast decoder, which used the estimated pose parameters to restore spatial context during reconstruction. Our model replaces this simple decoder with a large, pretrained diffusion model, which brings new opportunities for realism but also new challenges: we must decide how to best supply spatial grounding to a network that was not designed for object-centric slots. To this end, we consider two different variants for conditioning our diffusion-based decoder. The first variant (V1) uses implicit pose conditioning whereas the second variant (V2) use explicit pose conditioning as explained in the sequel.

\paragraph{Implicit Pose Conditioning (V1).}
Our baseline approach, shown in Figure~\ref{fig:video_method_diagram}, leverages a \textbf{dual conditioning design}. The disentangled object slots $\tilde{\mathbf{S}}_{t^*}^+$, which encode only identity and appearance, are routed exclusively to lightweight adapter layers added to the diffusion U-Net. In parallel, a global \textbf{register token} $\tilde{\mathbf{r}}_{t^*}$ is derived from per-frame DINO features, average-pooled and temporally aggregated. This token is directed to the frozen cross-attention layers of Stable Diffusion, where it supplies background structure, global lighting, and, critically, collective pose information. This clear division of labor—slots providing “what” and the register token providing “where”—maintains disentanglement while enabling the diffusion model to place objects correctly.

\begin{figure}[H]
    \centering
    \includegraphics[width=\linewidth]{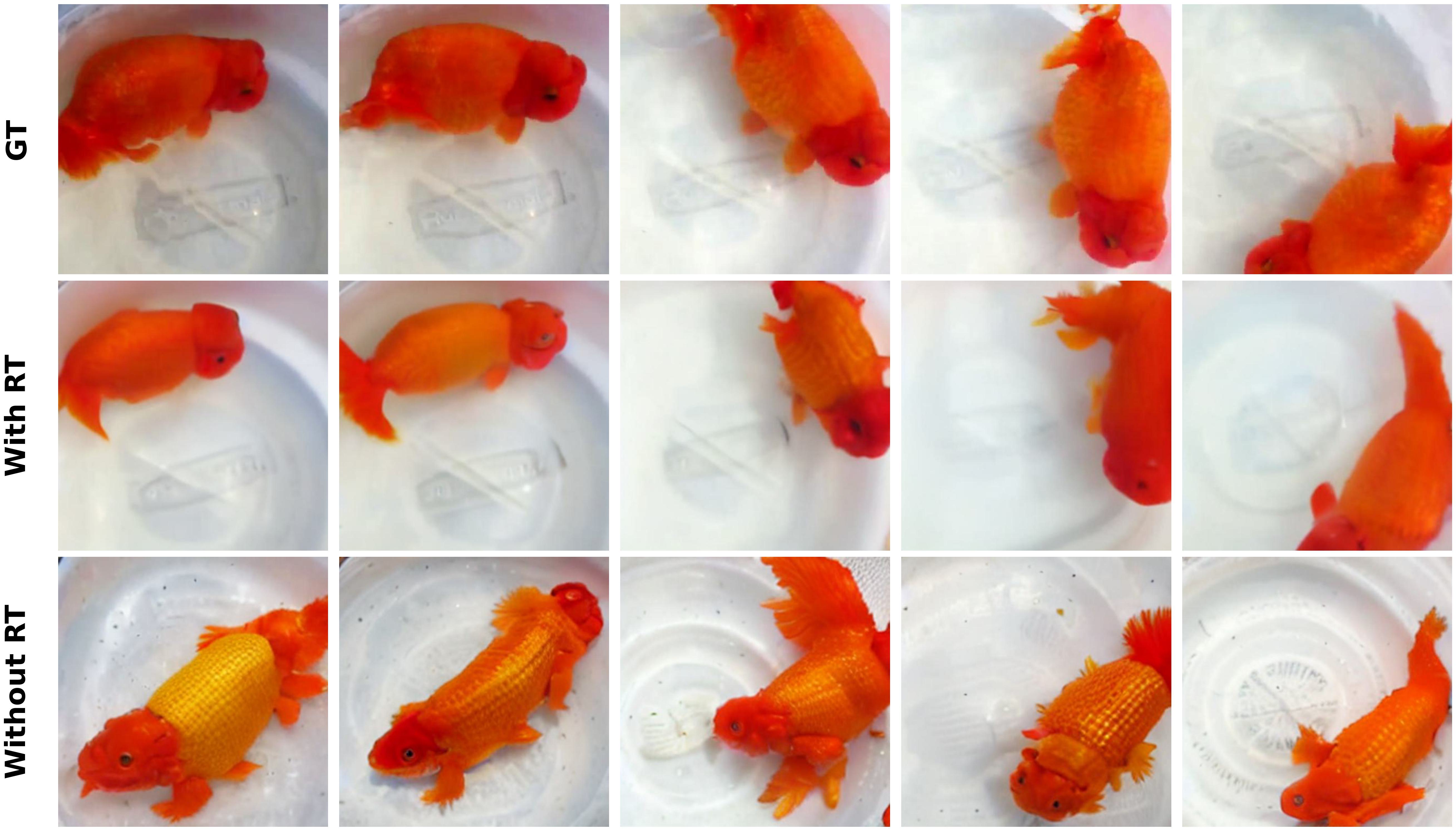}
    \includegraphics[width=\linewidth]{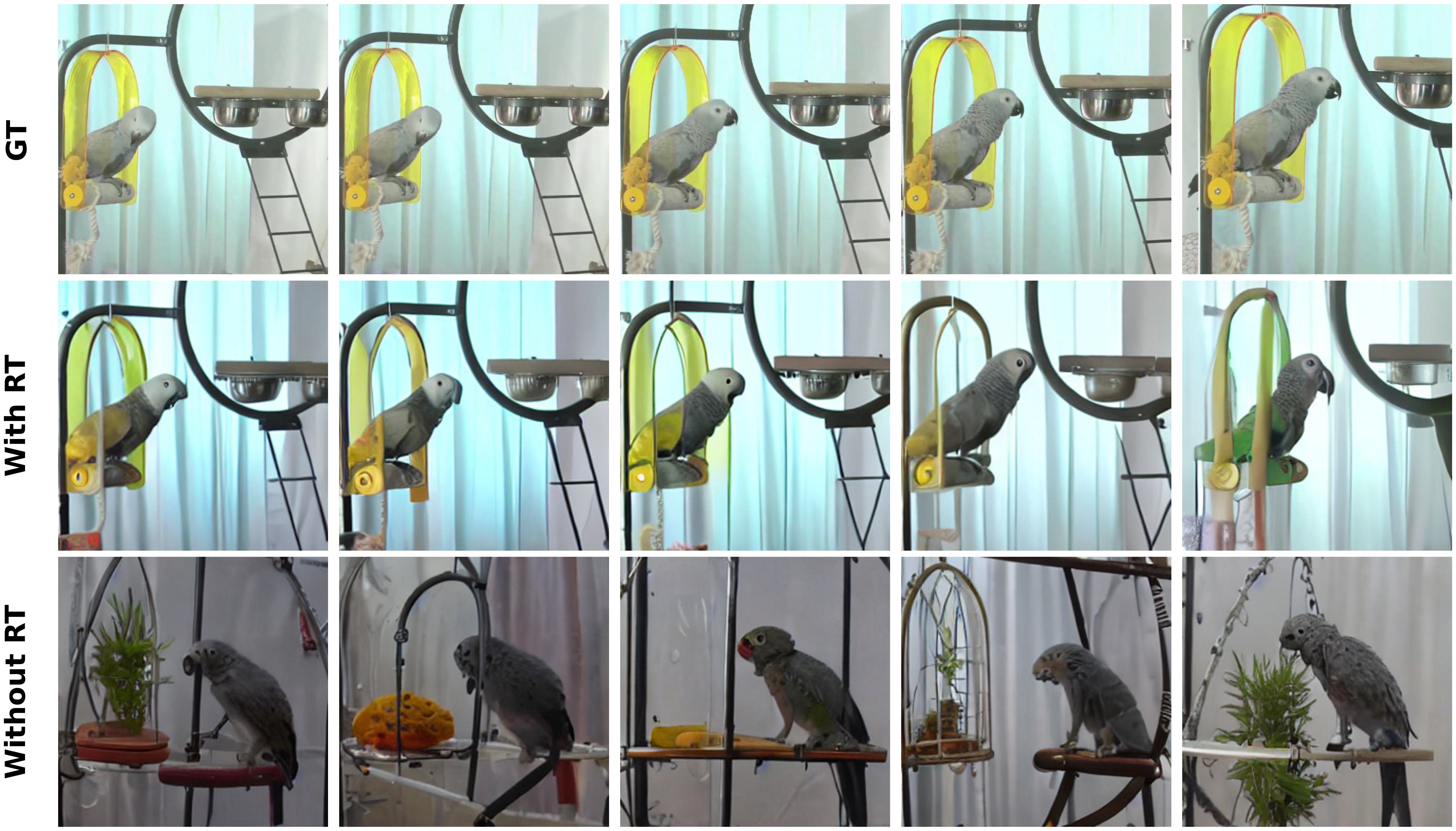}
    \caption[Register Token Effect on Generation Results]{\textbf{Effect of the register token (V1).} Top: ground truth sequence. Middle: generation \textbf{with} register token—objects are placed correctly and remain temporally consistent. Bottom: generation \textbf{without} register token—objects are synthesized but misplaced, with incorrect scales and orientations.}
    \label{fig:reg_token_vid_example}
\end{figure}

The role of the register token was empirically validated through ablations. As shown in Fig.~\ref{fig:reg_token_vid_example} and Table~\ref{table:video-gen-reg-token}, removing the register token at inference causes severe degradation in spatial coherence. Objects are still generated, since slot identity is preserved, but they appear misplaced, incorrectly scaled, or inconsistently oriented across frames. With the register token included, generated videos retain correct spatial layouts and maintain temporal consistency. This confirms that the register token provides the essential global grounding that allows pose-invariant slots to be used effectively within a diffusion framework.

\begin{figure*}[!htbp]
    \centering
    \includegraphics[width=0.95\textwidth]{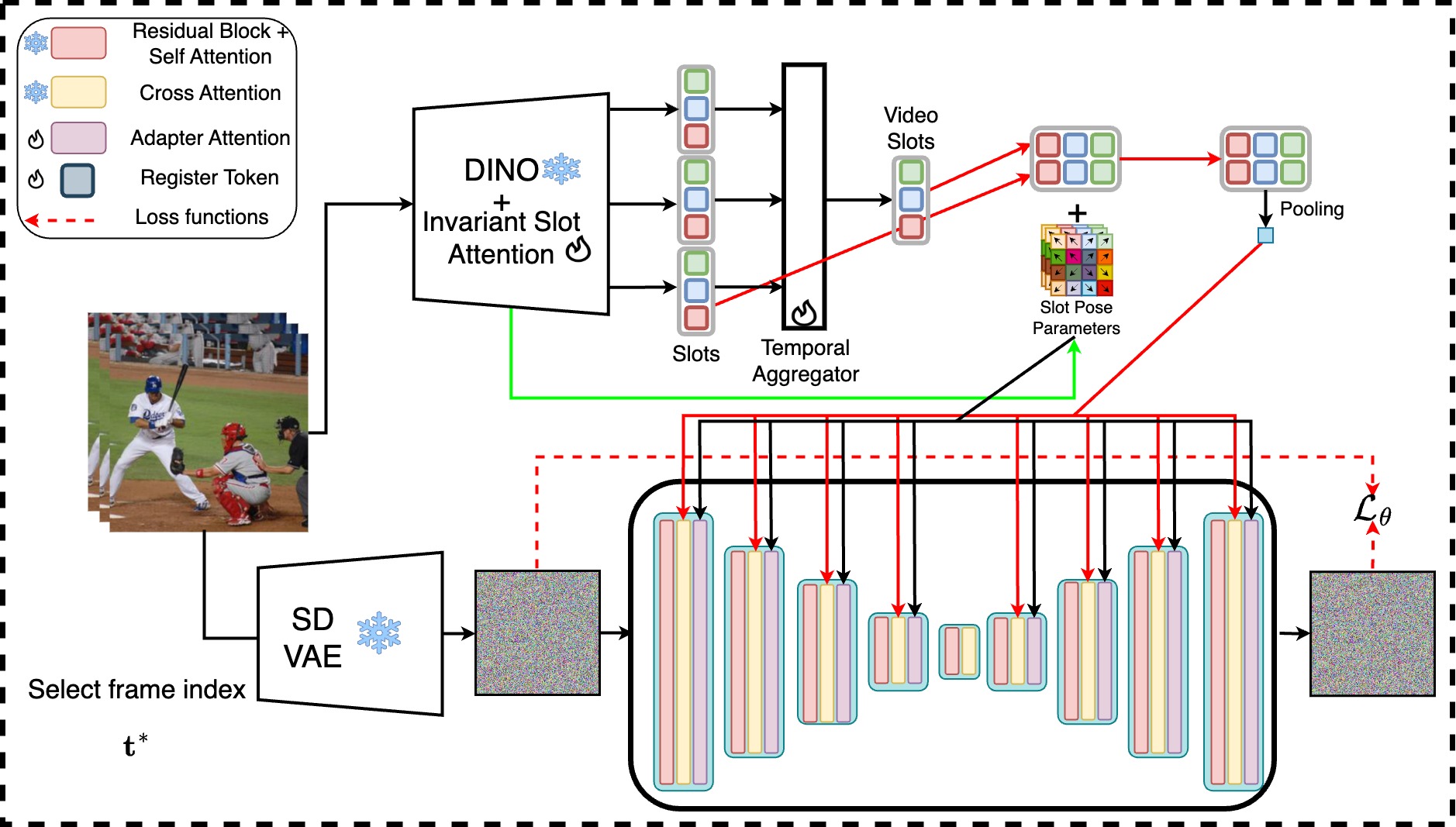}
    \caption[Explicit Pose Conditioning Architecture]{
        \textbf{Explicit Pose Conditioning (V2).}
        Overview of our extended framework with explicit pose conditioning. Each input frame is encoded by DINO and Invariant Slot Attention, producing disentangled slots. Per-slot pose parameters $(\mathbf{s}^j_p, \mathbf{s}^j_s)$ are used to construct relative coordinate grids, which are linearly projected and then \emph{broadcast} to match the dimensionality of the slot embeddings. The broadcasted pose features are added elementwise to the slots, producing pose-aware slot embeddings that capture both identity and spatial grounding. These slots are temporally aggregated and routed through adapter layers in the diffusion U-Net. In parallel, the register token is simplified to a slot average (without temporal aggregation) and fed into the frozen cross-attention layers. This architecture enables the decoder to directly exploit explicit pose information, improving object placement and compositional editing compared to the implicit variant (Figure~\ref{fig:video_method_diagram}).}
    \label{fig:video_pose_extension_diagram}
\end{figure*}

\paragraph{Explicit Pose Conditioning (V2).}
Although implicit conditioning enables strong results, its reliance on a global register token sometimes limits fine-grained editing. For example, during compositional replacement, implicit pose conditiong (V1) may generate the correct object but at the wrong position or scale. To address this, we introduce another variant (V2) based on explicit pose conditioning (Figure~\ref{fig:video_pose_extension_diagram}) that fuses per-slot pose directly into the slot embeddings before decoding.

ISA provides pose parameters $(\mathbf{s}^j_{p,t}, \mathbf{s}^j_{s,t})$ for each slot $j$, corresponding to translation and scale. Using these, we construct relative coordinate grids:
\begin{ceqn}
\begin{equation}
\mathbf{G}^j_{\text{rel}, t} = \frac{\mathbf{G}_{\text{abs}, t} - \mathbf{s}^j_{p,t}}{\mathbf{s}^j_{s,t}} \in \mathbb{R}^{N \times 2},
\end{equation}\end{ceqn}
where $\mathbf{G}_{\text{abs}, t}$ are absolute patch coordinates and $N$ is the number of spatial locations. These grids normalize coordinates with respect to each slot’s pose. 

To fuse pose and identity, we combine the slot embeddings with the projected relative grids as follows:
\begin{ceqn}
\begin{align}
\mathbf{\hat{G}}^j_{\text{rel},t} &= h(\mathbf{G}^j_{\text{rel},t}) \in \mathbb{R}^{N \times 1 \times D}, \\
\mathbf{S}_{\mathrm{b},t} &= \mathrm{broadcast}(\mathbf{S}_t) \in \mathbb{R}^{N \times K \times D}, \\
\mathbf{B}_t &= \mathbf{S}_{\mathrm{b},t} + \mathbf{\hat{G}}^j_{\text{rel},t} \in \mathbb{R}^{N \times K \times D}, \\
\mathbf{P}_t &= \tfrac{1}{N}\sum_{i=1}^N \mathbf{B}_t[i] \in \mathbb{R}^{K \times D}, \\
\mathbf{s}_t^j &= \mathrm{ReLU}\!\big(\mathrm{LayerNorm}(\mathbf{p}_t^j)\big) \in \mathbb{R}^D, \\
\mathbf{S}_t &= \{\mathbf{s}_t^j\}_{j=1}^K \in \mathbb{R}^{K \times D}, \\
\tilde{\mathbf{S}}_{1:L} &= \mathrm{Transformer}(\mathbf{S}_{1:L}), \quad 
\mathbf{S}_{1:L} = \mathrm{concat}(\mathbf{S}_1, \dots, \mathbf{S}_L) \in \mathbb{R}^{L \times K \times D}.
\end{align}\end{ceqn}

Here, $h(\cdot)$ is a linear projector mapping relative grids into the slot embedding space, and $\mathrm{broadcast}(\cdot)$ denotes tiling a slot embedding to match the spatial resolution $N$. The sum $\mathbf{B}_t$ merges slot appearance with relative position, which is then pooled across spatial dimensions to yield pose-aware slots $\mathbf{S}_t$. Finally, a temporal transformer aggregates slots across $L$ frames into $\tilde{\mathbf{S}}_{1:L}$.

In this design, the global register token is simplified: instead of being derived from DINO features and temporally aggregated, it is computed as a simple average of slots (before pose fusion):
\begin{ceqn}
\begin{equation}
\mathbf{r}_t = \tfrac{1}{K}\sum_{j=1}^K \tilde{\mathbf{s}}^j_t.
\end{equation}\end{ceqn}
This minimal token is passed to the frozen cross-attention layers, while the pose-aware slots condition the adapters. The result is a decoder that directly exploits pose parameters for spatial placement, enabling more precise and controllable editing.

\subsection{Comparative Discussion: Implicit vs. Explicit Conditioning}
The two conditioning strategies represent complementary solutions to bridging the gap between pose-invariant discovery and spatially grounded synthesis:
\begin{itemize}
    \item \textbf{Implicit (V1):} Separates identity (slots) from pose and background (register token). Simpler, robust to noise, and validated by ablations, but limited in precise spatial control during editing.
    \item \textbf{Explicit (V2):} Directly fuses pose into slots, simplifying the register token. Provides stronger control over object placement and improves editing reliability, but depends on accurate pose estimates and may be more brittle when these are noisy.
\end{itemize}
Together, these variants highlight two distinct pathways for adapting pose-invariant slots to diffusion-based synthesis: one emphasizing robustness, the other emphasizing controllability.

\section{Experimental Evaluation}
\label{sec:video_experiments}

We conducted a comprehensive set of experiments to rigorously evaluate our framework on the core tasks of unsupervised video object segmentation and temporally consistent video generation. We use our implicit pose conditioning version (V1) for all the results unless stated otherwise.
\begin{table}[b]
  \caption[Register Token Effect on Generation]{\textbf{Register token effect on video generation performance on YTVIS.} Register-token (RT) ablation.  
  The RT clearly boosts pixel accuracy (PSNR, SSIM), perceptual quality (LPIPS, FID) and temporal coherence (FVD) by carrying the pose information.}
  \label{table:video-gen-reg-token}
  \centering
  \small
  \setlength{\tabcolsep}{6pt}
  \begin{tabular}{lccccc}
    \toprule
    \textbf{Method} & PSNR$\uparrow$ & SSIM$\uparrow$ & LPIPS$\downarrow$ & FID$\downarrow$ & FVD$\downarrow$ \\
    \midrule
    w/o RT & 9.90 & 0.28 & 0.69 & 85.0 & 103.0 \\
    w/ RT  & \textbf{11.37} & \textbf{0.3933} & \textbf{0.5908} & \textbf{49.51} & \textbf{51.77} \\
    \bottomrule
  \end{tabular}
  \vspace{-0.5cm}
\end{table}

\subsection{Setup}

\textbf{Datasets.} Our evaluation was performed on two widely used real-world video benchmarks: \textbf{DAVIS17} \cite{perazzi2016benchmark}, containing high-quality sequences specifically tailored for video object segmentation, and \textbf{YouTube-VIS 2019} \linebreak \textbf{(YTVIS19)} \cite{Xu2018ECCV}, which features a greater diversity of complex scenes with significant variations in object appearance, pose, and background.

\textbf{Baselines.} For the segmentation task, we compare our model against \textbf{SOLV} \cite{aydemir2023self}, a recent state-of-the-art method in this area. It is important to note that SOLV uses a decoder that is specialized for mask generation and lacks generative capabilities. For the video generation task, as no prior methods directly address object-centric video generation from unsupervised representations, we establish a rigorous baseline. We compare against leading object-centric image models—\textbf{LSD} \cite{jiang2023lsd}, \textbf{SlotDiffusion} \cite{wu2023slotdiffusion}, and our own \textbf{SlotAdapt} \cite{akan2025slot-guided} from Chapter~\ref{chap:slotadapt} by training them on individual video frames. This comparison effectively highlights our framework's unique ability to generate coherent videos by explicitly modeling temporal dependencies.

\textbf{Implementation Details.} We use a frozen DINOv2 ViT-B/14 as the visual backbone and Stable Diffusion v1.5 as the decoder. All models are trained for 350,000 iterations on YTVIS19 and then fine-tuned for 50,000 iterations on DAVIS17, using a temporal window of $L=5$ frames (2 past, 1 present, 2 future). Only the ISA module, the temporal transformers, and the lightweight adapter layers are optimized during training. All experiments were conducted on two NVIDIA A40 GPUs.

\textbf{Evaluation Metrics.} We employ a comprehensive suite of metrics for both tasks.
\begin{itemize}
    \item \textbf{Segmentation:} We use the Foreground Adjusted Rand Index (FG-ARI) to measure the quality of clustering foreground pixels into distinct object segments. We also use the standard mean Intersection-over-Union (mIoU) to measure the pixel-level overlap accuracy for foreground objects. To ensure temporal consistency in evaluation, we apply the Hungarian matching algorithm \cite{kuhn1955hungarian} between predicted and ground-truth masks.
    \item \textbf{Generation:} To assess quality, we use five complementary metrics. Peak Signal-to-Noise Ratio (PSNR) and the Structural Similarity Index (SSIM) \cite{Wang2004TIP} are used to quantify pixel-level fidelity and structural preservation. Learned Perceptual Image Patch Similarity (LPIPS) \cite{Zhang2018CVPR} and Fréchet Inception Distance (FID) \cite{heusel2017fid} are used to measure perceptual quality and distributional similarity to real data. Crucially, we also employ Fréchet Video Distance (FVD) \cite{Unterthiner2018arXiv} to specifically assess temporal consistency and motion quality in the generated video sequences.
\end{itemize}

\subsection{Ablation Studies}
To validate our key architectural and methodological choices, we conducted a series of ablation studies on the YTVIS dataset, with results summarized in Table \ref{table:ablation-study}. The results clearly demonstrate the impact of each component. Using standard slot attention instead of ISA causes a drastic reduction in performance (mIoU drops from 40.57 to 27.09), confirming that learning pose-invariant representations is critical for the temporal domain. Furthermore, removing the temporal register aggregator or replacing our DINO-based register token with a simpler slot-averaging scheme leads to significant declines in both segmentation metrics. This underscores the importance of these components for capturing dynamic scene context effectively. Finally, we confirmed that our default 1-frame training strategy provides the best trade-off between performance and computational efficiency, as it yields results nearly on par with a more costly 5-frame training approach but with significantly lower overhead.

\begin{table}[t]
    \caption[Ablation study on YTVIS dataset]{\textbf{Ablation study on YTVIS dataset.} We systematically evaluate the contribution of key components in our unified framework by comparing against our full model configuration. The full model uses invariant slot attention, DINO register tokens, register aggregator, and 1-frame training. We analyze the impact of removing individual components and varying training strategies. \vspace{-0.3cm}
    }
    \label{table:ablation-study}
    \centering
    \small
    \setlength{\tabcolsep}{8pt}
    \begin{tabular}{lcc}
    \toprule
    \textbf{Method Configuration} & \textbf{mIoU} & \textbf{FG-ARI} \\
    \midrule
    Full Model & 40.57 & 22.40 \\
    \midrule
    \multicolumn{3}{l}{\textit{Component Ablations}} \\
    w/o Register Aggregator & 39.22 & 20.49 \\
    w/ Slot Avg Register Tokens & 36.87 & 18.06 \\
    w/ Standard Slot Attention & 27.09 & 11.06 \\
    \midrule
    \multicolumn{3}{l}{\textit{Training Strategy Ablations}} \\
    5-frame Training & 40.67 & 22.00 \\
    5-frame Training + Guidance & 41.02 & 22.69 \\
    \bottomrule
    \end{tabular}
    \vspace{-0.65cm}
\end{table}

\FloatBarrier
\subsection{Comparison to Baselines}

\subsubsection{Unsupervised Video Object Segmentation}

\textbf{Quantitative Results.} The quantitative evaluation of our framework on unsupervised video object segmentation is presented in Table~\ref{table:video-seg-quan}. Our unified model demonstrates highly competitive performance against contemporary methods, particularly when architectural distinctions are considered.

A crucial point of comparison involves the architectural divergence from the \textbf{SOLV} benchmark~\cite{aydemir2023self}. Our model is designed as a unified framework capable of both perception and generation tasks, while SOLV is a specialized method with a decoder architecture optimized specifically for segmentation mask prediction. This architectural difference manifests in how segmentation masks are produced. Our framework, by design, derives masks directly from the \textbf{encoder's} attention mechanism, reflecting the model's intrinsic object discovery process. Conversely, the primary results reported for SOLV are generated by its specialized \textbf{decoder}, which is trained explicitly to refine these initial representations into high-fidelity masks.

\begin{figure*}[htbp]
    \centering
    \includegraphics[width=0.9\linewidth]{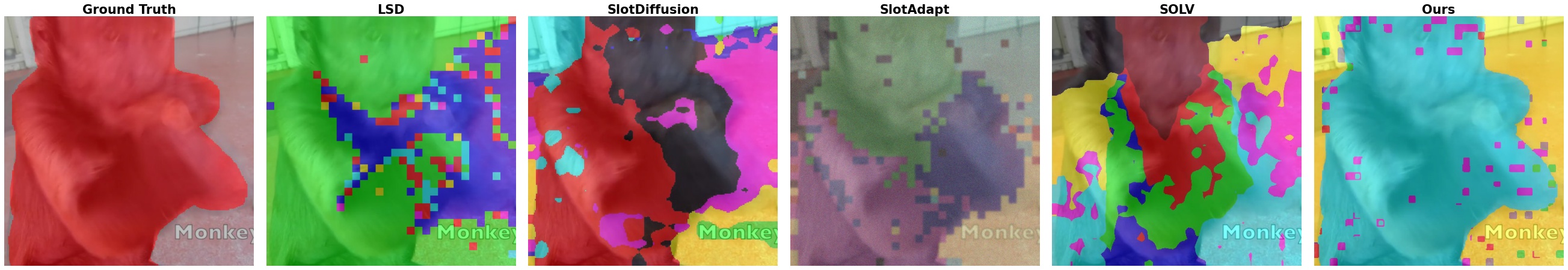}
    \includegraphics[width=0.9\linewidth]{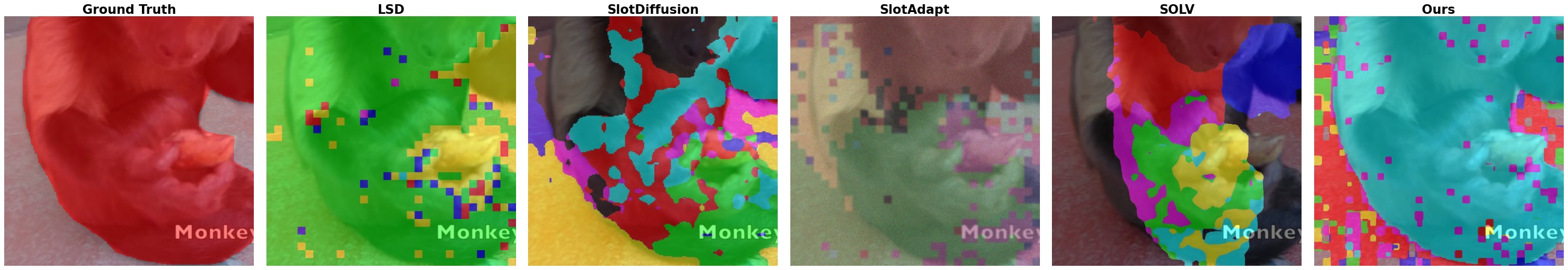}
    \includegraphics[width=0.9\linewidth]{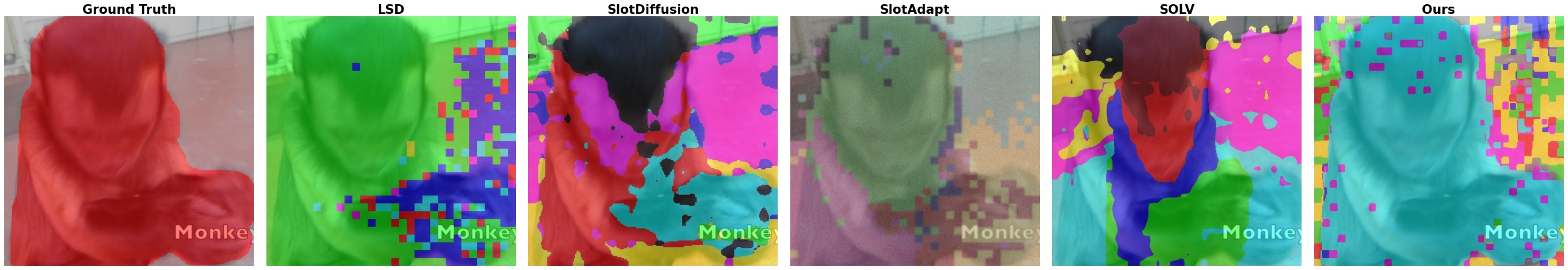}
    \includegraphics[width=0.9\linewidth]{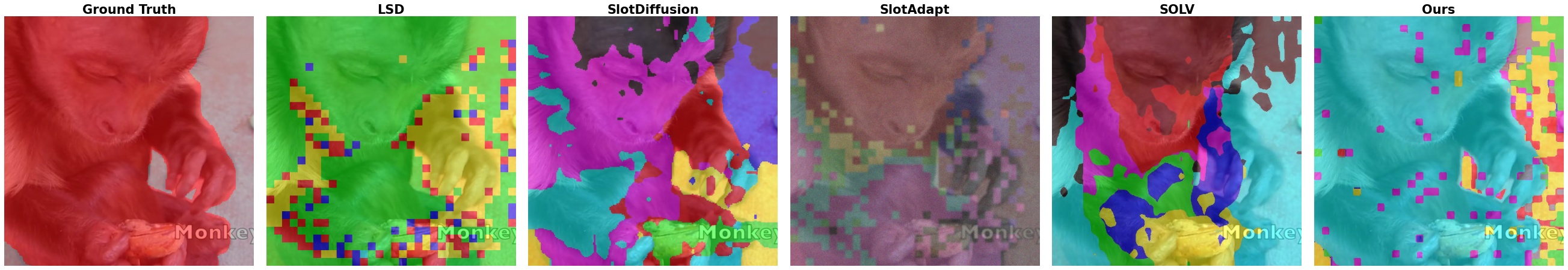}
    \includegraphics[width=0.9\linewidth]{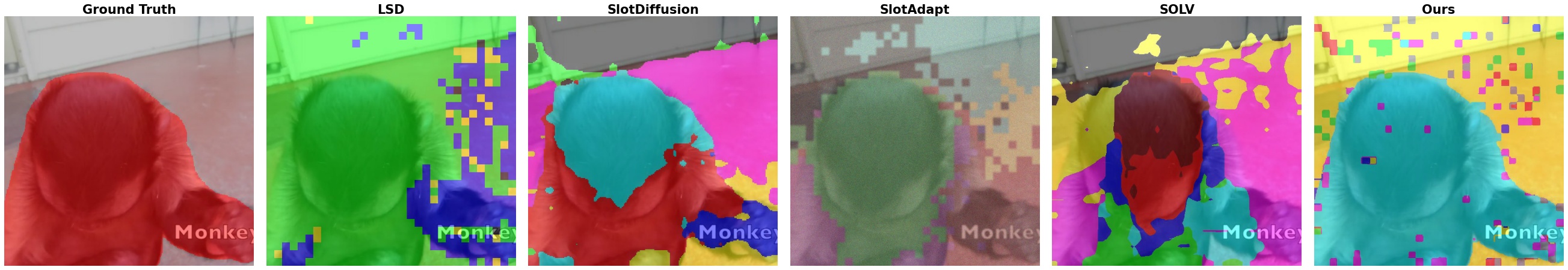}
    \caption[Segmentation Results on YTVIS]{\textbf{Segmentation Results on YTVIS.} Our method accurately delineates and tracks large objects undergoing significant shape and appearance changes across consecutive frames.}
    \label{fig:seg-comparison-ytvis-26}
\end{figure*}

\begin{figure*}[htbp]
    \centering
    \includegraphics[width=0.9\linewidth]{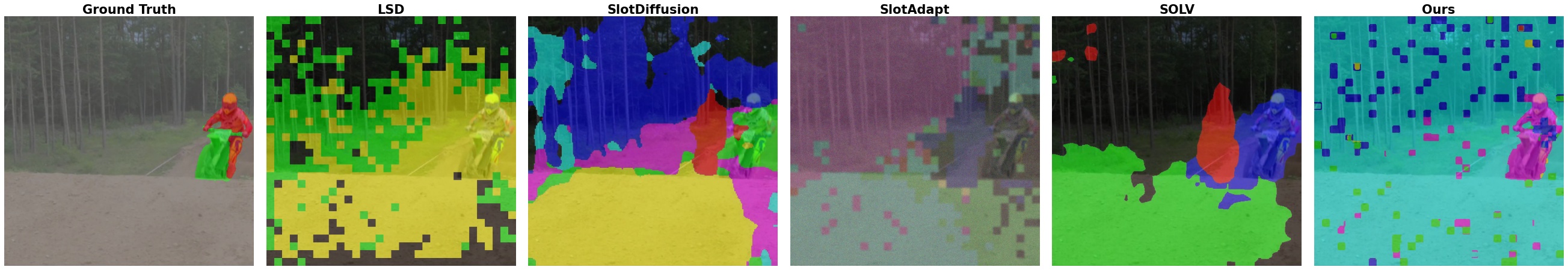}
    \includegraphics[width=0.9\linewidth]{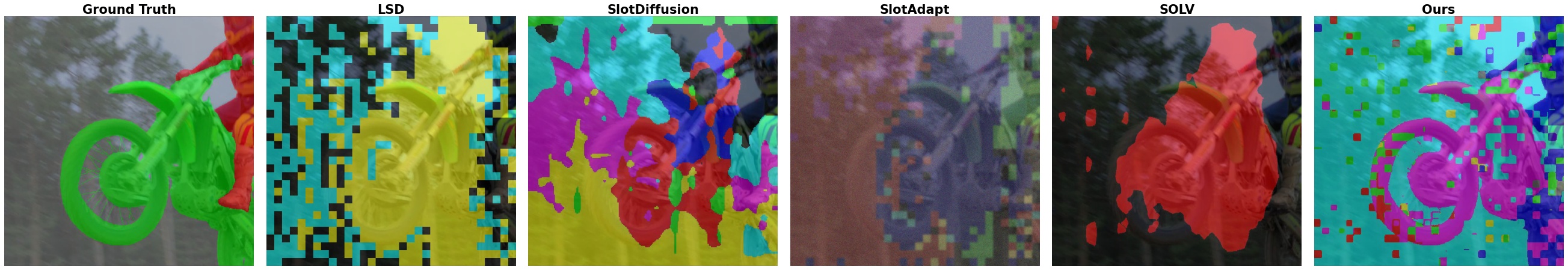}
    \includegraphics[width=0.9\linewidth]{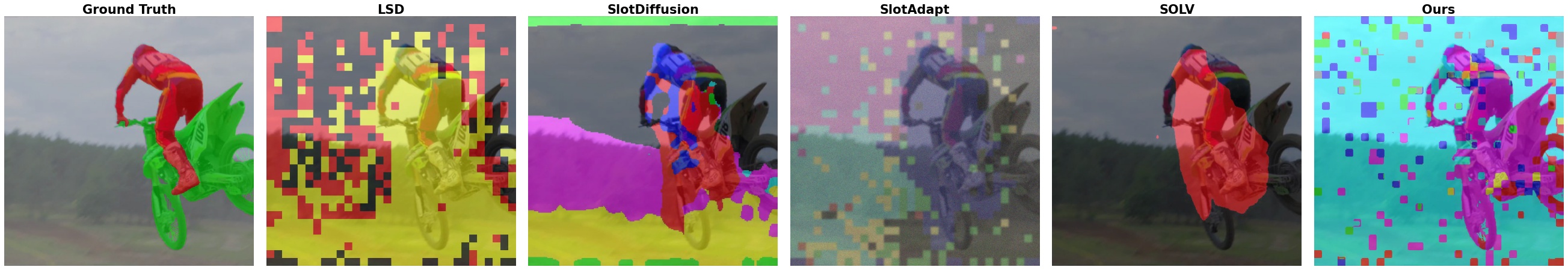}
    \includegraphics[width=0.9\linewidth]{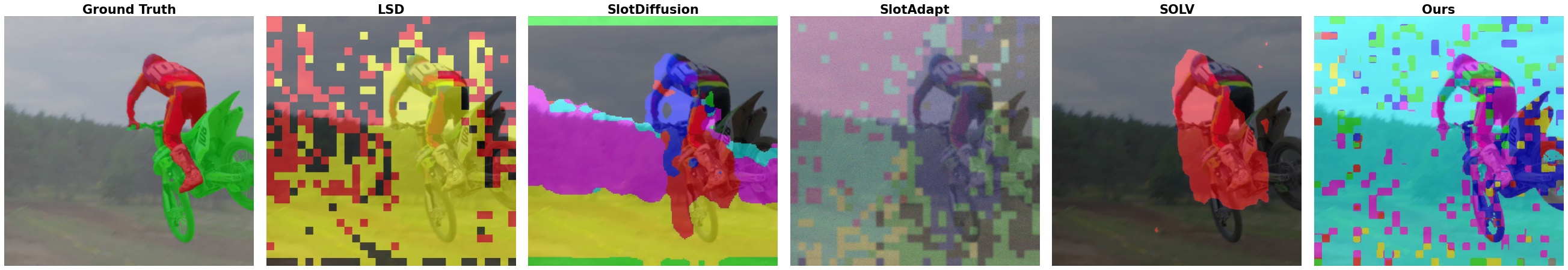}
    \includegraphics[width=0.9\linewidth]{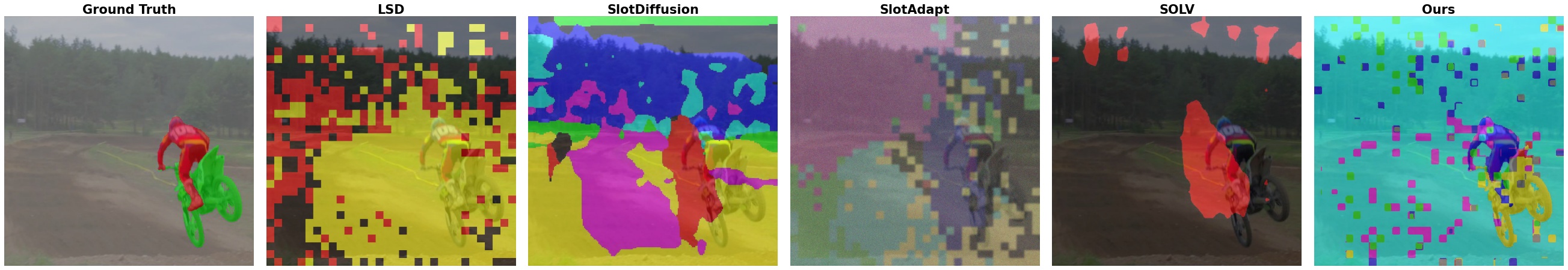}
    \caption[Segmentation Results on DAVIS17]{\textbf{Segmentation Results on DAVIS17.} Our method successfully distinguishes and tracks multiple closely positioned objects while maintaining stable segmentation masks across the entire temporal sequence.}
    \label{fig:seg-comparison-davis-23}
\end{figure*}

To facilitate a more direct comparison of the underlying representation quality, we therefore include results from SOLV's encoder-level masks (\textbf{SOLV-E}) in our evaluation. The results indicate that our method achieves state-of-the-art performance on the clustering-based metric (FG-ARI), surpassing all baselines including the full SOLV model. On the overlap-based metric (mIoU), our approach remains highly competitive with the fully-specialized SOLV decoder and substantially outperforms SOLV's own encoder-derived masks. This strong performance, attained without a specialized segmentation head, underscores the high quality of the object representations learned by our encoder and validates that our unified framework can perform perception at a level competitive with specialized methods, while uniquely offering powerful generative and compositional editing capabilities. \\
\textbf{Qualitative Analysis.} Figures \ref{fig:seg-comparison-ytvis-26} and \ref{fig:seg-comparison-davis-23} provide a comprehensive qualitative evaluation of our method's unsupervised video object segmentation performance on the YTVIS and DAVIS17 datasets, respectively. Each figure sequence displays five consecutive frames with segmentation masks overlaid, where different colors represent distinct object instances automatically discovered by our model. These results demonstrate our framework's ability to maintain consistent object identity tracking while preserving accurate object boundaries under various challenging conditions. \footnotetext[1]{The original SOLV results were reported at a higher resolution ($336\times504$) with a different aspect ratio, while our experiments are conducted at $224\times224$. This resolution difference partially accounts for the observed performance variations.}
\begin{table}[t]
    \caption[Segmentation Performance on Real-world Datasets]{\textbf{Unsupervised video object segmentation on real-world datasets.} We compare our method with state-of-the-art approaches on YTVIS and DAVIS datasets. For fair comparison with our encoder-based approach, we include SOLV-E (encoder attention masks) and SOLV-E + M (encoder attention masks with merging), alongside the full SOLV method which uses decoder-generated masks. Our approach demonstrates strong performance across both clustering-based (FG-ARI) and overlap-based (mIoU) evaluation metrics. \vspace{-0.3cm}
    }
    \label{table:video-seg-quan}
    \centering
    \small
    \setlength{\tabcolsep}{4pt}
    \begin{tabular}{lcccc}
    \toprule
    \multirow{2}{*}{\textbf{Method}} & \multicolumn{2}{c}{\textbf{YTVIS}} & \multicolumn{2}{c}{\textbf{DAVIS}} \\
    \cmidrule(lr){2-3} \cmidrule(lr){4-5}
    & mIoU & FG-ARI & mIoU & FG-ARI \\
    \midrule
    LSD~\cite{jiang2023lsd} & 29.55 & 14.07 & 29.55 & 14.35 \\
    SlotDiffusion~\cite{wu2023slotdiffusion} & 38.33 & 15.70 & 31.27 & 12.34 \\
    SlotAdapt~\cite{akan2025slot-guided} & 36.51 & 20.32 & 29.95 & 16.28 \\
    \midrule
    SOLV-E & 32.91 & 19.30 & 31.23 & 18.89 \\
    SOLV-E + M & 36.91 & 21.34 & 33.12 & 20.40 \\ 
    SOLV~\cite{aydemir2023self}\footnotemark[1] & \textbf{42.01} & 21.55 & \textbf{36.62} & 20.98 \\
    \midrule
    Ours & 40.57 & \textbf{22.40} & 34.93 & \textbf{21.60} \\
    \bottomrule
    \end{tabular}
    \vspace{-0.62cm}
\end{table}

Our method successfully handles several difficult scenarios: (1) rapid object motion and significant deformation across frames, (2) objects with high visual similarity or close spatial proximity, (3) substantial scale changes and partial occlusions, and (4) complex interactions between multiple objects in the scene. The segmentation masks remain stable and coherent across temporal sequences, with objects maintaining their assigned identities throughout the video duration. In contrast, baseline methods frequently exhibit segmentation instability, object identity confusion, or progressive degradation of boundary accuracy across frames.

\FloatBarrier

\subsubsection{Video Generation and Reconstruction}

\begin{figure*}[htbp]
    \centering
    \includegraphics[width=0.9\linewidth]{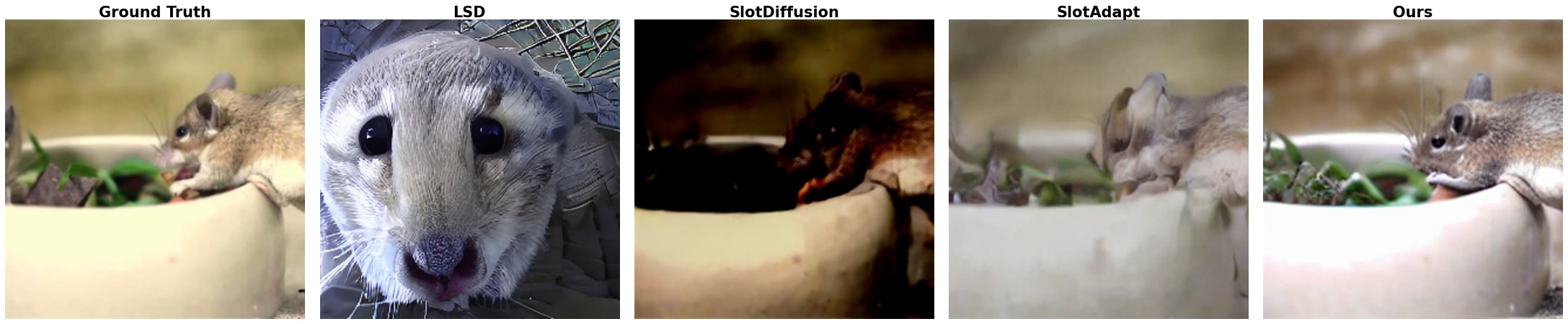}
    \includegraphics[width=0.9\linewidth]{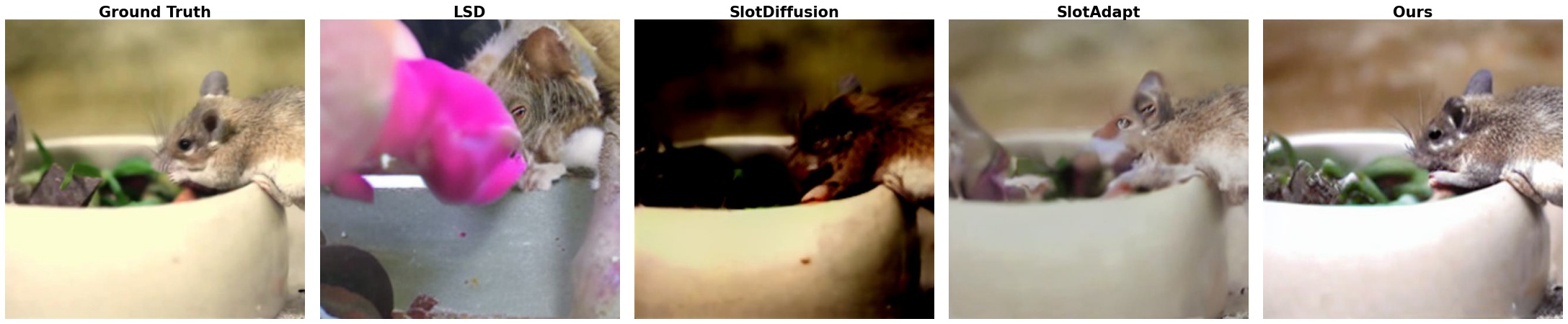}
    \includegraphics[width=0.9\linewidth]{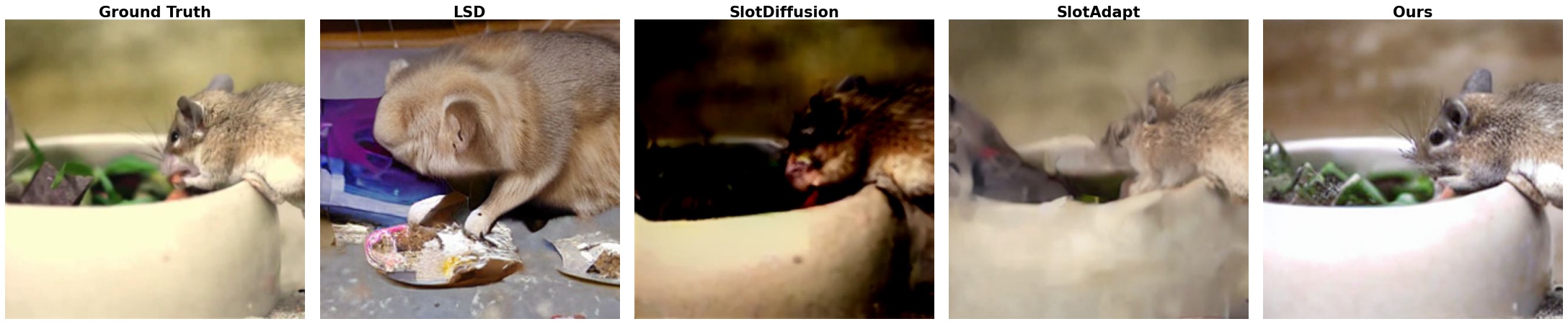}
    \includegraphics[width=0.9\linewidth]{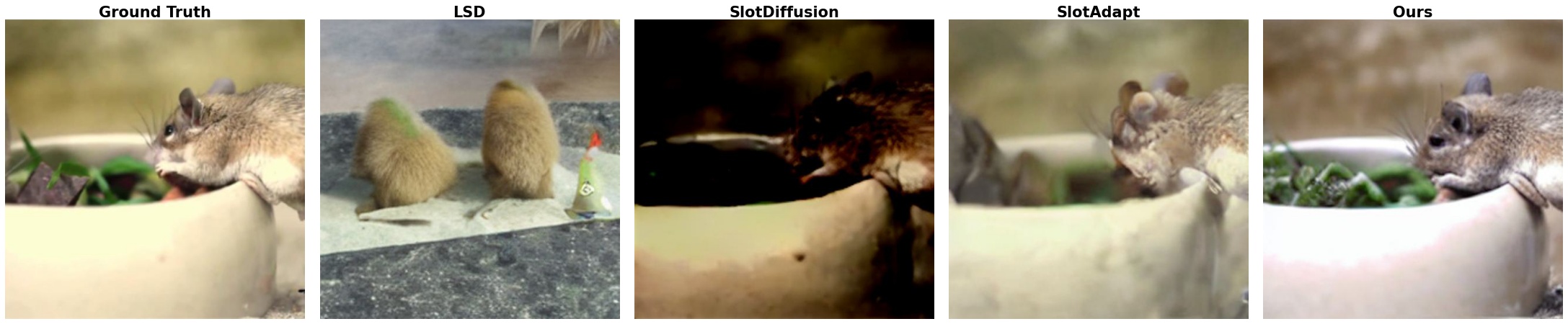}
    \includegraphics[width=0.9\linewidth]{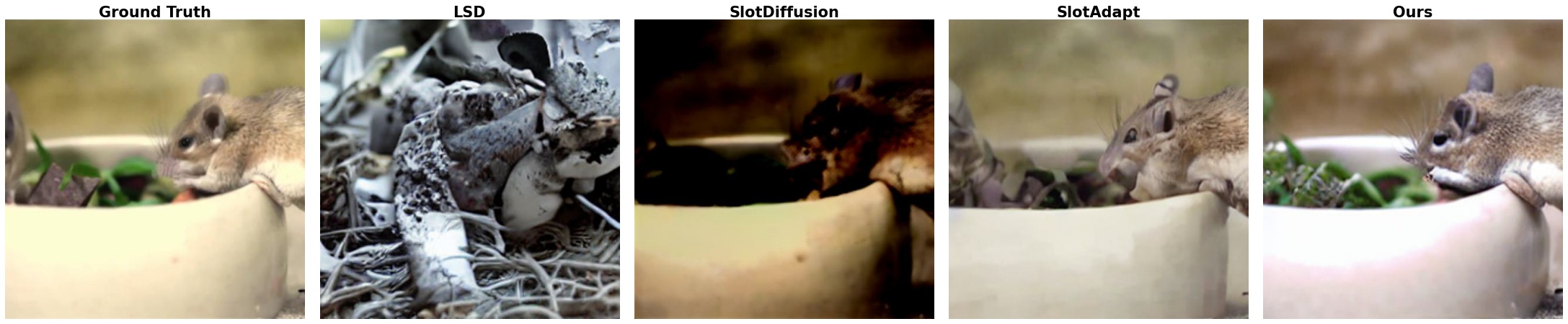}
    \caption[Video Generation Results on YTVIS]{\textbf{Video Generation Results on YTVIS.} Our method demonstrates superior object identity preservation and spatial coherence compared to baseline methods across five consecutive generated frames.}
    \label{fig:gen-comparison-ytvis-59}
\end{figure*}

\begin{figure*}[htbp]
    \centering
    \includegraphics[width=0.9\linewidth]{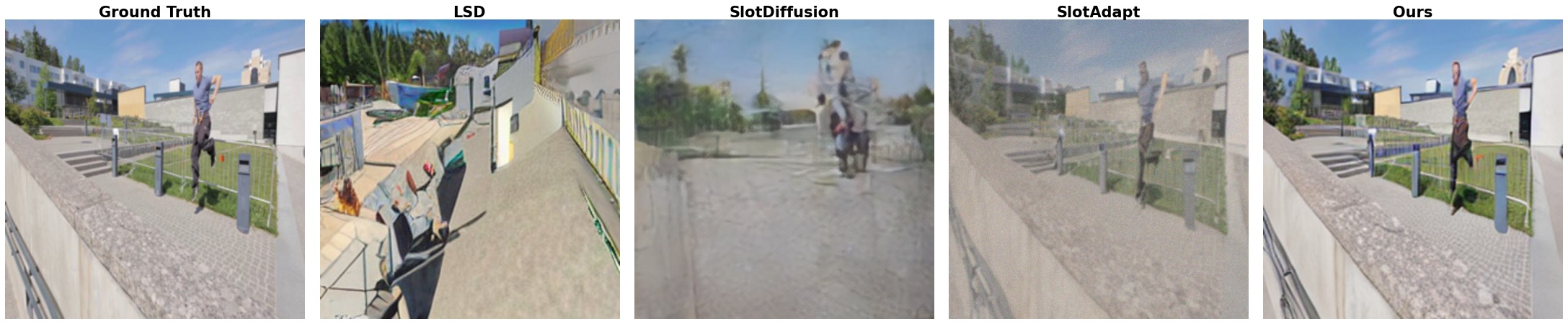}
    \includegraphics[width=0.9\linewidth]{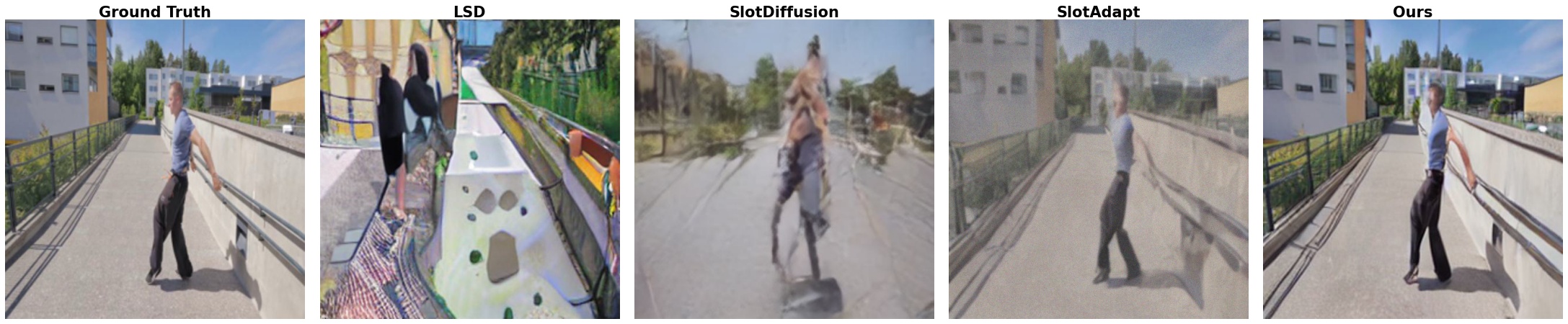}
    \includegraphics[width=0.9\linewidth]{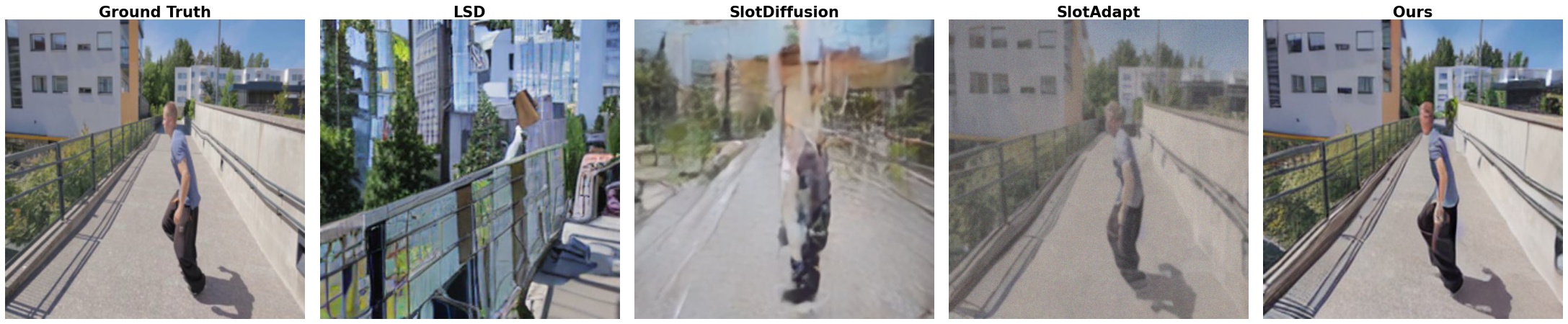}
    \includegraphics[width=0.9\linewidth]{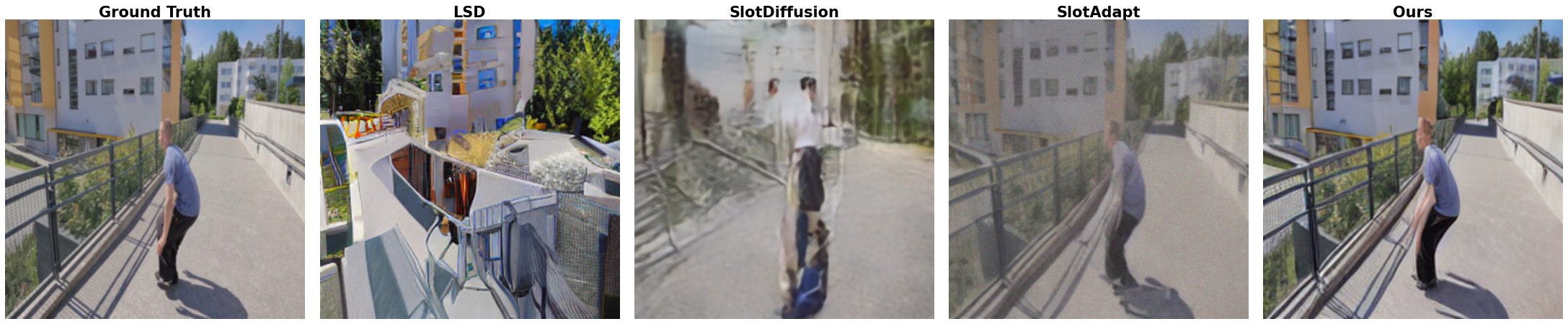}
    \includegraphics[width=0.9\linewidth]{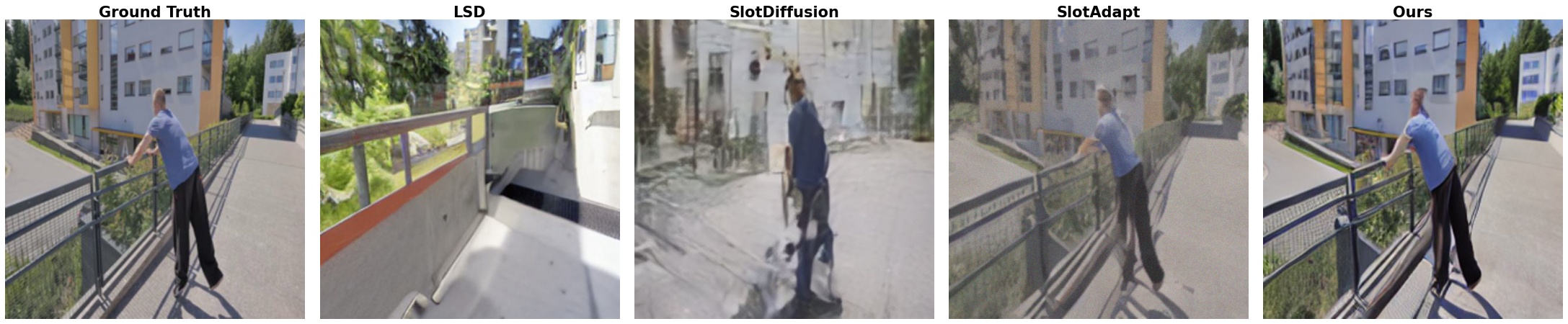}
    \caption[Video Generation Results on DAVIS17]{\textbf{Video Generation Results on DAVIS17.} Our approach maintains consistent object motion tracking and spatial layout coherence throughout the temporal sequence.}
    \label{fig:gen-comparison-davis-25}
\end{figure*}

\textbf{Quantitative Results.} The video generation performance is evaluated quantitatively in Table \ref{table:video-gen-quan}, which demonstrates our model's superior capabilities in video synthesis tasks. Our approach achieves state-of-the-art results on YTVIS and DAVIS17, outperforming baseline object-centric models across all metrics. The improvements in Fréchet Video Distance (FVD) are particularly notable, indicating superior temporal coherence and motion quality in our generated videos compared to baseline methods. Since FVD directly measures these temporal aspects, the substantial improvements provide strong empirical validation for the effectiveness of our temporal modeling architecture and training approach. \\
\textbf{Qualitative Analysis.} Figures \ref{fig:gen-comparison-ytvis-59} and \ref{fig:gen-comparison-davis-25} present detailed qualitative comparisons of our video generation results against baseline methods on the YTVIS and DAVIS17 datasets. Each figure sequence shows five consecutive frames generated by different methods, allowing for direct visual comparison of temporal consistency and visual quality. Our model produces consistently higher-quality video frames that exhibit improved temporal coherence compared to baseline approaches, which frequently suffer from flickering artifacts, inconsistent object appearances, or progressive degradation of object structures over time.

\begin{table*}[t]
    \caption[Generation Performance on Real-world Datasets]{\textbf{Video generation performance on real-world datasets.} We evaluate our method against state-of-the-art approaches on YTVIS and DAVIS datasets using comprehensive generation metrics. Our approach demonstrates superior performance across both pixel-level accuracy (PSNR, SSIM), perceptual quality measures (LPIPS, FID) and temporal coherence (FVD).\vspace{-0.3cm}}
    \label{table:video-gen-quan}
    \centering
    \small
    \setlength{\tabcolsep}{6pt}
    \begin{tabular}{lccccc}
    \toprule
    \textbf{Method} & PSNR$\uparrow$ & SSIM$\uparrow$ & LPIPS$\downarrow$ & FID$\downarrow$ & FVD$\downarrow$ \\
    \midrule
    \multicolumn{6}{c}{\textbf{YTVIS}} \\
    \midrule
    LSD~\cite{jiang2023lsd} & 9.64 & 0.2793 & 0.7770 & 100.68 & 121.41 \\
    SlotDiffusion~\cite{wu2023slotdiffusion} & 9.18 & 0.1867 & 0.6484 & 86.38 & 123.84 \\
    SlotAdapt~\cite{akan2025slot-guided} & 10.92 & 0.3669 & 0.6556 & 65.30 & 63.72 \\
    \midrule
    Ours & \textbf{11.37} & \textbf{0.3933} & \textbf{0.5908} & \textbf{49.51} & \textbf{51.77} \\
    \midrule
    \multicolumn{6}{c}{\textbf{DAVIS}} \\
    \midrule
    LSD~\cite{jiang2023lsd} & 9.58 & 0.0356 & 0.6079 & 84.30 & 75.05 \\
    SlotDiffusion~\cite{wu2023slotdiffusion} & 10.68 & 0.0340 & 0.6143 & 136.18 & 152.73 \\
    SlotAdapt~\cite{akan2025slot-guided} & 12.18 & 0.0674 & 0.2681 & 41.94 & 29.96 \\
    \midrule
    Ours & \textbf{12.38} & \textbf{0.0946} & \textbf{0.1886} & \textbf{28.43} & \textbf{16.17} \\
    \bottomrule
    \end{tabular}
\end{table*}

The generated sequences demonstrate our framework's ability to maintain stable object identities and spatial relationships across frames while preserving fine-grained visual details. Motion trajectories appear natural and physically plausible, and the overall scene composition remains coherent throughout the temporal sequence. These qualitative improvements align with the quantitative metrics, confirming that our unified slot-based approach successfully captures both the spatial structure and temporal dynamics necessary for high-quality video synthesis.

\FloatBarrier
\subsubsection{Compositional Video Editing}

The most distinctive contribution of our framework lies in enabling novel compositional video editing capabilities. While the implicit conditioning variant (V1) provides competitive segmentation and generation performance, its reliance on global register tokens for spatial context limits editing precision. In practice, V1 sometimes misplaces or mis-scales objects when performing edits. To address this limitation, we adopt the \textbf{explicit pose conditioning variant (V2)} for all editing tasks. By directly fusing per-slot pose parameters into the slot embeddings, V2 provides explicit spatial grounding, enabling precise object-level manipulations while maintaining both spatial coherence and temporal consistency across entire video sequences.

\begin{figure*}[htbp]
    \centering
    \includegraphics[width=0.8\linewidth]{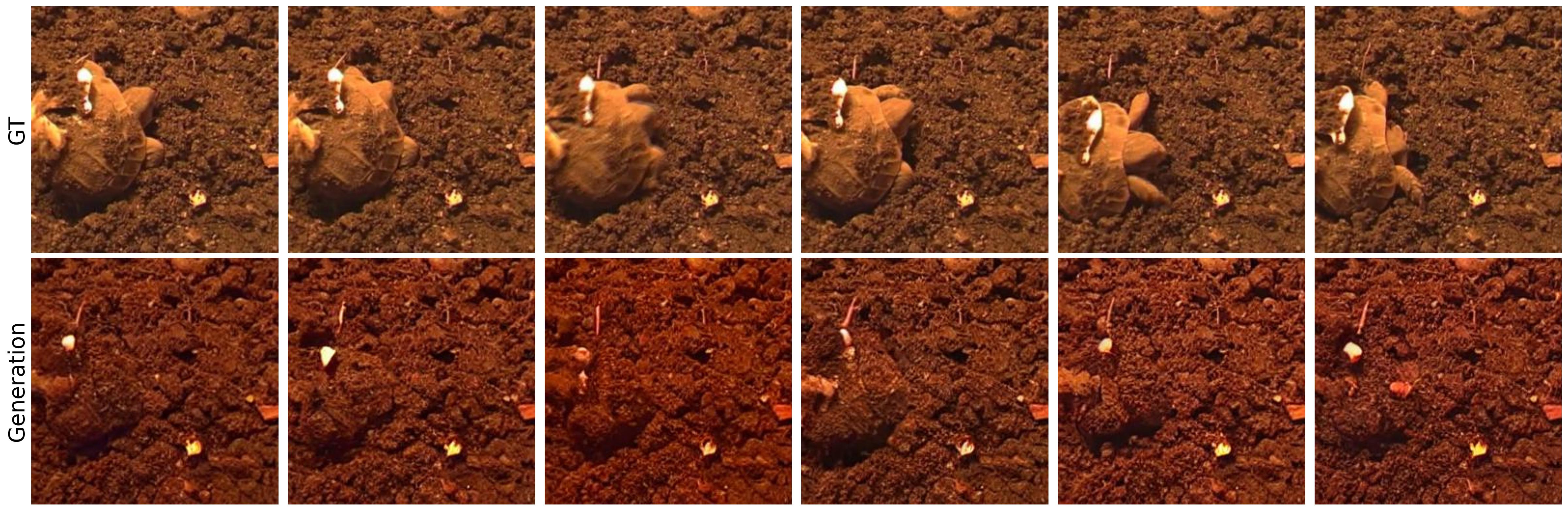}
    \includegraphics[width=0.8\linewidth]{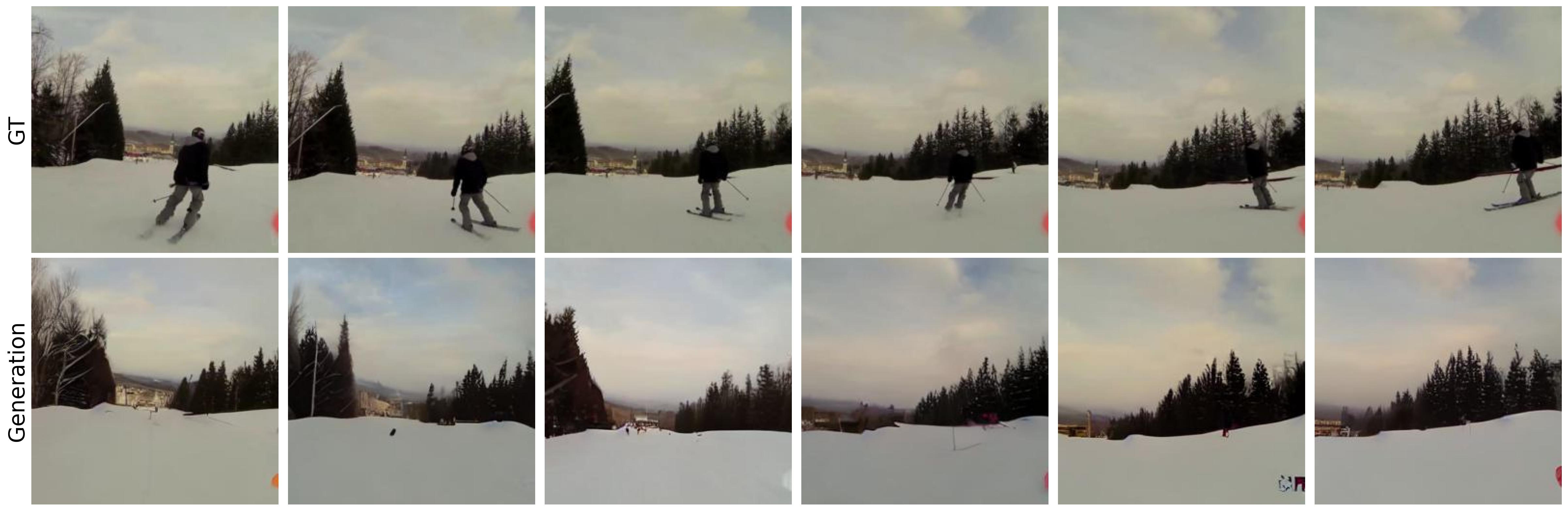}
    \includegraphics[width=0.8\linewidth]{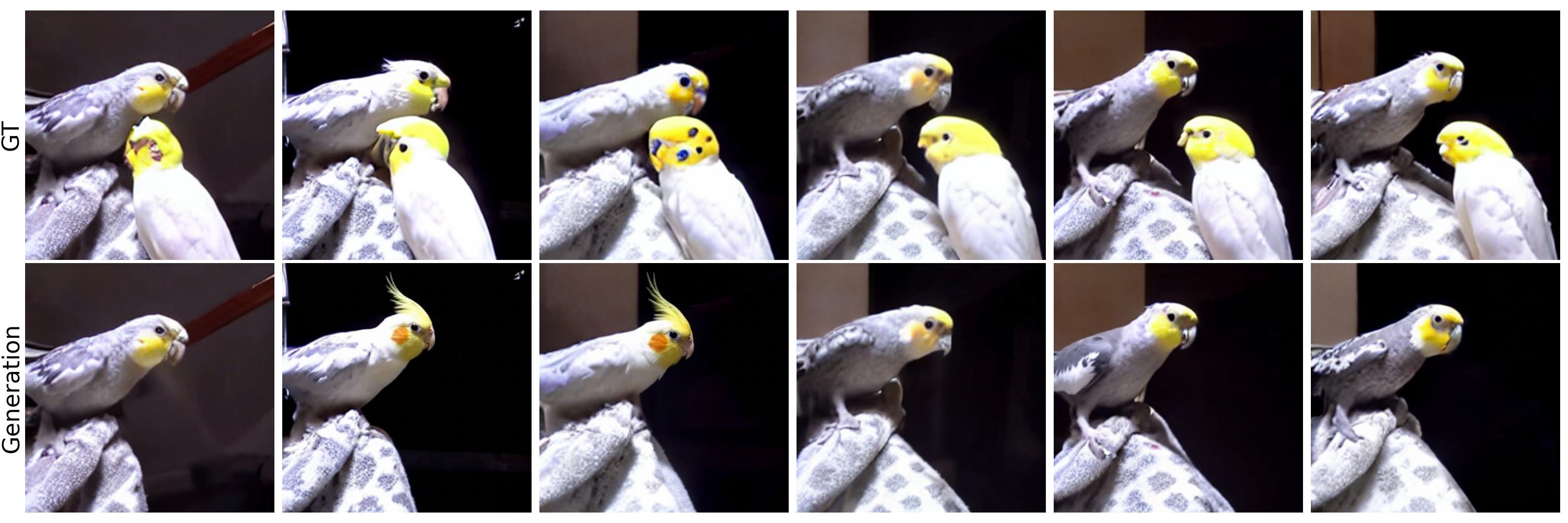}
    \caption[Object Removal Examples]{\textbf{Object Removal Examples.} Explicit conditioning (V2) enables clean object deletion across diverse video sequences. The removed entities are replaced with realistic background completions, while motion, lighting, and temporal dynamics of the remaining scene elements are preserved.}
    \label{fig:comp-gen-multi-frame-2}
\end{figure*}

Figure~\ref{fig:comp-gen-multi-frame-2} showcases targeted object removal across diverse scenarios. In each sequence, the specified object is successfully removed, and the model generates realistic background completions that seamlessly fill the vacated regions. The temporal dynamics of the remaining entities, as well as lighting, shadows, and occlusion patterns, remain consistent across frames. Even in the presence of complex interactions, scene coherence is preserved.

\begin{figure*}[htbp]
    \centering
    \includegraphics[width=0.8\linewidth]{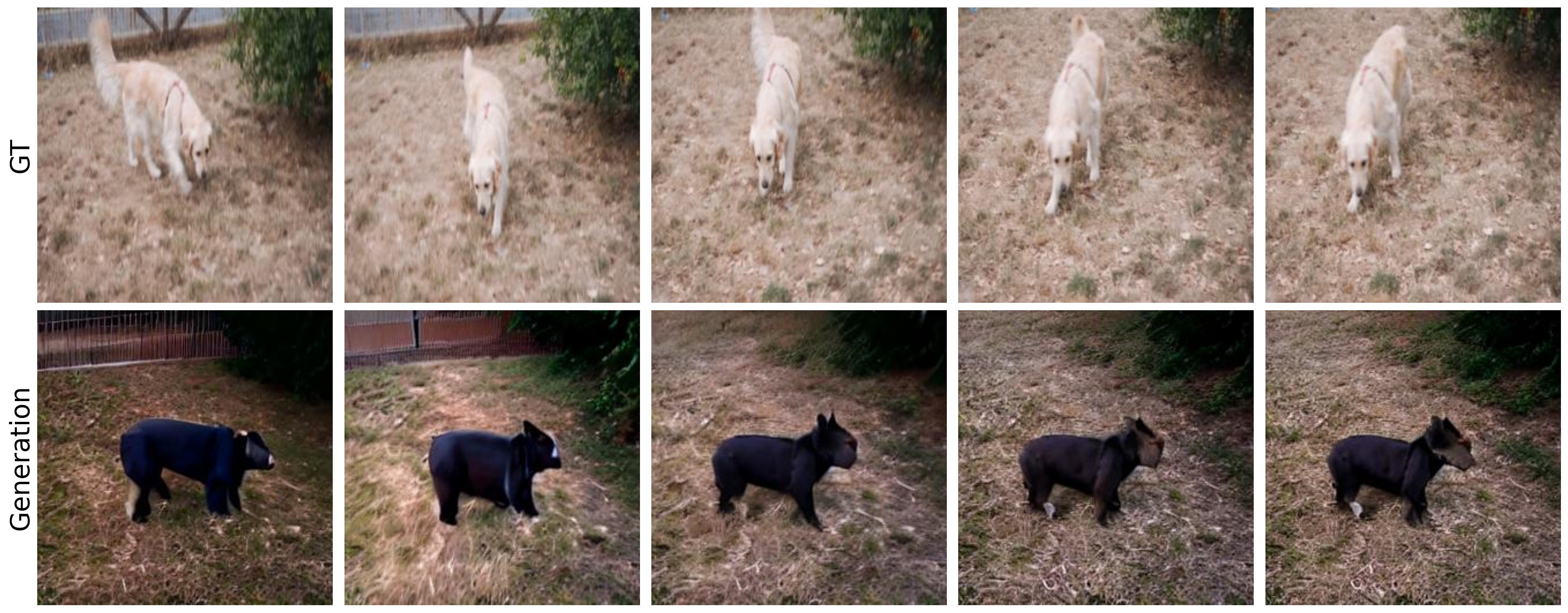}
    \includegraphics[width=0.8\linewidth]{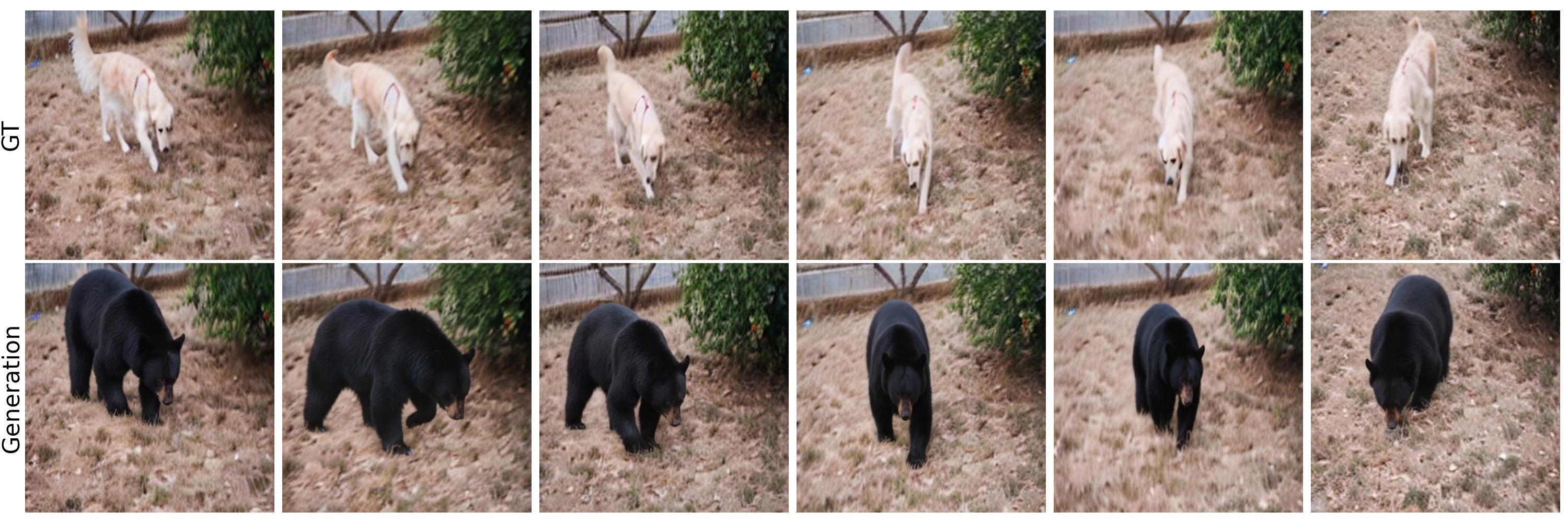}
    \caption[Object Replacement Comparison]{\textbf{Object Replacement Comparison.} Replacement results with implicit conditioning (V1) and explicit conditioning (V2). While V1 generates the correct new object but often with incorrect position or scale, V2 leverages explicit pose information to place and size the replacement (bear) correctly, producing spatially coherent results.}
    \label{fig:comp-gen-multi-frame-3}
\end{figure*}

Figure~\ref{fig:comp-gen-multi-frame-3} compares object replacement results from V1 and V2. Both models generate the correct new object identity, but V1 often struggles with positioning and scale, leading to spatial inconsistencies. In contrast, V2 accurately replaces the object—in this case, a bear—at the correct location and scale, achieving seamless integration into the scene. This highlights the advantage of explicit pose conditioning for reliable and controllable replacements.

\begin{figure*}[htbp]
    \centering
    \includegraphics[width=0.8\linewidth]{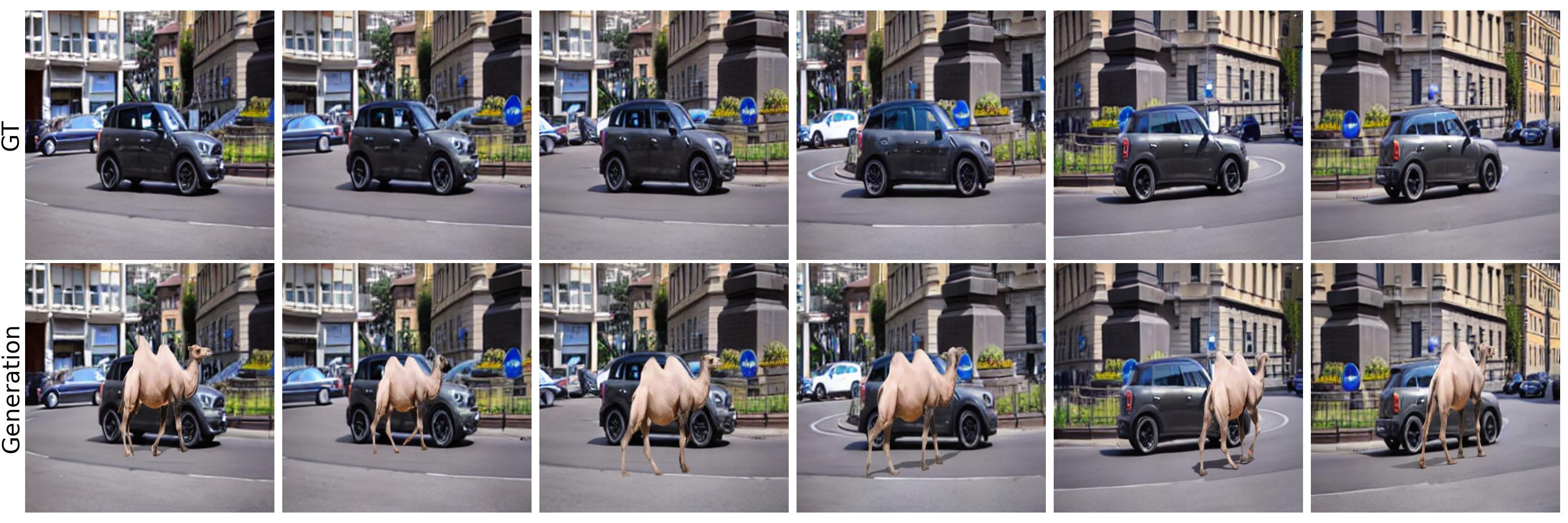}
    \caption[Object Addition Example]{\textbf{Object Addition Example.} Explicit conditioning (V2) allows insertion of a new object into the scene with the correct scale and position, while maintaining background integrity and temporal consistency across the sequence.}
    \label{fig:comp-gen-multi-frame-4}
\end{figure*}

Finally, Figure~\ref{fig:comp-gen-multi-frame-4} demonstrates object addition. A new entity is inserted into the sequence with the correct location and scale, while the background and existing scene elements remain temporally consistent. This example illustrates that explicit conditioning naturally extends to controlled addition, further broadening the range of compositional editing operations supported by our framework.

The quantitative evaluation of compositional editing performance is presented in Table~\ref{table:comp-gen-quan}. Both of our variants outperform prior baselines by a large margin. Comparing our two designs, explicit pose conditioning (Ours-V2) yields modest but consistent gains over the implicit variant (Ours-V1), particularly in perceptual quality (LPIPS, FID) and structural fidelity (SSIM). These results align with our qualitative observations, confirming that explicit pose information improves spatial precision and makes compositional generation more reliable.

\FloatBarrier

\begin{table}[h]
    \caption[Compositional generation performance]{\textbf{Compositional generation performance.} We evaluate compositional generation by mixing slots from different batch samples. Results are reported for both our implicit conditioning variant (Ours-V1) and explicit conditioning variant (Ours-V2). V2 achieves stronger performance, particularly on spatially precise metrics.\vspace{-0.3cm}}
    \label{table:comp-gen-quan}
    \centering
    \small
    \setlength{\tabcolsep}{3pt}
    \begin{tabular}{lcccc}
    \toprule
    \textbf{Method} & PSNR$\uparrow$ & SSIM$\uparrow$ & LPIPS$\downarrow$ & FID$\downarrow$ \\
    \midrule
    \multicolumn{5}{c}{\textit{YTVIS}} \\
    \midrule
    LSD & 8.05 & 0.0527 & 0.8734 & 127.49 \\
    SlotDiffusion & 7.83 & 0.0521 & 0.9486 & 129.33 \\
    SlotAdapt & 9.49 & 0.0575 & 0.7313 & 112.79 \\
    Ours-V1 & 10.07 & 0.0687 & 0.629  & 83.45 \\
    Ours-V2 & \textbf{10.42} & \textbf{0.0713} & \textbf{0.587} & \textbf{78.12} \\
    \midrule
    \multicolumn{5}{c}{\textit{DAVIS}} \\
    \midrule
    LSD & 5.22 & 0.0280 & 1.0234 & 198.23 \\
    SlotDiffusion & 6.04 & 0.0310 & 0.9912 & 187.13 \\
    SlotAdapt & 6.93 & 0.0320 & 0.9721 & 172.39 \\
    Ours-V1 & 8.27 & 0.0650 & 0.694  & 113.86 \\
    Ours-V2 & \textbf{8.64} & \textbf{0.0691} & \textbf{0.653} & \textbf{107.42} \\
    \bottomrule
    \end{tabular}
    \vspace{-0.65cm}
\end{table}

\FloatBarrier

\subsection{Discussion of Limitations}
\label{sec:video_limitations}
While this work represents a significant step forward, it is important to acknowledge its limitations. First, although our segmentation performance is strong for a unified model designed for both perception and generation, specialized methods that are solely optimized for segmentation can achieve higher overlap-based scores (mIoU). This represents an inherent trade-off: while unified models enable generative capabilities, they may not match specialized methods on segmentation-specific metrics.

Second, our framework's generative quality is ultimately dependent on a pretrained \textit{image} diffusion model. While our 1-frame training strategy is highly effective at adapting this image model for video, future work could explore integrating dedicated video diffusion backbones. Such an approach might allow for even more sophisticated, end-to-end modeling of temporal dynamics and could potentially improve long-range temporal coherence further.

Finally, like most current models in this field, our framework is evaluated on relatively short video clips of moderate resolution. Scaling these object-centric approaches to generate high-resolution, long-form video content remains a critical open challenge for future research. Moreover, while our method enables compositional editing operations such as object insertion, deletion, and replacement, these manipulations are not without limitations. Object addition can sometimes yield artifacts when inserted entities must interact with complex backgrounds or lighting conditions, and replacement may introduce inconsistencies in motion or occlusion handling. Addressing these limitations will require future advances in disentangled object-slot representations and dedicated mechanisms for interaction-aware editing.

\section{Chapter Summary}
\label{sec:video_summary}
This chapter extended the object-centric, compositional framework developed for images in Chapter 4 to the significantly more complex and dynamic domain of video. We introduced a novel architecture that combines a temporally-aware object encoder, featuring Invariant Slot Attention and a Transformer-based aggregator, with an efficient slot-conditioned diffusion decoder. Our key methodological innovations, particularly the use of ISA to learn pose-invariant representations and the 1-frame training strategy to adapt a pretrained image model for video dynamics, allowed our framework to achieve state-of-the-art performance in the nascent field of object-centric video generation.

More importantly, this work is the first to bridge the gap between unsupervised object discovery and high-quality, controllable content creation in video. By learning an explicitly manipulable, object-centric representation of dynamic scenes, we have enabled, for the first time, direct and flexible compositional video editing from purely unsupervised representations. Having established and validated a robust framework for both static and dynamic compositional synthesis, this thesis lays the groundwork for a new class of generative models endowed with a structured, human-like understanding of dynamic visual scenes. 

\newpage
\chapter{Conclusion}
\label{chap:conc}

This dissertation addresses a central goal of artificial intelligence: enabling machines to perceive and generate the visual world a structured, object-centric manner, closer to human vision. We do not experience our surroundings as a flat field of pixels, but as a space populated by distinct objects that move, interact, and persist over time. This work has shown that by building computational models that learn these fundamental components of visual scenes, it is possible to create generative systems capable of a deeper, more structured form of visual understanding. By developing a framework that unites object-centric decomposition with the generative power of pre-trained diffusion models, we addressed key challenges in static image generation and extended these solutions to video, enabling compositional synthesis and editing.

\section{Summary of Contributions}

The research presented in this thesis makes two primary and interconnected contributions to the fields of generative modeling and computer vision.

First, in Chapter~\ref{chap:slotadapt}, we introduced SlotAdapt, a novel and efficient framework for adapting large, pre-trained text-to-image diffusion models for object-centric image generation. This work directly addressed the ``pre-trained versus from-scratch" dilemma that has been a significant challenge in the field. By introducing lightweight adapter layers for dedicated slot-based conditioning and a register token mechanism to disentangle global scene information from object-specific representations, SlotAdapt successfully utilizes the significant generative capabilities of pretrained models while avoiding text-conditioning bias. Complemented by a self-supervised attention guidance loss, the framework achieves state-of-the-art performance in unsupervised object discovery on complex, real-world datasets. More importantly, it is the first to demonstrate high-fidelity, zero-shot compositional image editing—including object removal, replacement, and addition—from purely unsupervised representations, marking a substantial step towards more controllable and interpretable generative modeling.

Second, building upon this foundation, in Chapter~\ref{ch:video} we extended the compositional framework to the temporal domain with a novel method for object-centric video synthesis. This required a fundamental re-engineering of the static model to address the challenges of temporal consistency and motion modeling. By integrating Invariant Slot Attention (ISA) to learn pose-invariant object identities and a Transformer-based temporal aggregator to capture object dynamics, our model learns a coherent, manipulable representation of video content. A highly efficient 1-frame training strategy allows us to adapt a pretrained image diffusion model to understand and render video, bypassing the need to train a full video diffusion model from scratch. This approach not only sets a new state-of-the-art for object-centric video generation but, crucially, is the first to enable direct and flexible compositional video editing by manipulating learned, unsupervised video object slots. This work reframes video from a sequence of raw pixels into a dynamic, editable representation of persistent objects.

\section{Limitations and Future Directions}

While the proposed frameworks show clear progress, they also have several limitations that open directions for future research.

\textbf{Segmentation--Generation Trade-off.} Our model balances segmentation and generation in a single framework. This comes with a trade-off: segmentation scores are competitive but still below specialized segmentation-only models that can focus entirely on mask accuracy. One way forward is to add a lightweight segmentation head that refines the encoder’s boundaries without disturbing the generation pipeline. On the generative side, our approach relies on a pretrained image diffusion backbone. This limits long-range motion modeling and can also cause temporal flickering, especially on datasets such as YTVIS where the frame rate is low. Adopting dedicated video diffusion backbones could mitigate these issues by explicitly modeling frame-to-frame continuity, improving both temporal dynamics and visual consistency.

\textbf{Background Preservation in Editing.} When performing editing operations, the background is not always preserved correctly. This is partly due to the stochastic nature of diffusion models, which are built for sampling rather than reconstruction. As a result, they cannot perfectly recover the original background even when conditioned on the same slots. A possible solution is to combine inversion with feature injection during generation. For example, Plug-and-Play \cite{Tumanyan_2023_CVPR} uses intermediate attention features to guide the model so that only the edited object changes while the rest of the scene remains stable. A similar strategy could be applied here: first invert the source image to noise, then generate the manipulated version by injecting diffusion features while modifying the slots. This would anchor the unchanged parts of the scene more reliably.

\textbf{Challenges in Object Addition.} Editing operations do not perform equally well. Object removal works because the background is already represented in the register token and slots, so the diffusion model can plausibly fill the gap. Replacement also works well, because the new object inherits the anchor of the old slot: its pose, spatial context, and sometimes even associated shadows. Addition, however, is harder. There is no anchor in the scene for the new slot, so the model must invent both the object and its integration with the background. This often leads to objects that look pasted in, without natural blending, shadows, or depth cues. Future work could focus on adding local lighting models, depth- or plane-aware slot embeddings, or modules that harmonize inserted objects with their surroundings.

\textbf{Scaling to Long-Form, High-Resolution Video.} All experiments were on short, medium-resolution clips, which is standard in this field. Scaling to long, high-resolution video remains difficult. Doing so will require more efficient temporal aggregation, for example by compressing slot histories, and architectures that can keep object identities stable for hundreds or thousands of frames.

\textbf{Towards Interactive and Causal Reasoning.} The current framework allows compositional editing but does not model causal or physical relations between objects. A promising step is to combine object-centric slots with intuitive physics. Training in controlled environments with explicit physical rules could push slots to represent not only appearance but also mass, friction, or affordances. This would allow the model to go beyond recomposing scenes and predict realistic physical outcomes.

\textbf{Downstream Applications.} The slot-based representations are also useful for downstream tasks. For example, reinforcement learning agents could use them directly for decision-making in robotics or autonomous navigation. They could also be decoded into specific predictions like depth, flow, or semantics. This would give a flexible bridge between generative modeling and embodied AI systems.

\section{Concluding Remarks}

This thesis was motivated by the goal of developing generative models that interpret the world as a structured, compositional, and dynamic environment, rather than as a simple collection of pixels. By successfully integrating the principles of object-centric learning with the power of modern diffusion models, we have created a framework that learns to discover, represent, and generate objects with high fidelity in both images and videos. The ability to perform direct compositional editing through learned, unsupervised object representations represents a qualitative shift, moving generative AI from pure synthesis towards structured, object-level creation. The research presented here lays a foundation for future models with a more structured, object-centric understanding of dynamic visual scenes. This work opens new possibilities for creative applications, scientific simulation, and artificial intelligence that can reason about the world in terms of its constituent parts.

\bibliographystyle{apalike}
\bibliography{references}

\end{document}